\documentclass[10pt,journal,compsoc]{IEEEtran}

\ifCLASSOPTIONcompsoc
  \usepackage[nocompress]{cite}
\else
  \usepackage{cite}
\fi

\usepackage{graphicx}
\usepackage{amssymb,amsmath,epsfig}
\usepackage{algorithm}
\usepackage{algorithmic}
\usepackage{booktabs}
\usepackage{mathrsfs}
\usepackage{color}
\usepackage{textcomp}
\usepackage{bbding}
\usepackage{pifont}
\usepackage{wasysym}
\usepackage{amssymb}
\usepackage{subcaption}
\usepackage{cite}
\usepackage{amsmath,amssymb,amsfonts}
\usepackage{xcolor,colortbl}
\usepackage{overpic}
\usepackage{rotating}
\usepackage{microtype}
\usepackage{multirow}
\usepackage{makecell}
\usepackage{hhline}
\usepackage[american]{babel}
\usepackage{cuted}
\usepackage{ragged2e}

\definecolor{lightgray}{gray}{.92}
\definecolor{tinygray}{gray}{.96}

\newcommand{\etal}{\textit{et al}.}

\newcommand{\eg}{\textit{e}.\textit{g}.}

\newcommand{\tabincell}[2]{\begin{tabular}{@{}#1@{}}#2\end{tabular}}

\usepackage{multirow}
\usepackage{epsfig}
\usepackage{adjustbox}
\usepackage[pagebackref=false,
            breaklinks=true,colorlinks,
            citecolor=cyan,linkcolor=red,
            bookmarks=false]{hyperref}

\begin{document}
%
\title{Survey on Deep Face Restoration: From Non-blind to Blind and Beyond}

\author{
    Wenjie Li,
    Mei Wang,
    Kai Zhang,
    Juncheng Li,
    Xiaoming Li,
    Yuhang Zhang,
    Guangwei Gao$^*$,~\IEEEmembership{Senior Member,~IEEE,}
    Weihong Deng$^*$,~\IEEEmembership{Member,~IEEE}
    and Chia-Wen Lin,~\IEEEmembership{Fellow,~IEEE}
        \\
	\IEEEcompsocitemizethanks{
        \IEEEcompsocthanksitem ~\textit{$^*$: Corresponding author.}
		\IEEEcompsocthanksitem ~\textit{Wenjie Li, Mei Wang, Yuhang Zhang and Weihong Deng are with the Pattern Recognition and Intelligent System Laboratory, School of Artificial Intelligence, Beijing University of Posts and Telecommunications, Beijing, China. (e-mail: \{cswjli, wangmei1, zyhzyh, whdeng\}@bupt.edu.cn).}
        \IEEEcompsocthanksitem ~\textit{Kai Zhang is with the Computer Vision Lab, ETH Z\"{u}rich, Z\"{u}rich, Switzerland (e-mail:
        kai.zhang@vision.ee.ethz.ch).}
        \IEEEcompsocthanksitem ~\textit{Juncheng Li is with the School of Communication and Information Engineering, Shanghai University, Shanghai, China. (e-mail: cvjunchengli@gmail.com).}
        \IEEEcompsocthanksitem ~\textit{Xiaoming Li is with the Nanyang Technological University, Singapore. (e-mail:
        csxmli@gmail.com).}
        \IEEEcompsocthanksitem ~\textit{Guangwei Gao is with the Intelligent Visual Information Perception Laboratory, Institute of Advanced Technology, Nanjing University of Posts and Telecommunications, Nanjing, China. (e-mail: csggao@gmail.com).}
        \IEEEcompsocthanksitem ~\textit{Chia-Wen Lin is with the Department of Electrical Engineering, National Tsing Hua University, Hsinchu, Taiwan. (e-mail: cwlin@ee.nthu.edu.tw).}
	}
}

%


\IEEEtitleabstractindextext{
\begin{abstract}
\justifying
Face restoration (FR) is a specialized field within image restoration that aims to recover low-quality (LQ) face images into high-quality (HQ) face images. Recent advances in deep learning technology have led to significant progress in FR methods. In this paper, we begin by examining the prevalent factors responsible for real-world LQ images and introduce degradation techniques used to synthesize LQ images. We also discuss notable benchmarks commonly utilized in the field. Next, we categorize FR methods based on different tasks and explain their evolution over time. Furthermore, we explore the various facial priors commonly utilized in the restoration process and discuss strategies to enhance their effectiveness. In the experimental section, we thoroughly evaluate the performance of state-of-the-art FR methods across various tasks using a unified benchmark. We analyze their performance from different perspectives. Finally, we discuss the challenges faced in the field of FR and propose potential directions for future advancements. The open-source repository corresponding to this work can be found at \textit{\url{https://github.com/24wenjie-li/Awesome-Face-Restoration}}.
\end{abstract}

}

\maketitle

\IEEEdisplaynontitleabstractindextext

\IEEEraisesectionheading{\section{Introduction}}
\IEEEPARstart{F}{ace} restoration (FR) aims to improve the quality of degraded face images and recover accurate and high-quality (HQ) face images
from low-quality (LQ) face images.
This process is crucial for various downstream tasks such as face detection~\cite{yang2016wider}, face recognition~\cite{huang2008labeled}, and 3D face reconstruction~\cite{zhang2022learning}. The concept of face restoration was first introduced by Baker \emph{et al.}~\cite{baker2000hallucinating} in 2000. They developed a pioneering prediction model to enhance the resolution of low-resolution face images. Since then, numerous FR methods have been developed, gaining increasing attention from researchers in the field. Traditional FR methods primarily involve deep analysis of facial priors and degradation approaches. However, these methods often struggle to meet engineering requirements. With breakthroughs in deep learning technology, a multitude of deep learning-based methods specifically designed for FR tasks have emerged. Deep learning networks, utilizing large-scale datasets, are capable of effectively capturing diverse mapping relationships between degraded face images and real face images. Consequently, deep learning-based FR methods~\cite{chen2018fsrnet,wang2021towards} have demonstrated significant advantages over traditional methods, offering more robust solutions.

\begin{figure}[t]
\centering
\includegraphics[width=0.9\linewidth]{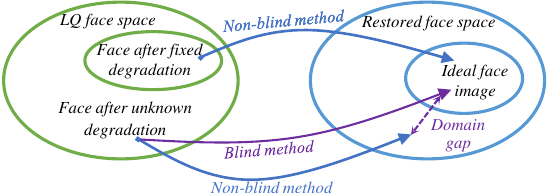}
\vspace{-3pt}
\caption{\small Domain interpretation of differences between non-blind and blind method. If the degradation factors affecting the stochastic LQ face differ from those assumed by the non-blind method (\eg, bicubic downsampling or fixed blur kernel), it can result in a significant domain gap between the restored face image and the ideal HQ face image.}
\label{fig:domain}
\vspace{-6mm}
\end{figure}

\begin{figure*}[t]
\centering
\includegraphics[width=1\textwidth]{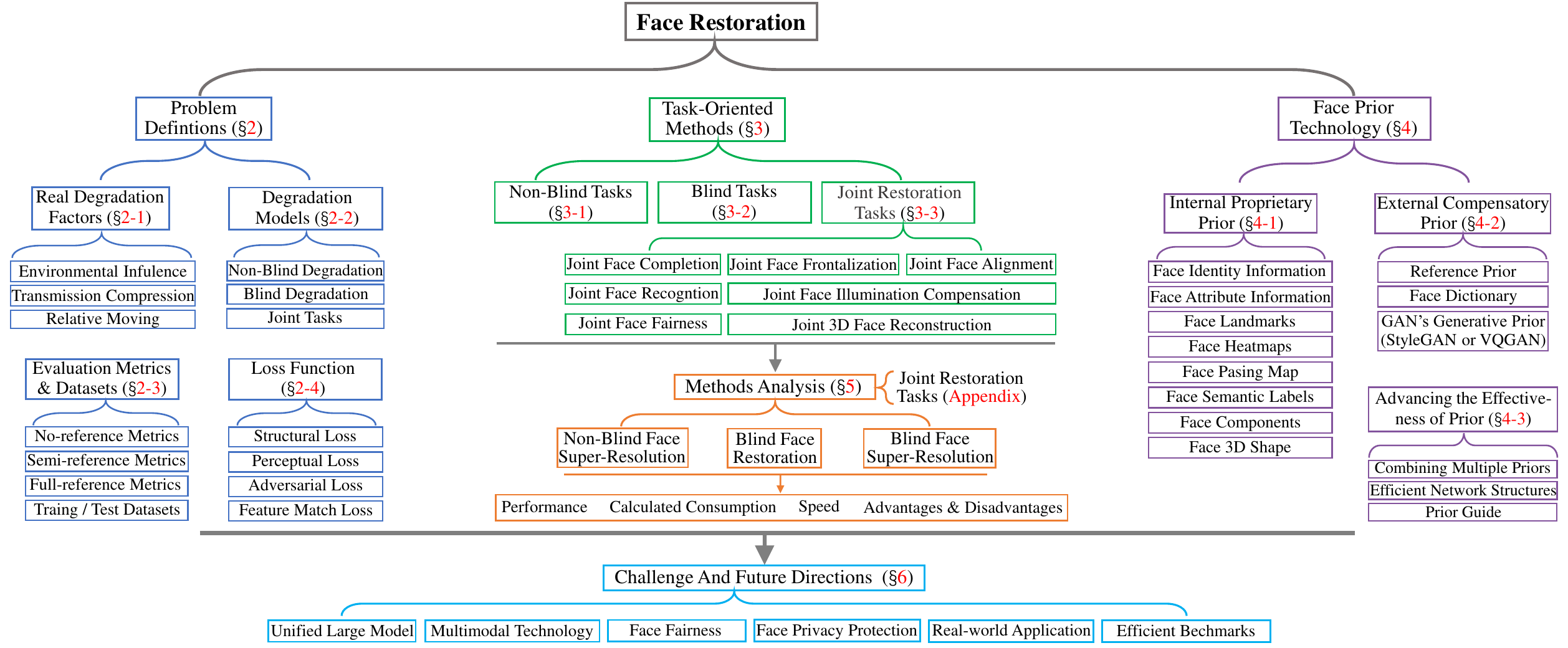}
\vspace{-5mm}
\caption{\small Outline of our deep learning-based face restoration survey.}
\label{overview}
\vspace{-4mm}
\end{figure*}

Most deep learning-based face restoration methods are trained using a fully supervised approach, where HQ face images are artificially degraded to synthesize paired LQ face images for training. In earlier non-blind methods~\cite{yu2016ultra, jiang2018deep, chen2018fsrnet}, HQ face images were degraded using fixed degradation techniques, typically bicubic downsampling. However, as shown in Fig.~\ref{fig:domain}, when the model is trained on LQ facial images synthesized in this specific manner, there can be a notable domain gap between the restored facial images and ideal HQ facial images. To address this issue, blind methods~\cite{li2018learning, wang2021towards, yang2021gan} have been developed. These methods simulate the realistic degradation process by incorporating an array of unknown degradation factors such as blur, noise, low resolution, and lossy compression. By considering more complex and diverse degradation scenarios and accounting for variations in poses and expressions, blind restoration methods have proven to be more applicable to real-world scenarios. Furthermore, a series of joint face restoration tasks have emerged to tackle specific challenges in face restoration~\cite{yu2017face, cheng2019identity, zhang2021recursive, zhong2020face, jalal2021fairness}. These tasks include joint face alignment and restoration~\cite{yu2017face}, joint face recognition and restoration~\cite{cheng2019identity}, joint illumination compensation and restoration~\cite{zhang2021recursive}, joint 3D face reconstruction and restoration~\cite{zhong2020face}, and joint face fairness and restoration~\cite{jalal2021fairness}. Building upon these advancements, our paper aims to provide a comprehensive survey of deep learning-based non-blind/blind face restoration methods and their joint tasks. By presenting this overview, we aim to shed light on the current state of development in the field, the technical approaches employed, the existing challenges, and potential directions.

\begin{table}[t]
\vspace{1.5mm}
\setlength\tabcolsep{2pt}
	\centering
	\footnotesize
	\caption{\small A summary of other deep learning based FR reviews.}
    \vspace{-2mm}
	\label{tab:different_reviews}
	\scalebox{0.95}{\begin{tabular}{l|l|l|l}
        \hline
		\toprule
        \rowcolor{lightgray}
        Year
		& Surveys        &   Related Topic  &   Venue  \\
        \hline
        \hline
		 2019 & Liu \etal \cite{liu2019survey}  & GAN-based face super-resolution & IET \\
		
		2021 & Jiang \etal \cite{jiang2021deep}  & Deep learning-based face super-resolution & CSUR  \\

        2023 & Wang \etal \cite{wang2022survey}  & Deep learning-based face restoration & Arxiv \\
		\hline
        \bottomrule
	\end{tabular}}
\vspace{-6mm}
\end{table}

Despite the rapid growth in the field of FR, there is a relative scarcity of reviews specifically focusing on deep learning-based FR methods. As depicted in TABLE~\ref{tab:different_reviews}, Liu \emph{et al.}~\cite{liu2019survey} provided a review of face super-resolution methods based on generative adversarial networks, but it solely focused on a specific technique within FR. Jiang \emph{et al.}~\cite{jiang2021deep} presented an overview of deep learning-based face super-resolution, covering FR tasks beyond super-resolution, but the emphasis remained on summarizing face super-resolution. Wang \emph{et al.}~\cite{wang2022survey} conducted a survey on FR, however, it adopted a classification pattern of sub-tasks in the image restoration domain, such as denoising, deblurring, super-resolution, and artifact removal. These patterns might not effectively generalize to existing FR methods, which could result in the omission of joint tasks related to FR. In contrast, our review provides a comprehensive summary of current FR methods from three distinct classification perspectives: blind, non-blind, and joint restoration tasks. By considering these perspectives, we not only encompass a broader range of methods related to FR but also clarify the characteristics of methods under different tasks. In the experimental section, while Wang's work~\cite{wang2022survey} primarily focused on blind methods, we conduct a comprehensive analysis of both blind and non-blind methods across various aspects. Furthermore, we provide a comparison of the methods within the joint tasks. As a result, our work provides an accurate perspective on non-blind/blind tasks and joint tasks, aiming to inspire new research within the community through insightful analysis.

The main contributions of our survey are as follows: ~\textbf{(I)} We compile the factors responsible for the degradation of real-world images and explain the degradation models used to synthesize diverse LQ face images. ~\textbf{(II)} We classify the field of FR based on blind, non-blind tasks and joint tasks criteria, providing a comprehensive overview of technological advancements within these domains. ~\textbf{(III)} Addressing the uncertainties stemming from the absence of consistent benchmarks in the field, we conduct a fair comparison of popular FR methods using standardized benchmarks. Additionally, we discuss the challenges and opportunities based on the experimental results.

Fig.~\ref{overview} provides an overview of the structure of this survey. In Section~\ref{RW}, we summarize the real-world factors contributing to the appearance of LQ face images and present corresponding artificial synthesis methods. We also discuss notable benchmarks used in the field. In Section~\ref{ME}, we introduce existing methods for different subtasks within FR. Section~\ref{PX} covers various popular priors and methods for enhancing prior validity in the restoration process. In Section~\ref{DS}, We conduct extensive experiments to compare state-of-the-art FR methods. Section~\ref{CS} addresses the challenges faced in FR and presents potential future directions. Finally, we conclude this survey in Section~\ref{ES}.

\section{Problem Definitions}~\label{RW}
In this section, we will discuss the presence of degradation factors in real-world scenarios, followed by an introduction to artificial degradation models. Additionally, we will cover commonly used loss functions, evaluation metrics, and datasets that are frequently employed in this field.


\begin{figure*}[t!]
\begin{minipage}{0.6\linewidth}
\begin{center}
\includegraphics[width=1\linewidth]{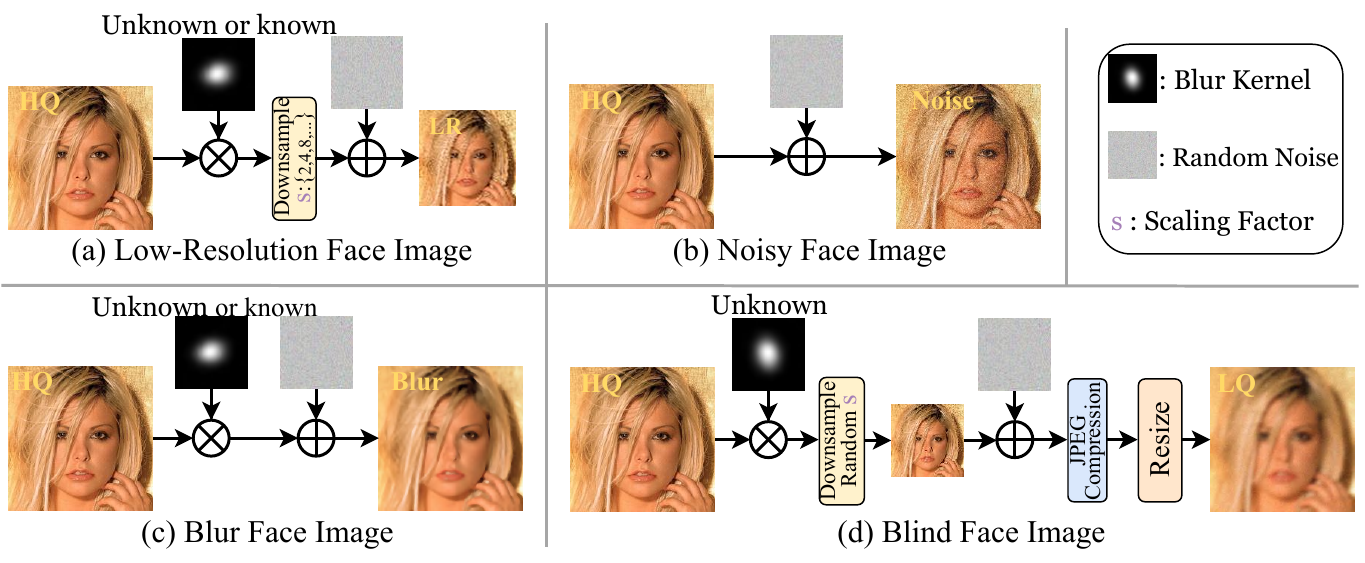}
\vspace{-9mm}
\end{center}
   \caption{\small Methods for generating various types of degradation facial images.}
\label{face_degradations}
\vspace{-3mm}
\end{minipage}
~
\begin{minipage}{0.39\linewidth}
\setlength\tabcolsep{1pt}
\vspace{-3mm}
    \captionof{table}{\small Summary of key evaluation metrics.}
    \vspace{-2mm}
	\label{tab:Evaluation_Metrics}
	\centering
	\footnotesize
	\scalebox{1}{\begin{tabular}{r||c}
		\toprule
		\rowcolor{lightgray}  Metrics   & Highlight   \\
		\hline
		\hline
		PSNR~\cite{hore2010image}   &    \tabincell{c}{Full reference, pixel-by-pixel comparison\\ of the differences between both.}
        \\ \hline  
		SSIM~\cite{wang2004image}  &  \tabincell{c}{Full reference, focus on differences in \\brightness, contrast, structure, \emph{etc.}}         \\ \hline  
        MS-SSIM~\cite{wang2003multiscale}  &  \tabincell{c}{Full reference, average SSIM for windows.} 
        \\ \hline  
        LPIPS~\cite{zhang2018unreasonable}  &  \tabincell{c}{Full reference, focus on the visual \\perceptual similarity between both.} 
        \\ \hline 
        IDD~\cite{wang2022restoreformer}  &  \tabincell{c}{Full reference, assess identity consistency.}  
        \\ \hline 
        FID~\cite{heusel2017gans}  &  \tabincell{c}{Semi-reference, measure the difference \\in distribution between both.} 
        \\ \hline 
        NIQE~\cite{mittal2012making}  &  \tabincell{c}{No reference, evaluate image naturalness.} 
        \\ \hline 
        MOS~\cite{hossfeld2016qoe}  &  \tabincell{c}{Subjective scoring by groups.} 
        \\ 
		\bottomrule
	\end{tabular}}
	\vspace{-3mm}
\end{minipage}
\end{figure*}

\begin{figure*}
\begin{center}
\includegraphics[width=0.99\linewidth]{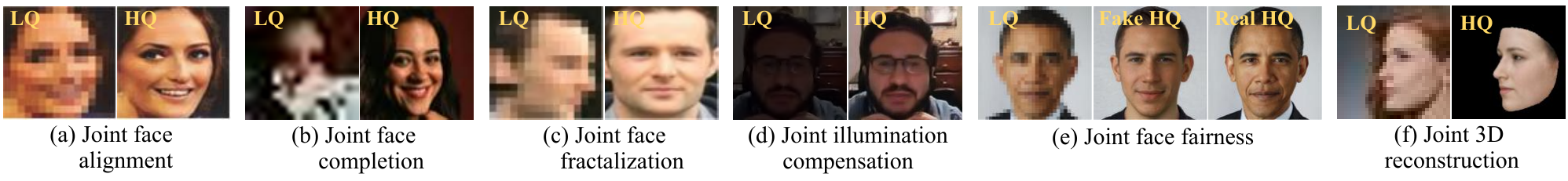}
\end{center}
\vspace{-4.5mm}
\caption{\small Demonstration of LQ and HQ face images for some joint face restoration tasks.}
\label{joint_tasks}
\vspace{-5mm}
\end{figure*}

\subsection{Real Degradation Factors}
In real-world scenarios, face images are susceptible to degradation during the imaging and transmission process due to the complex environment. The degradation of facial images is primarily caused by the limitations of the physical imaging equipment and external imaging conditions. We can summarize the main factors contributing to image degradation as follows: (1) Environmental influence: Particularly the low or high light conditions; (2) Camera shooting process: Internal factors related to the camera itself, such as optical imaging conditions, noise, and lens distortion, as well as external factors like relative displacement between the subject and the camera, such as camera shake or capturing moving face; (3) Compression during transmission: Lossy compression during image transmission and surveillance storage. To replicate realistic degradation, researchers have made various attempts. Initially, they utilized fixed blur kernels, such as Gaussian blur or downsampling, to simulate realistic blurring or low resolution. Later, randomized blur kernels were experimented with to improve robustness by introducing a wider range of degradation patterns. Additionally, considering the diversity of face-related tasks, extensive research has been conducted on joint FR tasks to recover LQ faces in specific scenes.

\vspace{-1mm}
\subsection{Degradation Models}\label{sec:degradation_models}
Due to the challenge of acquiring real HQ and LQ face image pairs, researchers often resort to using degradation models to generate synthetic LQ images ${I_{lq}}$ from HQ images ${I_{hq}}$. Generally, The ${I_{lq}}$ is the output of the ${I_{hq}}$ after degradation:
\begin{equation}
{I_{lq}} = D({I_{hq}};\delta ),
\end{equation}
where $D$ represents the degradation function and $\delta $ represents the parameter involved in the degradation process (\emph{e.g.}, the downsampling or noise or blur kernel). As shown in Fig.~\ref{face_degradations}, different $\delta $ can result in various types of degradation. Existing FR tasks can be categorized into four subtasks based on the type of degradation: face denoising, face deblurring, face super-resolution, and blind face restoration. The distinction between non-blind and blind lies in whether the degradation factors are known. TA subtask in face restoration is considered non-blind when the degradation factors are known and can be explicitly modeled. Conversely, if the degradation factors are unknown and cannot be precisely modeled, the FR task is classified as blind.

\begin{table*}[t]
    \setlength\tabcolsep{3pt}
	\centering
	\caption{\small Summary of benchmark datasets used in existing face restoration methods.}
    \vspace{-3pt}
	\label{tab:datasets}
	\scalebox{0.95}{\begin{tabular}{c|r||c|cccc|c|r}
        \hline
        \toprule
        \rowcolor{lightgray}
        Year & Dataset  & Size & Attributes & Landmarks & Parsing maps &  Identity  &  HQ-LQ  &  Methods \\
		\hline\hline

         2008
         & \textbf{LFW}~\cite{huang2008labeled}   
         & $13$K 
         & $73$ &\XSolidBrush &\XSolidBrush &\CheckmarkBold  
         & HQ  
         & C-SRIP~\cite{grm2019face}, LRFR~\cite{lai2019low}, DPDFN~\cite{jiang2020dual}, \emph{etc.}\\

        2010
        & \textbf{Multi-PIE}~\cite{gross2010multi}   
        & $755.4$K 
        & \XSolidBrush & \XSolidBrush &\XSolidBrush & \CheckmarkBold  
        & HQ  
        & FSGN~\cite{song2019joint}, CPGAN~\cite{zhang2020copy}, MDCN~\cite{shen2020exploiting}, \emph{etc.} \\

	  2011
		& \textbf{AFLW}~\cite{koestinger2011annotated}    
        & $26$K 
        & \XSolidBrush & $21$ & \XSolidBrush & \XSolidBrush  
        & HQ  
        &  FAN~\cite{kim2019progressive}, JASRNet~\cite{yin2020joint}, \emph{etc.} \\

        2011
        & \textbf{SCFace}~\cite{grgic2011scface}    
        & $4.2$K 
        & \XSolidBrush & $4$ & \XSolidBrush & \XSolidBrush  
        & HQ 
        & MNCE~\cite{jiang2018deep}, SISN~\cite{lu2021face}, CTCNet~\cite{gao2023ctcnet}, \emph{etc.}\\
        
        2012
		& \textbf{Helen}~\cite{le2012interactive}    
        & $2.3$K  
        & \XSolidBrush& $194$ & \CheckmarkBold &\XSolidBrush  
        & HQ  
        & DIC~\cite{ma2020deep}, SAAN~\cite{zhao2020saan}, SCTANet~\cite{bao2023sctanet}, \emph{etc.}\\


        2014
		& \textbf{CASIA-WebFace}~\cite{yi2014learning}  
        & $494.4$K  
        & \XSolidBrush &  2 &  \XSolidBrush & \CheckmarkBold  
        & HQ  
        & MDFR~\cite{tu2021joint}, C-SRIP~\cite{grm2019face}, GFRNet~\cite{li2018learning}, \emph{etc.}\\ 

        2015
		& \textbf{CelebA}~\cite{liu2015deep} 
        & $202.6$K 
        & $40$ & $5$ & \XSolidBrush & \CheckmarkBold  
        & HQ   
        & FSRNet~\cite{chen2018fsrnet}, SPARNet~\cite{chen2020learning}, SFMNet~\cite{wang2023spatial}, \emph{etc.}\\ 

        2016
		& \textbf{Widerface}~\cite{yang2016wider} 
        & $32.2$K 
        & \XSolidBrush & \XSolidBrush &\XSolidBrush &\XSolidBrush  
        & HQ  
        & Se-RNet~\cite{yu2021semantic}, SCGAN~\cite{hou2023semi}, \emph{etc.}\\ 

	  2017
        & \textbf{LS3D-W}~\cite{bulat2017far}  
        & $230$K  
        & \XSolidBrush & $68$ & \XSolidBrush & \XSolidBrush  
        & HQ   
        & Super-FAN~\cite{bulat2018super}, SCGAN~\cite{hou2023semi}, \emph{etc.}\\

        2017
		& \textbf{Menpo}~\cite{zafeiriou2017menpo} 
        & $9$K 
        & \XSolidBrush & $68/39$ & \XSolidBrush & \XSolidBrush 
        & HQ   
        & SAM3D~\cite{hu2020face, hu2021face}, \emph{etc.}\\

	  2018
		& \textbf{VGGFace2}~\cite{cao2018vggface2}   
        & $3310$K  
        & \XSolidBrush& \XSolidBrush & \XSolidBrush& \CheckmarkBold  
        & HQ  
        & GFRNet~\cite{li2018learning}, ASFFNet~\cite{li2020enhanced}, GWAInet~\cite{dogan2019exemplar}, \emph{etc.}\\ 

        
        2019
		& \textbf{FFHQ}~\cite{karras2019style}  
        &  $70$K 
        & \XSolidBrush& $68$ &\XSolidBrush & \XSolidBrush 
		& HQ  
        & mGANprior~\cite{gu2020image}, GFPGAN~\cite{wang2021towards}, VQFR~\cite{gu2022vqfr}, \emph{etc.} \\ 

        2020
        & \textbf{CelebAMask-HQ}~\cite{lee2020maskgan}  
        & $30$K  
        & \XSolidBrush & \XSolidBrush &\CheckmarkBold & \XSolidBrush 
		& HQ  
        & MSGGAN~\cite{karnewar2020msg}, GPEN~\cite{yang2021gan}, Pro-UIGAN~\cite{zhang2022pro}, \emph{etc.} \\

	  2022
		& \textbf{EDFace-Celeb-1M}~\cite{zhang2022edface}  
        & $1700$K  
        & \XSolidBrush & \XSolidBrush& \XSolidBrush & \XSolidBrush
		& HQ-LQ  
        & STUNet~\cite{zhang2022blind}, \emph{etc.}\\  

        2022
        & \textbf{CelebRef-HQ}~\cite{li2023learning}  
        & $10.6$K  
        & \XSolidBrush & \XSolidBrush& \XSolidBrush & \CheckmarkBold
		& HQ  
        & DMDNet~\cite{li2023learning}, \emph{etc.}\\
        \hline

		\bottomrule
	\end{tabular}}
    \vspace{-5pt}
\end{table*}

\noindent$\bullet$~\textbf{Non-blind Degradation Models.} (I) The non-blind task primarily focuses on face super-resolution (FSR)~\cite{ma2020deep}, also known as face hallucination~\cite{grm2019face}. As shown in Fig.~\ref{face_degradations} (a), its degradation model involves degrading a high-resolution (HR) face image into a low-resolution (LR) face image. When the blur kernel is pre-determined and remains constant, such as a Gaussian blur kernel or any other well-defined blur kernel, FSR can be categorized as a non-blind task. The degradation model can be described as follows:
\begin{equation}
{I_{lr}} = ({I_{hq}} \otimes {k_f}){ \downarrow _s} + n,
\end{equation}
where ${I_{lr}}$ represents the LR face image, ${I_{hq}}$ represents the HR face image, $ \otimes $ represents the convolutional operation, ${k_f}$ represents the fixed blur kernel, ${ \downarrow _s}$ denotes the downsampling operation with scale factor $s$, typically set to 4, 8, 16 and 32, and $n$ represents the additive Gaussian noise. Additionally, most researchers directly employ this degradation model to simplify the  FSR's degradation process as:
\begin{equation}
{I_{lr}} = ({I_{hq}}){ \downarrow _s}.
\label{bicubic}
\end{equation}

(II) Face denoising~\cite{cheng2021nbnet} and face deblurring~\cite{shen2018deep} primarily focus on removing additive noise from face images or simulating the removal of motion blur in a realistic face captured by a camera. Similarly, as shown in Fig.~\ref{face_degradations} (b) and (c), when the blur kernel remains constant, they can be classified as non-blind tasks. Their degradation model can be described separately as:
\begin{equation}
{I_n} = {I_{hq}} + n,
\end{equation}
\begin{equation}
{I_b} = {I_{hq}} \otimes {k_f} + n,
\end{equation}
where ${I_n}$ represents the face image containing noise, ${I_b}$ represents the blurred image, ${I_{hq}}$ represents the clean HQ face image, ${k_f}$ represents the fixed blur kernel and $n$ represents the additive Gaussian noise.

\noindent$\bullet$~\textbf{Blind Degradation Models.} (I) When the blur kernel in degradation models is randomly generated or composed of multiple unknown blur kernels, the nature of the blur kernel becomes essentially unknown. In such cases, both face super-resolution~\cite{bulat2018super} and face deblurring~\cite{lai2022face} can be classified as blind tasks. As shown in Fig.~\ref{face_degradations} (a) and (c), their degradation processes can be described separately as follows:
\begin{equation}
{I_{lr}} = ({I_{hq}} \otimes {k_u}){ \downarrow _s} + n,
\label{blind_SR}
\end{equation}
\begin{equation}
{I_b} = {I_{hq}} \otimes {k_u} + n,
\end{equation}
where ${k_u}$ is the unknown blur kernel, and the remaining variables have the same meanings as described above for non-blind face super-resolution and face deblurring.

(II) Since the above tasks focus on a single type of degradation, they face challenges in handling severely degraded face images encountered in real-world scenarios. Blind face restoration~\cite{chen2021progressive,he2022gcfsr,wang2022restoreformer} aims to address this limitation by considering more complex degradations, making it the most prominent task in the field currently. GFRNet~\cite{li2018learning} is a pioneering work in blind face restoration by introducing a more intricate degradation model aimed at simulating realistic deterioration for the first time. As shown in Fig.~\ref{face_degradations} (d), the degradation model in blind face restoration encompasses random noise, unknown blur, arbitrary scale downsampling, and random JPEG compression artifacts. This degradation process can be formulated as follows:
\begin{equation}
{I_{lq}} = \{ JPE{G_q}(({I_{hq}} \otimes {k_u}){ \downarrow _{{s_r}}} + {n_r})\} { \uparrow _{{s_r}}},
\label{blind}
\end{equation}
where $I_{lq}$ and $I_{hq}$ represent the low-quality and high-quality face images, respectively. $JPE{G_q}$ represents JPEG compression operation with arbitrary quality factor, ${k_u}$ represents an unknown blur kernel. ${ \downarrow _{{s_r}}}$ and ${ \uparrow _{{s_r}}}$ represent down-sampling and up-sampling operations with arbitrary scale factors ${s_r}$, respectively. ${n_r}$ represents random noise.

\noindent$\bullet$~\textbf{Joint Tasks.}
Due to the multitude of joint tasks, we do not introduce the degradation models for each of them individually. Fig.~\ref{joint_tasks} showcases several examples of joint tasks, depicted from left to right: (a) Joint face alignment and restoration~\cite{yu2017face}: This task addresses the challenge of misaligned faces by aligning and restoring them. (b) Joint face completion and restoration~\cite{cai2019fcsr}: The objective is to handle face occlusions and restore the missing regions in the face image. (c) Joint face frontalization and restoration~\cite{yu2019can}: This task focuses on recovering frontal faces from side faces, enhancing their appearance and quality. (d) Joint face illumination compensation and restoration~\cite{zhang2021recursive}: This task aims to restore faces captured in low-light conditions, compensating for the lack of illumination. (e) Joint face fairness and restoration~\cite{jalal2021fairness}: This task aims to improve the accuracy of face restoration across different human races, promoting fairness and inclusivity. (f) Joint 3D face reconstruction~\cite{zhang2022learning}: This task aims to improve the accuracy of 3D reconstruction of low-quality faces. In each case, the HQ face images are represented on the right, while the degraded LQ face images corresponding to each specific task are shown on the left. These joint tasks are designed to address face restoration challenges in specific scenarios and hold practical significance in their respective domains.

\subsection{Evaluation Metrics And Datasets}
We have compiled a selection of the most widely used evaluation metrics in the field of FR, as presented in TABLE~\ref{tab:Evaluation_Metrics}. We classify these metrics into three groups: full-reference metrics, which necessitate paired HQ face images; semi-reference metrics, which only require unpaired HQ face images; and no-reference metrics, which don't involve any face images for measurement. Additionally, more metrics can be found at \textit{\url{https://github.com/chaofengc/Awesome-Image-Quality-Assessment}}. Furthermore, we summarize commonly used benchmark datasets for FR in TABLE~\ref{tab:datasets}, including the number of face images, the facial features included, the availability of HQ-LQ pairs, and previous methods that have utilized these datasets. For datasets that only provide HQ images, we need to synthesize the corresponding LQ images using the degradation model introduced in Sections~\ref{sec:degradation_models}.

\begin{figure*}
\begin{center}
\includegraphics[width=0.99\linewidth]{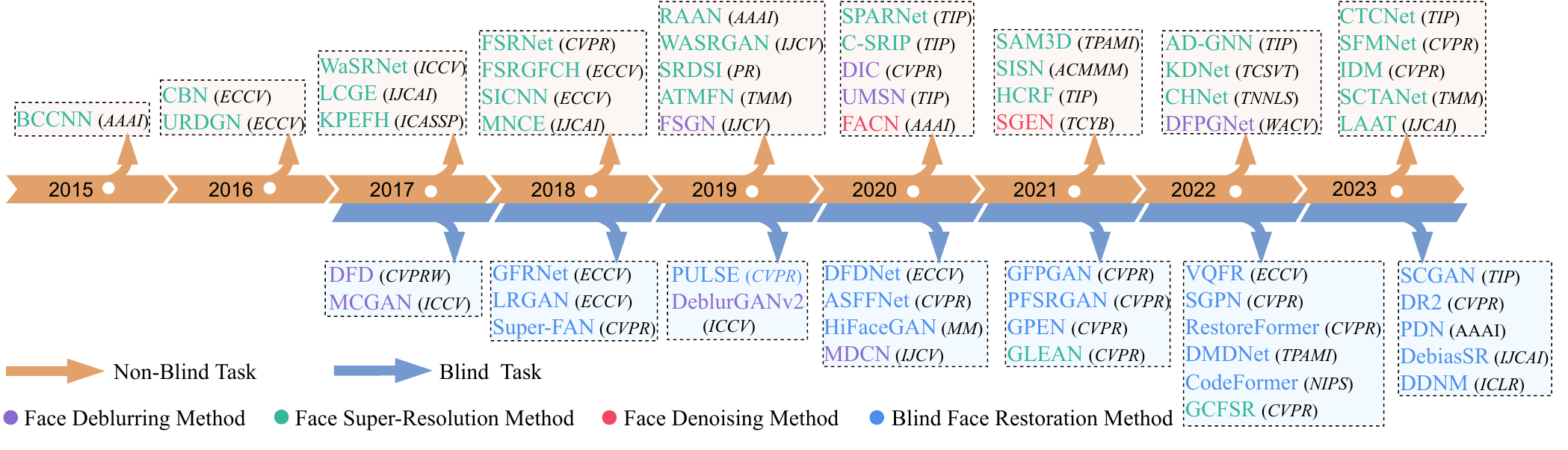}
\vspace{-5.5mm}
\end{center}
   \caption{Milestones of deep learning-based non-blind/blind task methods, including its name and venues.}
\label{Non_and_Blind_timeline}
\vspace{-4mm}
\end{figure*}

\subsection{Loss Function}
The researchers aim to estimate the approximation of the HQ face image ${I_{hq}}$, denoted as ${{\hat I}_{hq}}$, from the LQ face image ${I_{lq}}$, following:
\begin{equation}
{{\hat I}_{hq}} = {D^{ - 1}}({I_{lq}},\delta ) = F({I_{lq}},\theta ),
\end{equation}
where $F$ represents the face restoration method and $\theta $ represents the parameters of the method. During the training, the optimization process can be formulated as follows:
\begin{equation}
\hat{\theta} = \mathrm{argmin} \, L(\hat{I}_{hq}, I_{hq}),
\end{equation}
where ${\hat \theta}$ represents the optimization parameter in the training process, $L$ represents the loss between ${\hat I}_{hq}$ and ${I_{hq}}$. Different loss functions can yield varying results in face restoration. Initially, researchers commonly used structural losses; however, these losses have limitations, such as over-smoothing the output images. To overcome these limitations, perceptual losses and adversarial losses were developed. Furthermore, because of the structured nature of faces, a large number of face-specific losses have also been proposed. 

\noindent$\bullet$~\textbf{Structural loss.} 
Structural losses are employed to minimize the structural differences between two face images. The most commonly used structural losses are pixel-wise losses, which include $L1$ loss~\cite{yang2021gan,wang2021towards,wang2022restoreformer} and the $L2$ loss~\cite{chen2018fsrnet,li2020enhanced,chen2021progressive}. They can be formulated as
\begin{equation}
{L_i} = {\left\| {{I_{hq}}(h,w,c) - {{\hat I}_{hq}}(h,w,c)} \right\|_i},i \in \{ 1,2\} ,
\end{equation}
where $h$, $w$, and $c$ represent the height, width, and number of channels, respectively. The pixel-level loss also encompasses the Huber loss~\cite{kalarot2020component} and Carbonnier penalty function. Furthermore, in addition to the pixel-level losses, textural losses have been developed. These include the SSIM loss~\cite{grm2019face}, which promotes image textural similarity, and the cyclic consistency loss~\cite{hou2023semi}, which facilitates cooperation between recovery and degradation processes. While minimizing these structural losses encourages the restored image to closely match the ground truth image in terms of pixel values, resulting in a similar structure between the two face images and a higher PSNR value, there is a disadvantage. The recovered face image, however, tends to be too smooth and lacks fine details.

\noindent$\bullet$~\textbf{Perceptual loss.} 
The perceptual loss is intended to enhance the visual quality of the recovered images by comparing them to the ground truth images in the perceptual domain using a pre-trained network, such as VGG, Inception \emph{etc.}. The prevalent approach is to calculate the loss based on features extracted from specific intermediate or higher layers of the pre-trained network, as these features represent high-level semantic information within the image. Denoting the $l$-th layer involved in the computation of the pre-trained network as ${{\varphi _l}}$, its perceptual loss $L_{per}^l$ can be expressed as follows:
\begin{equation}
L_{per}^l = {\left\| {{\varphi _l}({I_{hq}}(h,w,c)) - {\varphi _l}({{\hat I}_{hq}}(h,w,c))} \right\|_2},
\end{equation}

\noindent$\bullet$~\textbf{Adversarial loss.} 
The adversarial loss is a common type of loss used in GAN-based face restoration methods~\cite{chen2021progressive,he2022gcfsr,wang2022restoreformer}. In this setup, the generator $G$ aims to generate an HQ face image to deceive the discriminator $D$, while the discriminator $D$ strives to distinguish between the generated image and the ground-truth image. The generator and discriminator are trained alternately to generate visually more realistic images. The loss can be expressed as follows:
\begin{equation}
{L_{adv,D}} = {E_{{I_{hq}}}}[\log (1 - D(G({I_{hq}}))) + \log (D({I_{hq}}))],
\end{equation}
\begin{equation}
{L_{adv,G}} = {E_{{I_{hq}}}}[\log (1 - D(G({I_{hq}})))],
\end{equation}
where ${L_{adv,G}}$ and ${L_{adv,D}}$ are the adversarial losses of the generator and discriminator, respectively. It is worth noting that the use of adversarial loss can sometimes result in training instabilities, so careful parameter tuning is necessary. Furthermore, although models trained with adversarial loss can generate visually appealing results, they may also introduce artifacts, resulting in less faithful face images.

\noindent$\bullet$~\textbf{Feature match loss.} The structured nature of the human face allows for the integration of specific structural features into the supervised process, leading to improved accuracy in restoration. These features include face landmarks~\cite{chen2018fsrnet}, face heatmaps~\cite{bulat2018super}, 3D face shape~\cite{hu2020face}, semantic-aware style~\cite{chan2021glean}, face parsing~\cite{chen2021progressive}, facial attention~\cite{kim2019progressive}, face identity~\cite{grm2019face}, and facial components~\cite{wang2021towards}. Among these, the face landmarks loss is widely utilized and can be described as
\begin{equation}
{L_{landmarks}} = \frac{1}{N}\sum\limits_{n = 1}^N {{{\left\| {l_{x,y}^n - {\hat l}_{x,y}^n} \right\|}_2}} ,
\end{equation}
where $N$ is the number of facial landmarks, and ${l_{x,y}^n}$ and ${\hat l}_{x,y}^n$ represent the coordinates of the $n$-th landmark point in the HQ face and the recovered face, respectively. Face-specific losses take into account the specific characteristics and details of facial images. By incorporating these losses, the model can better preserve facial attributes, improve facial details, and enhance the overall visual quality of the restored face.

\begin{figure*}
\begin{center}
\includegraphics[width=0.99\linewidth]{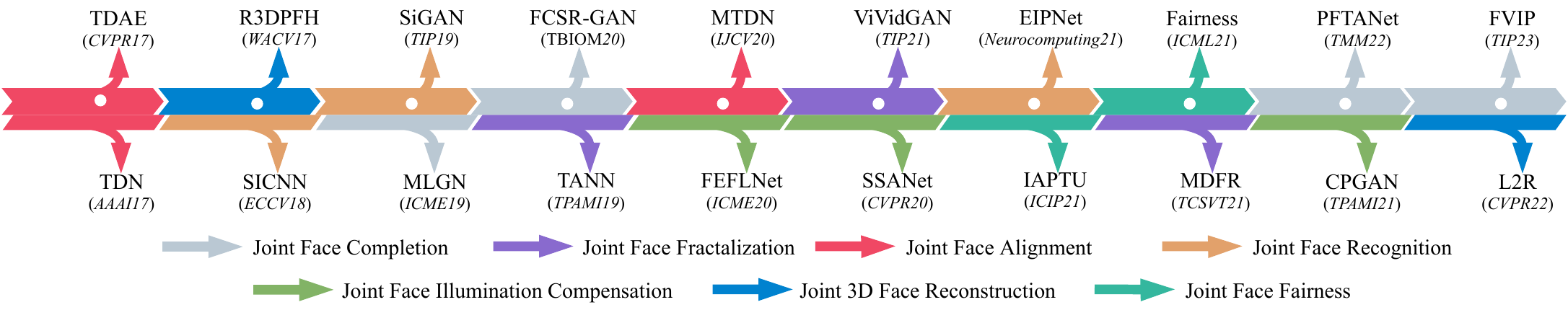}
\end{center}
\vspace{-4mm}
\caption{\small Milestones of the Joint Face Restoration methods, including its name and venues.}
\vspace{-3mm}
\label{joint_tasks_timeline}
\end{figure*}

\begin{figure*}
\begin{center}
\includegraphics[width=0.99\linewidth]{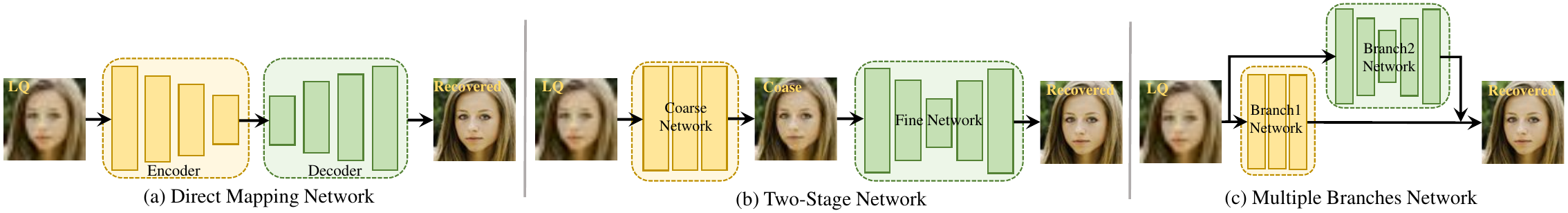}
\end{center}
\vspace{-4mm}
\caption{\small Summary of the architecture of general methods for non-blind face restoration.}
\label{general_network}
\vspace{-4mm}
\end{figure*}

\section{Task-oriented Methods}~\label{ME}
In this section, we will summarize and discuss the methodology for each of the three types of face restoration tasks: non-blind tasks, blind tasks, and joint restoration tasks. Fig.~\ref{Non_and_Blind_timeline} illustrates several notable methods in recent years that focus on non-blind and blind tasks. Fig.~\ref{joint_tasks_timeline} showcases several landmark methods in recent years that specialize in joint face restoration tasks.

\subsection{Non-blind Tasks}
The initial attempts in the field of FR primarily focused on non-blind methods. Earlier non-blind methods did not consider facial priors and directly mapped LQ images to HQ images, as depicted in Fig.~\ref{general_network} (a). One pioneering work is the bi-channel convolutional neural network (BCCNN) proposed by Zhou \emph{et al.}~\cite{zhou2015learning}, which significantly surpasses previous conventional approaches. This network combines the extracted face features with the input face features and utilizes a decoder to reconstruct HQ face images, leveraging its strong fitting capability. Similarly, other methods~\cite{feng2016face,lu2017face,chen2019sequential} also adopt direct LQ to HQ mapping networks. Subsequently, non-blind methods incorporated novel techniques, such as learning strategies and prior constraints, into the mapping network to achieve more robust and accurate face restoration. Specifically, as shown in Fig.~\ref{general_network} (b), one class of methods adopts a two-stage approach for face restoration, consisting of roughing and refining stages. For example, CBN~\cite{zhu2016deep} employs a cascaded framework to address the performance limitations observed in previous methods when dealing with misaligned facial images. LCGE~\cite{song2017learning}, MNCE~\cite{jiang2018deep}, and FSGN~\cite{song2019joint} generate facial components that approximate real landmarks and enhance them by recovering details. FSRNet~\cite{chen2018fsrnet} obtains a rough face image through a network and then refines it using a heatmap and a resolving map of facial landmarks. DIDnet~\cite{cheng2021face} and ATSENet~\cite{li2020learning} utilize facial identity or attributes to enhance the features extracted by the initial network and recover face images with higher confidence. FAN~\cite{kim2019progressive} employs a facial attention prior loss to constrain each incremental stage and gradually increase the resolution. Another class of methods adopts a multi-branch structure for facial restoration, as depicted in Fig.~\ref{general_network} (c). For example, KPEFH~\cite{li2018face} utilizes multiple branches in the network to predict key components of the face separately. FSRGFCH~\cite{yu2018face} enhances the quality of facial details by predicting the face component heatmap with an additional branching in the network. UMSN~\cite{yasarla2020deblurring} employs multiple branches to predict regions of different semantic categories of the face separately and then combines them.

\begin{figure}
\begin{center}
\vspace{0mm}
\includegraphics[width=1\linewidth]{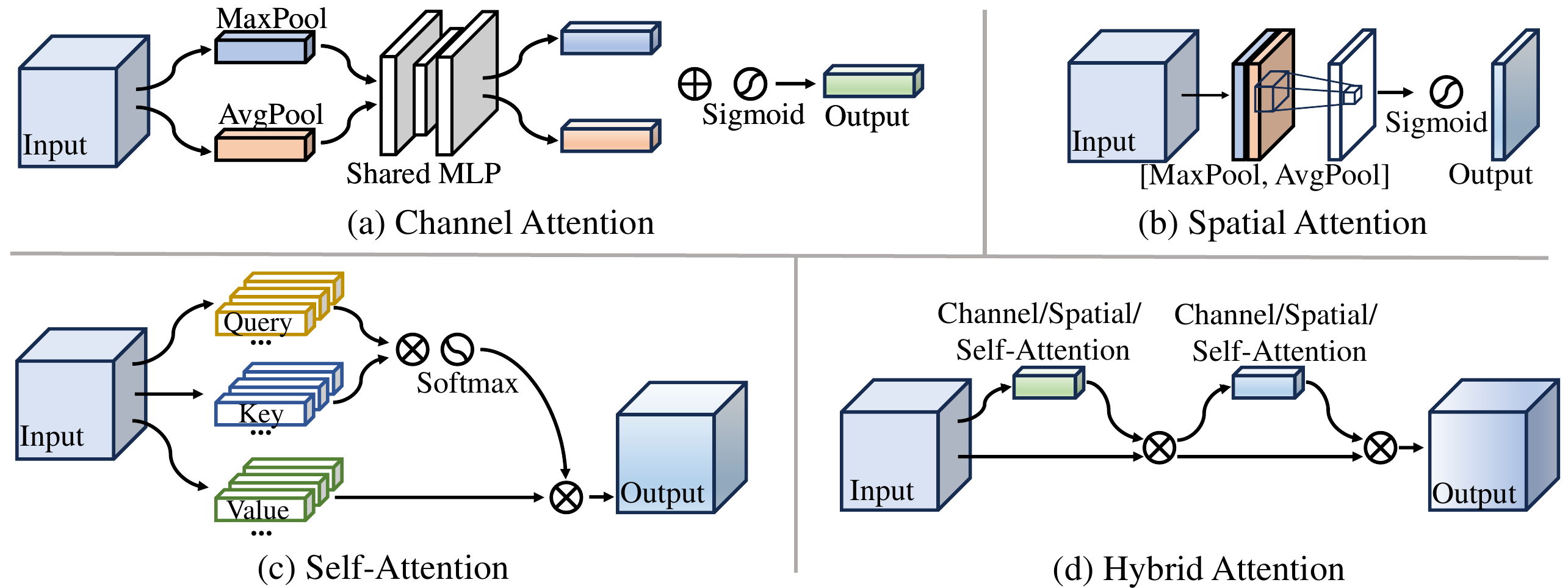}
\end{center}
\vspace{-4mm}
\caption{\small The architecture of Attention-based methods.}
\label{attention}
\vspace{-4mm}
\end{figure}

Attention mechanisms have demonstrated their effectiveness in image restoration methods~\cite{zhang2018image,liang2021swinir}. Subsequently, there has been a significant focus on integrating attention mechanisms~\cite{cao2017attention} to enhance the handling of important facial regions. Various networks based on attention mechanisms have been developed, as illustrated in Fig.~\ref{attention}. Attention can be categorized into four types: channel attention, spatial attention, self-attention, and hybrid attention. Channel attention-based approaches~\cite{jiang2019atmfn,xin2019residual,zhao2020saan,chudasama2021comsupresnet} emphasize the relative weights between different feature channels in the model, enabling selective emphasis on important channels. Spatial attention-based approaches~\cite{chen2020learning,hu2021face,lu2022rethinking} focus on capturing spatial contextual information about features, enabling the model to prioritize features relevant to key face structures. Self-attention-based approaches~\cite{wang2022faceformer,qi2023efficient,LAATransformer} mainly capture global facial information, yielding excellent performance. Some approaches~\cite{kim2019progressive,chen2020learning} also enhance individual attention mechanisms to better suit the specific requirements of FR tasks. Hybrid attention-based approaches~\cite{bao2022attention,bao2023sctanet,gao2023ctcnet} combine the aforementioned three main types of attention, aiming to leverage the advantages of different attention types to improve the overall performance of restoration models. Furthermore, some approaches leverage specific types of prior to guide the network. For instance, SAAN~\cite{zhao2020saan} incorporates the face parsing map, FAN~\cite{kim2019progressive} incorporates the face landmark, SAM3D~\cite{hu2020face,hu2021face} incorporates the 3D face information, HaPSR~\cite{wang2021heatmap} incorporates the face heatmap, and CHNet~\cite{lu2022rethinking} incorporates the face components. To direct attention more precisely, some methods have started to artificially delineate and recover different regions of the face image. WaSRNet~\cite{huang2017wavelet} employs wavelet transform to convert various regions of the image into coefficients and then performs restoration processing at different levels in the wavelet coefficient domain. SRDSI~\cite{hu2019face} uses PCA to decompose faces into low-frequency and high-frequency components and then employs deep and sparse networks to recover these two parts, respectively. SFMNet~\cite{wang2023spatial} integrates information extracted from its spatial and frequency branches, enhancing the texture of the contour.

\begin{figure*}
\begin{center}
\includegraphics[width=0.99\linewidth]{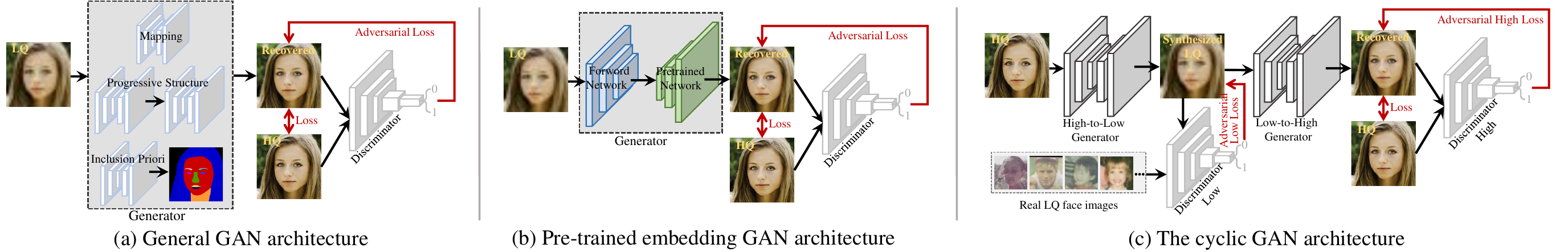}
\end{center}
\vspace{-4mm}
\caption{\small Summary of architecture of GAN-based methods for face restoration.}
\label{GAN_framework}
\vspace{-4mm}
\end{figure*}

The Generative Adversarial Network (GAN) has gained significant popularity due to its ability to generate visually appealing images. It consists of a generator and a discriminator. The generator's role is to produce realistic samples to deceive the discriminator, while the discriminator's task is to distinguish between the generator's output and real data. GAN architectures used in FR can be classified into three types: general GAN, pre-trained embedded GAN, and cyclic GAN. Non-blind methods primarily employ the general GAN structure depicted in Fig.~\ref{GAN_framework} (a). In 2016, Yu \emph{et al.}~\cite{yu2016ultra} introduced the first GAN-based face super-resolution network (URDGN). This network utilizes a discriminative network to learn fundamental facial features, and a generative network leverages adversarial learning to combine these features with the input face. Since then, many different GAN-based face restoration methods have been extended in the non-blind task, showing promising recovery results. Some methods focus on designing progressive GANs, including two- or multi-stage approaches~\cite{dou2020pca,karnewar2020msg,lin2020learning}. Others concentrate on embedding face-specific prior information, such as facial geometry~\cite{yu2021semantic,li2021organ,wang2022propagating}, facial attributes~\cite{lu2018attribute}, or identity information~\cite{zhang2020supervised} into the GAN framework. It is worth noting that given the excellent performance of GAN, many non-GAN-based methods~\cite{chen2018fsrnet,ma2020deep,chen2020learning,bao2022attention,gao2023ctcnet,bao2023sctanet} also provide a GAN version of their approach for reference. 

\begin{figure}
\begin{center}
\vspace{-1mm}
\begin{overpic}[width=1\linewidth]{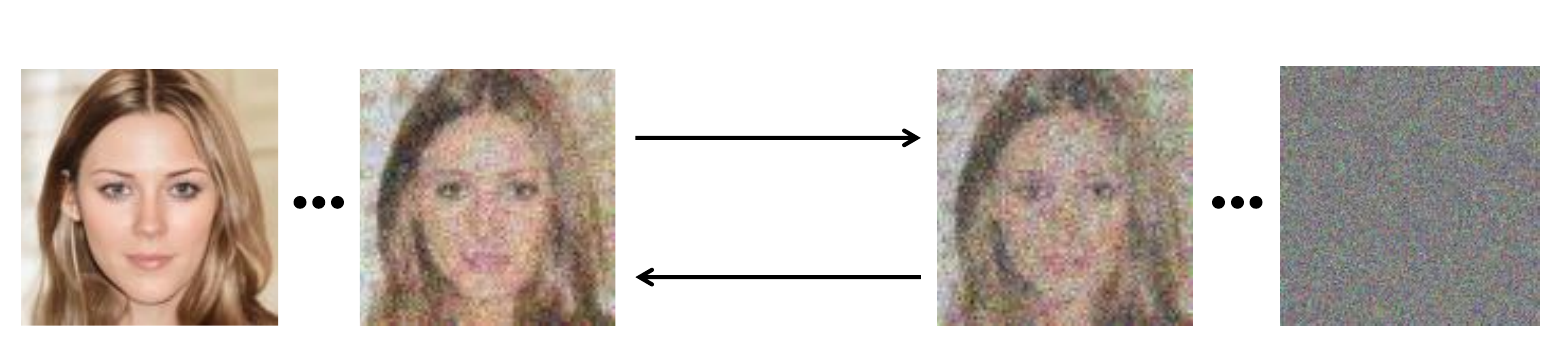}
\put(42,14.1){\color{black}{\scriptsize $q({y_t}|{y_{t\!-\!1}})$}}
\put(40,5.3){\color{black}{\scriptsize ${p_\theta }({y_{t\!-\!1}}|{y_t},x)$}}
\put(1.8,17.9){\color{black}{\scriptsize ${y_0} \sim p(y|x)$}}
\put(27.8,17.9){\color{black}{\scriptsize ${y_{t\!-\!1}}$}}
\put(66.8,17.9){\color{black}{\scriptsize ${y_t}$}}
\put(81.0,17.9){\color{black}{\scriptsize ${y_T} \sim \mathcal{N}(0,I)$}}
\end{overpic}
\end{center}
\vspace{-2.5mm}
\caption{\small The diffusion denoising principle involves two main processes: forward diffusion process $q$ and reverse inference process $p$. In the $q$, Gaussian noise is gradually added to the target image from left to right. In contrast, $p$ iteratively denoises the target image, proceeding from right to left.}
\label{diffusion_model}
\vspace{-5mm}
\end{figure}

However, GAN-driven methods often suffer from pattern collapse, resulting in a lack of diversity in the generated images. The diffusion probabilistic model (DDPM) have been proposed as an alternative approach. As shown in Fig.~\ref{diffusion_model}, given samples drawn from an unknown conditional distribution $p(y|x)$, the input-output image pair is denoted as $D = \{ {x_i},{y_i}\} $. DDPM learns the parameter approximations of $p(y|x)$ through a stochastic iterative refinement process that maps the source image $x$ to the target image $y$. Specifically, DDPM starts with a purely noisy image ${y_T} \sim \mathcal{N}(0,I)$, and the model refines the image through successive iterations $({y_{T\! - \!1}},{y_{T\! - \!2}},...,{y_0})$ based on the learned conditional transformation distribution ${p_\theta }({y_{t\! - \!1}}|{y_t},x)$, refining the image until ${y_0} \sim p(y|x)$. In 2022, SRDiff~\cite{li2022srdiff} introduced a diffusion-based model for face super-resolution.  It incorporated residual prediction throughout the framework to accelerate convergence. Then, SR3~\cite{saharia2022image} achieved super-resolution by iterative denoising the conditional images generated by the denoising diffusion probabilistic model, resulting in more realistic outputs at various magnification factors. IDM~\cite{gao2023implicit} combined an implicit neural representation with a denoising diffusion model. This allowed the model to continuous-resolution requirements and provide HQ face restoration with improved scalability across different scales.

\subsection{Blind Tasks} 
In practical applications, researchers have observed that methods originally designed for non-blind tasks often struggle to effectively handle real-world LQ face images. Consequently, the focus of face restoration is gradually shifting towards blind tasks to address a broader range of application scenarios and challenges associated with LQ images. One of the earliest blind methods is DFD~\cite{chrysos2017deep}, introduced by G. Chrysos \emph{et al.}, which employs a modified ResNet architecture for blind face deblurring. Then, MCGAN~\cite{xu2017learning} leveraged GAN techniques to significantly improve the model's robustness in tackling blind deblurring tasks. However, this approach exhibits limited efficacy when encountering more complex forms of degradation. As a result, subsequent endeavors in the realm of blind tasks have predominantly employed GAN-driven methodologies. Some methods adopt the general GAN structure depicted in Fig.~\ref{GAN_framework} (a). For example, DeblurGAN-v2~\cite{kupyn2019deblurgan}, HiFaceGAN~\cite{yang2020hifacegan}, STUNet~\cite{zhang2022blind}, GCFSR~\cite{he2022gcfsr}, and FaceFormer~\cite{li2022faceformer} all design novel and intricate network architectures for blind face restoration. Additionally, many methods use more complex GAN networks with prior information. GFRNet~\cite{li2018learning}, ASFFNet~\cite{li2020enhanced} and DMDNet~\cite{li2023learning} utilize a bootstrap network with reference to prior to guide the recovery network, employing a two-stage strategy for better face restoration. MDCN~\cite{shen2020exploiting} and PFSRGAN~\cite{chen2021progressive} employ a two-stage network consisting of a face semantic label prediction network and a recovery parsing network for reconstruction. Furthermore, Super-FAN~\cite{bulat2018super}, DFDNet~\cite{li2020blind}, and RestoreFormer~\cite{wang2022restoreformer} integrate face structure information or face component dictionary into GAN-based algorithms to enhance the quality of blind LQ facial images.

\begin{figure*}
\begin{center}
\includegraphics[width=0.99\linewidth]{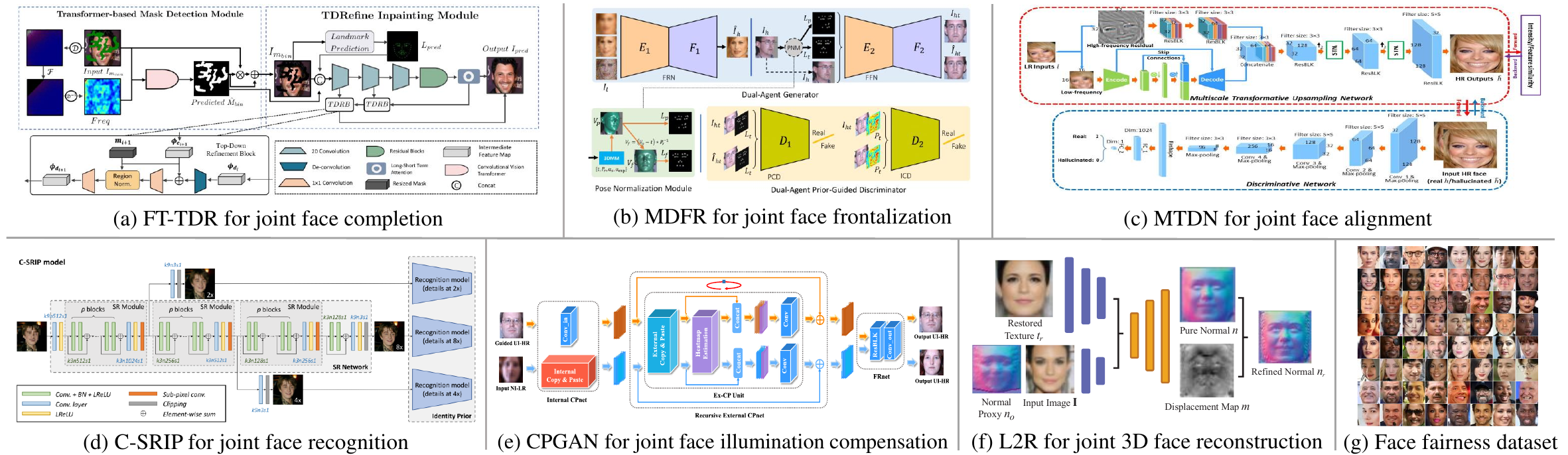}
\end{center}
\vspace{-4mm}
   \caption{\small Examples of methods for joint tasks. (a): FT-TDR~\cite{wang2022ft} for joint face complementation; (b): MDFR~\cite{tu2021joint} for joint face frontalization; (c): MTDN~\cite{yu2020hallucinating} for joint face alignment; (d): C-SRIP~\cite{grm2019face} for joint face recognition; (e): CPGAN~\cite{zhang2021recursive} for joint face illumination complementation; (f): L2R~\cite{zhang2022learning} for joint 3D face reconstruction; (g): EDFace-Celeb-1M~\cite{zhang2022edface} dataset for joint face fairness.}
\label{joint_tasks_expamles}
\vspace{-4mm}
\end{figure*}

Pre-trained GAN-based models have become the most popular approach in the field of blind face restoration since generative models~\cite{karras2019style,karras2020analyzing} can produce realistic and HQ face images. As shown in Fig.~\ref{GAN_framework} (b), the pre-trained GAN embedding architecture involves adding an additional pre-trained generative GAN~\cite{karras2019style,esser2021taming} into the generator network. For example, GPEN~\cite{yang2021gan} incorporates a pre-trained StyleGAN as a decoder within a U-network. It utilizes features extracted from the input by the decoder to refine the decoder's output, significantly improving restoration results compared to the general GAN structure. GFPGAN~\cite{wang2021towards} goes a step further by integrating features from various scales within the encoder through spatial transformations into a pre-trained GAN employed as a decoder. Other networks, such as GLEAN~\cite{chan2021glean}, Panini-Net~\cite{wang2022panini}, SGPN~\cite{zhu2022blind}, DEAR-GAN~\cite{hu2023dear}, DebiasSR~\cite{lianalyzing}, PDN~\cite{wang2023gan}, and others, also embrace this architecture. They incorporate a pre-trained StyleGAN or its variations into a GAN generator, complementing it with their individually crafted network architectures to cater to their specific application requirements. To further enhance the fidelity of the generated images, methods like VQFR~\cite{gu2022vqfr}, CodeFormer~\cite{zhou2022towards}, and others employ pre-train VQGAN to enhance facial details. They achieve this by employing discrete feature codesets extracted from HQ face images as prior. The discrete codebook prior, acquired within a smaller agent space, significantly reduces uncertainty and ambiguity compared to the continuous StyleGAN prior. 


Another category of blind methods focuses on addressing the challenge of obtaining paired LQ and HQ images in real-world scenarios. Inspired by CycleGAN~\cite{zhu2017unpaired}, as shown in Fig.~\ref{GAN_framework} (c), LRGAN~\cite{bulat2018learn} employs an cyclic GAN architecture consisting of two GAN networks. The initial high-to-low GAN generates LQ images that mimic real-world conditions and pairs them with corresponding HQ images. Subsequently, the second low-to-high GAN network is used to restore and enhance the quality of the generated LQ face images for restoration purposes. SCGAN~\cite{hou2023semi} takes a step further by guiding the generation of paired LQ images through the creation of degenerate branches from HQ images. This approach further reduces the domain gap between the generated LQ and the authentic LQ images. Additionally, diffusion-denoising techniques for blind tasks aim to improve robustness in severely degraded scenarios when compared to non-blind tasks. DR2~\cite{wang2023dr2} employs this technique to enhance the robustness of the blind restoration process and reduce artifacts often observed in the output face images. DDPM~\cite{wang2022zero} refines the spatial content during backpropagation to improve the robustness and realism of the restoration in challenging scenarios. DIFFBFR~\cite{qiu2023diffbfr} takes a different approach by initially restoring the LQ image and subsequently employing an LQ-independent unconditional diffusion model to refine the texture, rather than directly restoring the HQ image from a noisy input. 

\subsection{Joint Restoration Tasks}
In this section, we will discuss some essential components of FR, which include joint face completion and restoration, joint face frontalization, joint face alignment, joint face recognition, joint face illumination compensation, joint 3D face reconstruction, and joint face fairness. And we have shown representative methods for each of them in Fig.~\ref{joint_tasks_expamles}.

\noindent$\bullet$~\textbf{Joint Face Completion.} It is an important branch of FR, as real-world captured face images may suffer from both blurring and occlusion. One class of methods focuses on normal-resolution complements. MLGN~\cite{liu2019facial} and Swin-CasUNet~\cite{zeng2022swin} directly employ general networks for completion, but their fidelity is unsatisfactory. Given that accurately estimating occluded facial features is the key challenge in face completion, integrating prior information empowers models to infer critical details such as facial contours under occlusion. As a result, facial priors are extensively integrated into the majority of methods. For example, ID-GAN~\cite{ge2020occluded} uses facial identity, SwapInpaint~\cite{li2021swapinpaint} uses reference face, PFTANet~\cite{zhang2022pluralistic} employs face semantic labels, FT-TDR~\cite{wang2022ft} utilizes face landmarks, and others (~\cite{bai2023fine}) uses face components. Another class of methods focuses on low-resolution face completion, where the initial methods~\cite{yang2018hallucinating,cai2019fcsr,gao2023jdsr} address occluded parts through patching first before performing restoration work. However, this type of method can result in a significant accumulation of errors in the final results. In contrast, MFG-GAN~\cite{liu2023joint} utilizes graph convolution and customized loss functions to achieve end-to-end restoration. UR-GAN~\cite{zhang2022pro} utilizes landmarks guidance to progressively fix occluded and LQ faces.

\noindent$\bullet$~\textbf{Joint Face Frontalization.} Existing FR methods are primarily designed for frontal faces, and when applied to non-frontal faces, artifacts in the reconstructed results become evident. The first attempt to address this issue was made by TANN~\cite{yu2019can}. It utilized a discriminative network to enforce that the side-face generated face image should be close to the front-face image, aligning the faces in the same plane. Subsequently, VividGAN~\cite{zhang2021face} employed a fractalization network combined with a fine feature network to further optimize the face details under fractalization. MDFR~\cite{tu2021joint} introduced a 3D pose-based module to estimate the degree of face fractalization. It proposed a training strategy that integrates the recovery network with face fractalization end-to-end. Furthermore, inspired by the aforementioned methods, some approaches~\cite{li2020if,duan2021simultaneous} also combine the tasks of face completion and frontalization to address them jointly.

\begin{figure*}
\centering
\includegraphics[width=0.99\textwidth]{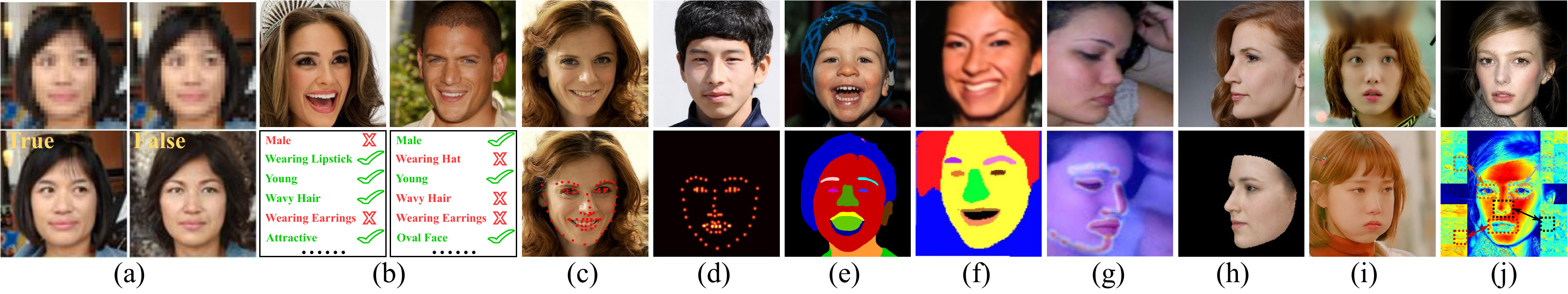} \vspace{-3mm}
\caption{\small Visualization of popular Facial priors. (a) Facial identity; (b) Facial attributes; (c) Facial landmarks; (d) Facial heatmaps; (e) Facial parsing map; (f) Facial semantic labels; (g) Facial components; (h) 3D Face; (i) Facial reference priors; (j) Facial dictionary.}
\label{priori_examples}
\vspace{-5mm}
\end{figure*}

\noindent$\bullet$~\textbf{Joint Face Alignment.} Most FR methods require the use of aligned face training samples for optimal performance. Therefore, researchers have developed various methods for joint face alignment. Yu \emph{et al.} were among the first to attempt embedding a spatial transformation layer as a generator and utilizing a discriminator to improve the alignment and upsampling. They developed TDN~\cite{yu2017face} and MDTN~\cite{yu2020hallucinating} using this approach. To handle possible noise in unaligned faces, they also developed a method~\cite{yu2017hallucinating} that incorporates downsampling and upsampling within the TDN framework to minimize the noise's impact. JASRNet~\cite{yin2020joint} achieves quality alignment in parallel by supervising facial landmarks and HQ face images. Another approach~\cite{abbasi2021identity} utilizes a face 3D dictionary alignment scheme to accomplish alignment.

\noindent$\bullet$~\textbf{Joint Face Recognition.} Some restoration methods~\cite{menon2020pulse,gu2020image} may result in recovered face images that diverge from their original identities, making them unsuitable for downstream face recognition tasks. 
Since face recognition heavily relies on local features such as the eyes, many priors struggle to accurately emphasize these specific areas. One swift solution involves applying a pre-trained face recognition model after the restoration. This helps determine whether the restored face image aligns with the ground truth in terms of identity, enhancing restoration accuracy by incorporating identity-related prior knowledge. Some examples of these methods include SICNN~\cite{zhang2018super}, LRFR~\cite{lai2019low}, 
 and others~\cite{ataer2019verification,mathai2019does,kim2021edge}. C-SRIP~\cite{grm2019face} improves upon this approach by recovering multiple scales of face images through different branches and supervising the recovered face images at different scales using a pre-trained face recognition network. Furthermore, some methods, including SiGAN~\cite{hsu2019sigan}, FH-GAN~\cite{bayramli2019fh}, WaSRGAN~\cite{huang2019wavelet}, and others, further enhance performance by incorporating discriminators into the restoration process. 

\noindent$\bullet$~\textbf{Joint Face Illumination Compensation.} Due to the unsatisfactory restoration performance of current algorithms on low-light LQ faces, this task has garnered significant attention. The main challenge in this task is detecting facial contours under low light conditions. As the first work, SeLENet~\cite{le2019selenet} segments the input low-light face into human face normals and light coefficients. It then augments the existing lighting coefficients to complete the lighting compensation process. CPGAN~\cite{zhang2020copy} employs an internal CPNet to accomplish detail restoration from the input facial image.  Additionally, it utilizes an external CPNet to compensate for background lighting using externally guided images. Furthermore, Zhang \emph{et al.}~\cite{zhang2021recursive} further improves CPGAN by introducing landmark constraints and recursive strategies. Ding \emph{et al.}~\cite{ding2020learning} employs a face localization network to detect facial landmarks, and then utilize these landmarks to better restore face contours and key features. Later, NASFE~\cite{yasarla2021network} introduces an automatic search strategy to discover an optimized network architecture specifically designed for the given task. 

\noindent$\bullet$~\textbf{Joint 3D Face Reconstruction.} With the advancement in 3D technology, there has been growing interest in achieving 3D face reconstruction from LQ face images or recovering reconstructed LR 3D faces. 
R3DPFH~\cite{zhong2020face} focused on predicting corresponding HQ 3D face meshes from LR faces containing noise. Utilizing the Lucas-Kanade algorithm, Qu \emph{et al.}~\cite{qu2017robust} aimed to improve the accuracy of 3D model fitting. Furthermore, Li \emph{et al.}~\cite{li20213d} and Uddin \emph{et al.}~\cite{uddin2022incomplete} utilized techniques for 3D point clouds to infer HR mesh data from LQ or incomplete 3D face point clouds. In contrast to the aforementioned methods, L2R~\cite{zhang2022learning} directly reconstructed HQ faces from LQ faces by learning to recover fine-grained 3D details on the proxy image.

\noindent$\bullet$~\textbf{Joint Face Fairness.} Existing datasets often fail to adequately represent the distribution of human races, which can introduce biases towards specific racial groups in trained methods. One class of approaches focuses on algorithmic fairness by employing suitable algorithms to mitigate racial bias. Ajil \emph{et al.}~\cite{jalal2021fairness} define theoretical concepts of race fairness and implement their defined notion of conditional proportional representation through a posteriori sampling, which helps achieve fairer face restoration. Noam \emph{et al.}~\cite{gat2021identity} enhance the feature extractor to better capture facial features, attributes, and racial information, by incorporating multifaceted constraints to reduce racial bias. Another class of approaches tackles the problem by building more ethnically balanced and comprehensive datasets. Zhang \emph{et al.}~\cite{zhang2022edface} developed the EDFace-Celeb-1M dataset, which covers 1.7 million photographs from different countries with relatively balanced ethnicity. Subsequently, Zhang \emph{et al.}~\cite{zhang2022blind} synthesized datasets for FR, namely EDFace-Celeb-1M and EDFace-Celeb-150K, which have made significant contributions to the progress of face fairness by providing more diverse and representative data.

\section{Face Prior Technology}~\label{PX}
Considering the inherent structured attributes of faces, many methods in the aforementioned tasks have chosen to incorporate facial priors to enhance restoration outcomes. To provide a better understanding of the diverse roles played by these priors in face restoration, this section focuses on exploring the technology of facial priors. We present these priors in Fig.~\ref{priori_examples} for reference. Based on whether they additionally utilize the structural information of the external face, we categorize these priors into two classes: internal proprietary prior-based methods and external compensatory prior-based methods. A summary of representative methods can be found in TABLE~\ref{tab:overview}. In the following sections, we will discuss these two classes of methods and their network structures in detail. 
It is worth noting that a few methods~\cite{yu2022multi,zhu2022blind} utilize both priors. 

\begin{table*}[htbp]
\setlength\tabcolsep{1pt}
 \renewcommand\arraystretch{0.95}
	\begin{center}
	 \caption{\small An overview of representative blind / non-blind and joint face restoration methods based on deep learning techniques.}
    \vspace{-1.5mm}
          \scalebox{0.96}{\begin{tabular}{r|c||c|c|c|c}
             \hline  
             \toprule
			 \cellcolor{lightgray}{Methods} & \cellcolor{lightgray}{Publication} & \cellcolor{lightgray}{Prior} & \cellcolor{lightgray}{Task}  & \cellcolor{lightgray}{Improved Technology} & \multicolumn{1}{c}{\cellcolor{lightgray}{Highlight}}    \\
			 \hline \hline 

			BCCNN~\cite{zhou2015learning}    & AAAI 2015 & \multirow{7}{*}{\tabincell{c}{Plain}} &\multirow{16}{*}{\tabincell{c}{Non-Blind \\ Task}} & Plain & \tabincell{c}{Bi-channel CNN}   \\ 
			
			ATMFN~\cite{jiang2019atmfn} & TMM 2019 & & & GAN+Attention & \tabincell{c}{Channel Attention Mechanism}   \\   
			
			SPARNet~\cite{chen2020learning}  & TIP 2020  & &  & Attention & \tabincell{c}{Spatial Attention Mechanism}     \\

            MSG-GAN~\cite{karnewar2020msg}  & CVPR 2020  & &  & GAN &Multi-Scale GAN \\

            
            Faceformer~\cite{wang2022faceformer}  & TCSVT 2022 &  &  & Attention & \tabincell{c}{Self-Attention Mechanism}  \\

            SCTANet~\cite{bao2023sctanet}  & TMM 2022 &  &  & Attention & \tabincell{c}{Self-attention / Spatial Attention}  \\

            IDM~\cite{gao2023implicit}  & CVPR 2023 &  & & Diffusion Model & \tabincell{c}{Diffusion Probabilistic Models}  \\ \cline{1-3}  \cline{5-6}

            LCGE~\cite{song2017learning} & IJCAI 2017  &\multirow{8}{*}{\tabincell{c}{Internal \\ Proprietary Prior}} &   & Prior & \tabincell{l}{Facial Components Prior} \\ 


            AEDN~\cite{yu2018super}  & CVPR 2018  &  & & GAN+Prior & \tabincell{l}{Given Facial Attribute Prior}  \\ 

            FSRNet~\cite{chen2018fsrnet}  & CVPR 2018  &  & & Prior & \tabincell{l}{Facial Landmarks / Parsing Maps Prior}  \\ 
                        
			FACN~\cite{xin2020facial} & AAAI 2020  & &   & Prior & \tabincell{l}{Estimated Facial Attribute Prior} \\ 
            
            DIC~\cite{ma2020deep} & CVPR 2020  &  & & Attention+Prior & \tabincell{l}{Facial Landmarks / Components Prior}  \\ 

            PAP3D~\cite{hu2021face} & TPAMI 2021  &  & & Attention+Prior & \tabincell{l}{3D Facial Prior / Spaital Attention}  \\ 
           
            HCRF~\cite{liu2021features}   & TIP 2021  &  & & Prior & Facial Semantic Labels Prior \\ \cline{1-3}  \cline{5-6}

			GWAInet~\cite{dogan2019exemplar}  & CVPR 2019  &\multirow{2}{*}{\tabincell{c}{External \\ Compensatory Prior}}  & & GAN+Prior & High Quality Images As Reference Prior \\ 

            KDFSRNet~\cite{wang2022propagating}   & TCSVT 2022  &  & & Prior & Pre-trained Teacher's Knowledge As Generative Prior \\
            \hline
            \hline

            LRGAN~\cite{bulat2018learn}      & ECCV 2018  & \multirow{6}{*}{\tabincell{c}{Plain}} & \multirow{22}{*}{\tabincell{c}{Blind Task}}  & GAN & Unsupervised / Two-Stage GAN  \\ 
      
			HiFaceGAN~\cite{yang2020hifacegan}  & MM 2020  &  &  & GAN &  Semantic-Guided Generation \\

            GCFSR~\cite{he2022gcfsr}  & CVPR 2022  &  &    & GAN &  Generative And Controllable Framework \\

            SCGAN~\cite{hou2023semi}  & TIP 2023  &  &    & GAN & Unsupervised / Semi-Cycled GAN \\

            DR2~\cite{wang2023dr2}  & CVPR 2023  &  &    & Diffusion Model & Diffusion-based Robust Degradation Remover \\

            IDDM~\cite{zhao2023towards}  & ICCV 2023  &  &    & Diffusion Model & Iteratively Learned System \\
			\cline{1-3}  \cline{5-6}

			Super-FAN~\cite{bulat2018super} & CVPR 2018  &\multirow{5}{*}{\tabincell{c}{Internal \\ Proprietary Prior}} &   & GAN+Prior & \tabincell{l}{Facial Heatmap Prior} \\ 

            MDCN~\cite{shen2020exploiting}  & IJCV 2020     & &  & GAN+Prior & Semantic labels Prior\\ 

            UMSN~\cite{yasarla2020deblurring}  & TIP 2020     & &  & GAN+Prior & Facial Components Prior\\ 
   
			PSFRGAN~\cite{chen2021progressive}  & CVPR 2021     & &  & GAN+Prior & Facial Parsing Maps Prior\\ 
		    
		    SGPN~\cite{zhu2022blind}  & CVPR 2022     & &  & GAN+Prior & 3D Faical / Pre-trained Generative Prior\\ \cline{1-3}  \cline{5-6}

			GFRNet~\cite{li2018learning} & ECCV 2018   & \multirow{11}{*}{\tabincell{c}{External \\Proprietary Prior}} &  & GAN+Prior & High Quality Images As Reference Prior   \\

            PULSE~\cite{menon2020pulse}  & CVPR 2020     & &  & GAN+Prior & Pre-trained StyleGAN's Generative Prior \\ 

            ASFFNet~\cite{li2020enhanced}  & CVPR 2020     & &  & GAN+Attention+Prior & Reference / Landmark Prior\\ 

            DFDNet~\cite{li2020blind}  & ECCV 2020     & &  & GAN+Prior & Faical Component Dictionaries Prior\\ 
			
			GFPGAN~\cite{wang2021towards} & CVPR 2021  & &  & GAN+Prior & Pre-trained StyleGAN's Generative Prior\\


            DMDNet~\cite{li2023learning} & TPAMI 2022  & &  & GAN+Prior & Faical Component Dictionaries Prior / Reference Prior\\

            RestoreFormer~\cite{wang2022restoreformer} & CVPR 2022  & &  & GAN+Attention+Prior & Faical Component Dictionaries\\

            VQFR~\cite{gu2022vqfr} & ECCV 2022  & &  & GAN+Prior & Pre-trained VQGAN's Codebook Prior\\

            CodeFormer~\cite{zhou2022towards} & NIPS 2022  & &  & GAN+Prior & Pre-trained VQGAN's Codebook Prior\\

            DebiasFR~\cite{lianalyzing} & IJCAI 2023  & &  & GAN+Prior & Pre-trained StyleGAN's Generative Prior
            \\  \hline \hline

			TDN~\cite{yu2017face} & AAAI 2017 & \multirow{4}{*}{\tabincell{c}{Plain}} & \multirow{16}{*}{\tabincell{c}{Joint Task}}  & GAN & CNN / Joint Face Alignment  \\ 
   
		    TANN~\cite{yu2019can} & TPAMI 2019     & &  & GAN & CNN / Joint Face Frontalization  \\

            MTDN~\cite{yu2020hallucinating} & IJCV 2020     & &  & GAN & CNN / Joint Face Alignment  \\
           	
            EDFace-Celeb-1M~\cite{zhang2022edface} & TPAMI 2022 &  & & Plain & Dataset / Joint Face Fairness  \\ \cline{1-3}  \cline{5-6}

            SICNN~\cite{zhang2018super} & ECCV 2018 & \multirow{10}{*}{\tabincell{c}{Internal
            \\ Proprietary Prior}} &  & Prior & Identity Prior / Joint Face Recognition  \\ 

            FCSR-GAN~\cite{cai2019fcsr} & TBIOM 2020   &  & & GAN+Prior & Landmark/ Semantic labels / Joint Face Compensation  \\
   
		    JASRNet~\cite{yin2020joint} & AAAI 2020     & &  & Prior & Landmark Prior / Joint Face Alignment \\ 

            ID-GAN~\cite{yin2020joint} & TCSVT 2020     & &  & GAN+Prior & Semantic labels / Identity Prior / Joint Face Recognition \\ 

            SiGAN~\cite{hsu2019sigan} & TIP 2019     & &  & GAN+Prior & Identity Prior / Joint Face Recognition \\ 

            MDFR~\cite{tu2021joint} & TCSVT 2021   &  & & GAN+Prior & Landmark/ 3D Facial Prior / Joint Face Frontalization  \\
           	
            
            FT-TDR~\cite{wang2022ft} & TMM 2022   &  & & GAN+Attention+Prior &  Landmark Prior/ Self-Attention / Joint Face Completion  \\
            
            L2R~\cite{zhang2022learning} & CVPR 2022   &  & & GAN+Prior &  Generative / 3D Prior / Joint 3D Face Reconstruction  \\

            FVIP~\cite{yang2023deep} & TIP 2023   &  & & GAN+Prior &  3D Face Prior / Joint Face Completion  \\
            \cline{1-3}  \cline{5-6}

		    CPGAN~\cite{zhang2020copy} & CVPR 2020   & \multirow{3}{*}{\tabincell{c}{External
            \\ Proprietary Prior}} &  & GAN+Attention+Prior & Reference Prior / Joint Illumination Compensation \\ 
           	
            ViVidGAN~\cite{zhang2021face} & TIP 2021 &  & & GAN+Attention+Prior & Reference Prior / Joint Illumination Compensation  \\ 

            IAPTU~\cite{zhang2021face} & ICIP 2021 &  & & GAN+Prior & Pre-trained Generative Prior / Joint Face Fairness  \\ 
            
            \bottomrule
  		\end{tabular}}     
		\label{tab:overview}
	\end{center}
    \vspace{-5mm}
\end{table*}

\subsection{Internal Proprietary Prior}
This type of method primarily utilizes knowledge about the attributes and structural features inherent to the face itself. It incorporates information such as identity, facial features, and contours to guide the face restoration process. Common techniques employed in this approach include identity recognition, facial landmarks creation, semantic labeling maps, and more.

The first type of information used is the face's own 1D information, such as identity prior and attribute prior. Identity prior refers to information related to an individual's identity, indicating whether the restored face corresponds to the same person as the ground truth. Integrating identity prior to the restoration process enhances the model's ability to faithfully recover facial features. Methods based on identity prior, such as SICNN~\cite{zhang2018super}, FH-GAN~\cite{bayramli2019fh}, IPFH~\cite{cheng2019identity}, C-SRIP~\cite{grm2019face}, and others, aim to maintain identity consistency between the restored image and the HQ face image. During training, these frameworks typically include a restoration network and a pre-trained face recognition network. The face recognition network serves as an identity prior, determining whether the restored face belongs to the same identity as the HQ face, thereby improving the identity accuracy of the restored face. The face attribute prior provides 1D semantic information about the face for face restoration, such as attributes like long hair, age, and more. This prior aids the model in understanding and preserving specific facial characteristics during the restoration process. 
For instance, incorporating age attributes into the restoration process assists models in accurately preserving natural textures such as skin wrinkles. Earlier methods, such as EFSRSA~\cite{yu2018super}, ATNet~\cite{li2019deep}, ATSENet~\cite{li2020learning}, AACNN~\cite{lee2018attribute}, and others, directly connect the attribute information to the LQ image or its extracted features. Other methods, like AGCycleGAN~\cite{lu2018attribute} and FSRSA~\cite{yu2018super}, use a discriminator to encourage the network to pay more attention to attribute features during restoration. However, these methods may experience significant performance degradation when attributes are missing. To address this issue, attribute estimation methods~\cite{xin2019residual,xin2020facial} have been proposed. These approaches design appropriate attribute-based losses that enable the network to adaptively predict attribute information. RAAN~\cite{xin2019residual} utilizes three branches to separately predict face shape, texture, and attribute information. It emphasizes either face shape or texture based on the attribute channel. FACN~\cite{xin2020facial} introduces the concept of capsules to enhance the recovered face. This is achieved by performing multiplication or addition operations between the face attribute mask estimated by the network and the semantic or probabilistic capsule obtained from the input. 


Another class of methods emphasizes the use of the face's unique 2D geometric or 3D spatial information as priors. Facial landmarks~\cite{kim2019progressive,ma2020deep,wang2022ft} and facial heatmaps~\cite{wang2021heatmap,wang2022ft,yu2022multi} are examples of these priors, representing coordinate points or probability density maps that indicate key facial components such as the eyes, nose, mouth, and chin. They provide accurate and detailed facial location information. Methods like DIC~\cite{ma2020deep} utilize the predicted coordinates of facial landmarks from the prior estimation network to guide the restoration network. However, using a large number of facial landmarks may lead to error accumulation in coordinate estimation, particularly for severely degraded face images, resulting in distortion of the restored facial structure. In contrast, facial parsing maps~\cite{chen2018fsrnet,chen2021progressive,yu2021semantic} and facial semantic labels~\cite{shen2018deep,yasarla2020deblurring,shen2020exploiting} are more robust to severe degradation as they segment the face into regions. Even if some regions are severely degraded, intact regions can still guide the restoration process. Moreover, these priors contain more comprehensive facial information, enabling the restoration model to better understand the overall facial structure and proportions, leading to more coherent restorations. However, these priors may involve multiple semantic labels for different facial regions, requiring more complex networks~\cite{yasarla2020deblurring,chen2021progressive} to address semantic ambiguity. On the other hand, facial components~\cite{yu2018face,lu2022rethinking,bai2023fine} provide a straightforward representation of critical facial features, reducing the need for complex models while effectively guiding the restoration process. In addition to the aforementioned 2D facial priors, Hu \emph{et al.}~\cite{hu2020face} introduced the use of a 3D face prior to handle faces with large pose variations. Subsequent 3D prior-based methods~\cite{zhu2022blind,dey20223dfacefill,yang2023deep} demonstrated their robustness in handling complex facial structures and significant pose changes. There are also methods~\cite{li2020learning,zhang2022pluralistic} that strive to achieve more comprehensive restoration by synergistically combining multiple internal proprietary priors.

\subsection{External Compensatory Prior}
Methods that leverage external priors primarily rely on externally guided faces or information sources derived from external HQ face datasets to facilitate the face restoration process. These external priors can take various forms, including reference priors, face dictionary priors, and pre-trained generative priors.

Reference prior-based methods~\cite{li2018learning,li2020enhanced,dogan2019exemplar} utilize HQ face images of the same individual as a reference to enhance the restoration of a target face image. The challenge lies in effectively handling reference faces with varying poses and lighting conditions. GFRNet~\cite{li2018learning} is the pioneering work in this field. It employs a sub-network called WarpNet, coupled with a landmark loss, to rectify pose and expression disparities present in the reference face. This enables the model to effectively utilize reference faces that exhibit differences compared to the face undergoing restoration. GWAInet~\cite{dogan2019exemplar} utilizes the structure of the generative network of the GAN and achieves favorable results without relying on facial landmarks. Subsequently, ASFFNet~\cite{li2020enhanced} further enhances performance by refining the selection of the guide face and improving the efficiency of feature fusion between the guide face and the image to be recovered. 

However, the above methods require reference images for both training and inference, which limits their applicability in various scenarios. To address this limitation, DFDNet~\cite{li2020blind} employs a strategy that creates a facial component dictionary. Initially, a dictionary comprising facial elements such as eyes, nose, and mouth is categorized from an HQ face dataset. During the training phase, the network dynamically selects the most analogous features from the component dictionary to guide the reconstruction of corresponding facial parts. RestoreFormer~\cite{wang2022restoreformer} integrates Transformer architecture and leverages the face component loss to more effectively utilize the potential of the facial component dictionary. DMDNet~\cite{li2023learning} leverages external facial images as well as other images of the same individual to construct two distinct facial dictionaries. This process enables a gradual refinement from the external dictionary to the personalized dictionary, resulting in a coarse-to-fine bootstrapping approach.


\begin{figure*}[t!]
\setlength\tabcolsep{2pt}
\begin{minipage}{0.325\linewidth}
\begin{subfigure}[h]{\linewidth}
\includegraphics[width=0.99\linewidth]{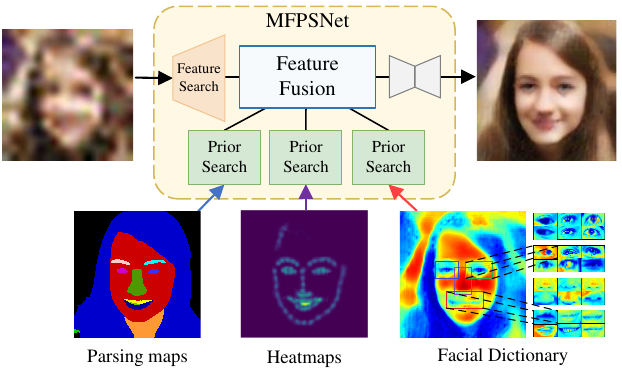}
\centering
\end{subfigure}
\vspace{-1.7mm}
\captionof{figure}{\small The representative method MFPSNet~\cite{yu2022multi} for enhancing restoration process using multiple priors.}
\label{fig:MFPSNet}
\end{minipage}
~
\begin{minipage}{0.33\linewidth}
\begin{subfigure}[h]{\linewidth}
\begin{tabular}{cccc}
\includegraphics[width=0.27\linewidth]{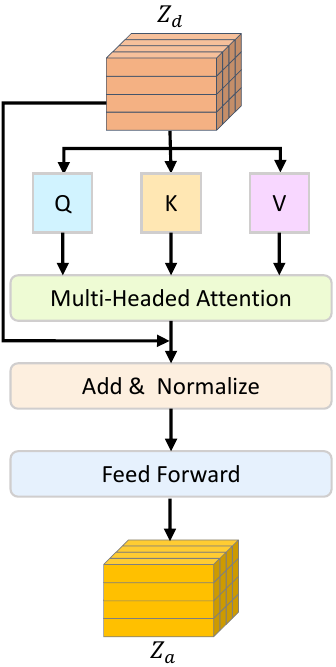} & \includegraphics[width=0.27\linewidth]{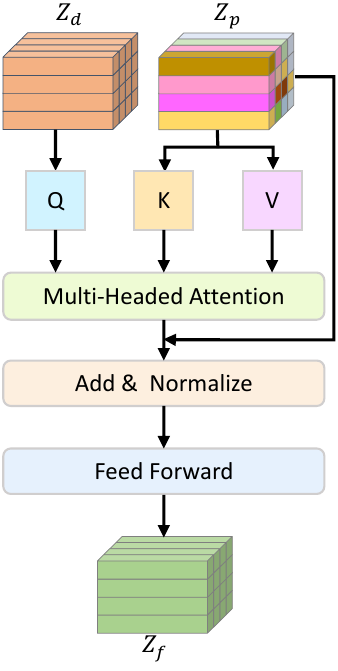} & \multicolumn{2}{c}{\includegraphics[width=0.37\linewidth]{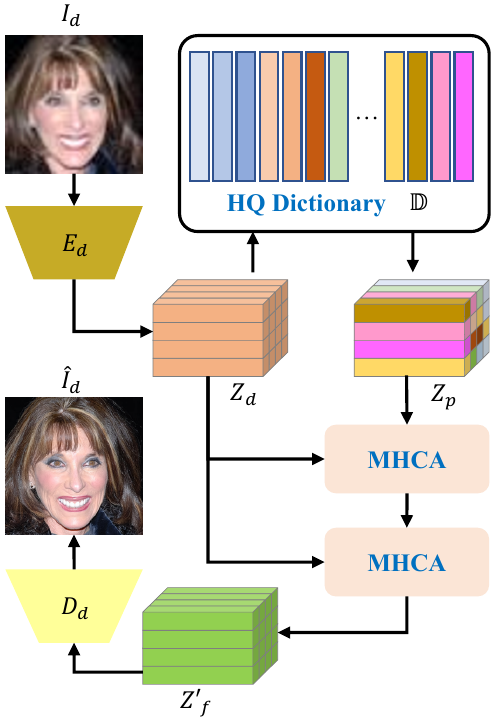}} 
\vspace{-1.5mm}\\
\tiny(a) MHSA & \tiny(b) MHCA & \multicolumn{2}{c}{\tiny(c) RestoreFormer} \\
\centering
\end{tabular}
\end{subfigure}
\vspace{-7mm}
\captionof{figure}{\small The representative method RestoreFormer~\cite{wang2022restoreformer} for designing networks that more efficiently utilize priors.}
\label{fig:Restoreformer}
\end{minipage}
~
\begin{minipage}{0.315\linewidth}
\vspace{3.5mm}
\begin{subfigure}[h]{\linewidth}
\includegraphics[width=0.99\linewidth]{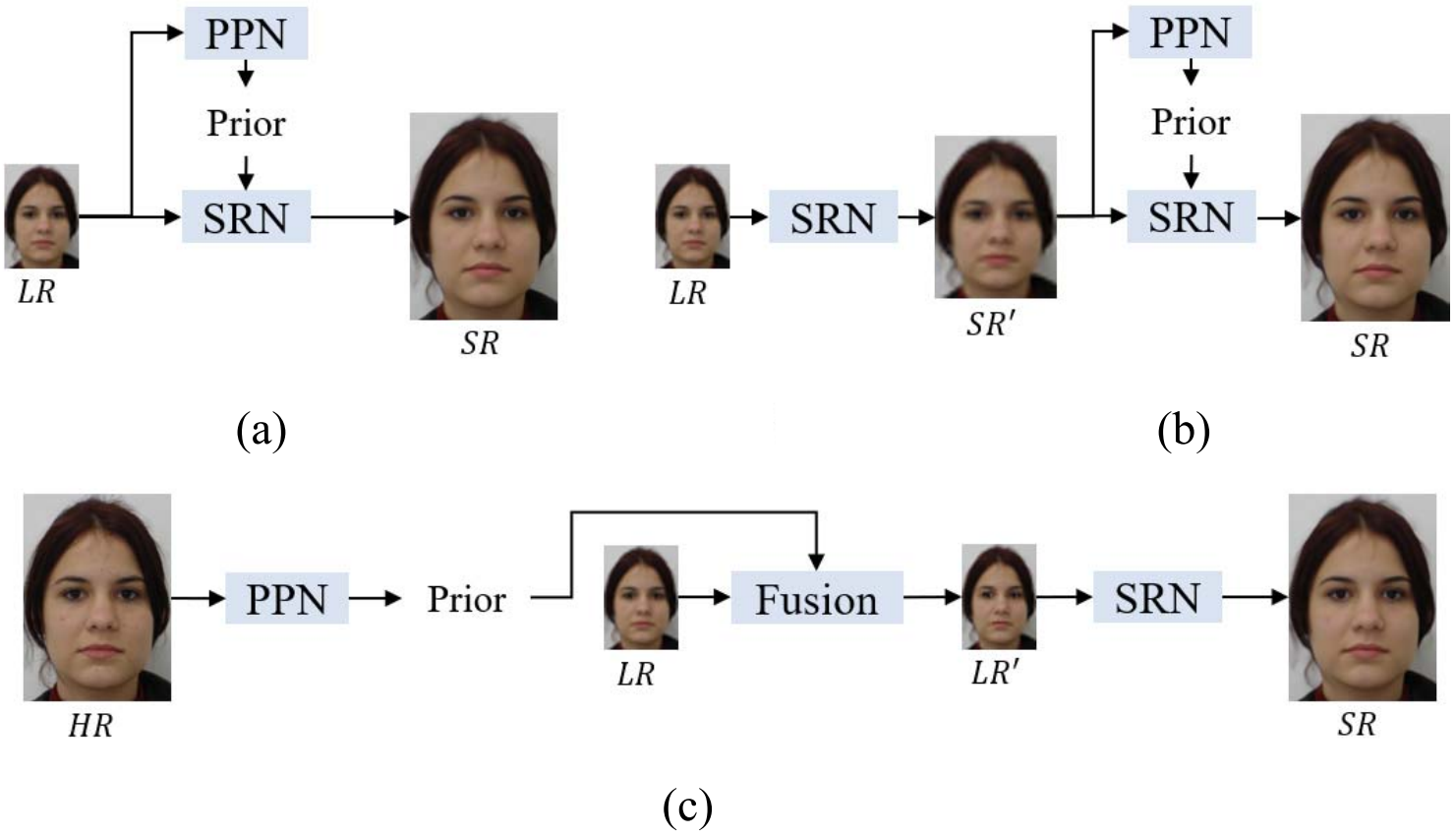}
\centering
\end{subfigure}
\vspace{-2.5mm}
\captionof{figure}{\small The method CHNet~\cite{lu2022rethinking} for enhancing the effectiveness of priors by changing the prior guidance approach.}
\label{fig:CHNet}
\end{minipage}
\vspace{-4.5mm}
\end{figure*}

Unlike face dictionary that requires manual separation of facial features, pre-trained face GAN models~\cite{karras2019style,karras2020analyzing,esser2021taming} can automatically extract information beyond facial features, including texture, hair details, and more. This makes approaches based on pre-trained generative priors simpler and more efficient. PULSE~\cite{menon2020pulse} is a pioneering breakthrough in FR that utilizes generative prior. It identifies the most relevant potential vectors in the pre-trained GAN feature domain for the input LQ face. Subsequently, mGANprior~\cite{gu2020image} enhances the PULSE method by incorporating multiple potential spatial vectors derived from the pre-trained GAN. However, these methods are complex and may struggle to ensure fidelity in restoration while effectively leveraging the input facial features. Approaches like GLEAN~\cite{chan2022glean}, GPEN~\cite{yang2021gan}, and GFPGAN~\cite{wang2021towards} integrate a pre-trained GAN into their customized networks. They employ GAN's generative prior to guide the forward process of the network, effectively leveraging the input facial features and leading to improved fidelity in restoration. Subsequent techniques~\cite{wang2022panini,hou2022feature,hu2023dear,wang2023gan,lianalyzing} aim to enhance the efficacy of pre-trained GAN priors by investigating optimal strategies for integrating pre-trained GANs with forward networks or exploring more efficient forward networks. SGPN~\cite{zhu2022blind} incorporates a 3D shape prior along with the generative prior to enhance restoration, combining both spatial and structural information. Apart from approaches based on pre-trained StyleGAN~\cite{karras2019style,karras2020analyzing}, there is another category of methods built upon pre-trained VQGAN~\cite{esser2021taming}. The key advantage of VQGAN lies in its utilization of a vector quantization mechanism, enabling accurate manipulation of specific features within the generated face images. Additionally, the training of it is more stable compared to some variants of StyleGAN. VQFR~\cite{gu2022vqfr} leverages discrete codebook vectors from VQGAN, using optimally sized compression patches and a parallel decoder to improve detail and fidelity in the restored outcomes. Codeformer~\cite{zhou2022towards} integrates Transformer technology into its network architecture, achieving a favorable trade-off between quality and fidelity with a controlled feature conversion module. Zhao \emph{et al.}~\cite{zhao2022rethinking} explores the utilization of pre-training priors, aiming to strike a harmonious equilibrium between generation and restoration aspects.

\subsection{Advancing the Effectiveness of Prior}
In this section, we will delve into approaches aimed at enhancing the effectiveness of prior knowledge for facial restoration. These approaches include combining multiple priors, developing efficient network structures, and adopting the prior guide approach. 

\noindent$\bullet$~\textbf{Combining Multiple Priors.} Since different prior are suitable to different scenarios, the effectiveness of prior utilization diminishes significantly when inappropriate priors are used. To address this issue, some methods enhance the effectiveness of individual prior in facial restoration by incorporating multiple priors during the restoration process, leveraging the flexible complementarity of various prior information. Fig.~\ref{fig:MFPSNet} illustrates MFPSNet~\cite{yu2022multi}, which utilizes multiple priors including face parsing maps, face landmarks, and face dictionary to assist in restoration. Compared to approaches relying on a single prior, MFPSNet exhibits better robustness in highly blurry scenes. In general, some methods~\cite{yasarla2020deblurring,li2023learning,chen2018fsrnet,li2020learning} make use of either multiple internal proprietary priors or multiple external compensating priors. For example, UMSN~\cite{yasarla2020deblurring} employs both face semantic labels and facial components as priors. DMDNet~\cite{li2023learning} utilizes both facial dictionaries and external reference faces. Additionally, some methods~\cite{wang2021towards,zhu2022blind,wang2022restoreformer} combine internal proprietary priors with external compensating priors. For instance, SGPN~\cite{zhu2022blind} leverages a 3D face shape prior alongside a pre-trained GAN prior. However, employing an approach that utilizes multiple priors requires increased computational resources for prior estimation and often demands a larger dataset for modeling. 

\noindent$\bullet$~\textbf{Efficient Network Structures.} Initial methods~\cite{chen2018fsrnet,ma2020deep} primarily focused on utilizing simple residual block structures for prior fusion, although these structures were not always optimal solutions. Subsequently, some methods~\cite{li2020enhanced,wang2021towards,gu2022vqfr,zhou2022towards} aimed to design more efficient networks for prior fusion or estimation to enhance restoration performance. As depicted in Fig.~\ref{fig:Restoreformer}, RestoreFormer~\cite{wang2022restoreformer} designs a custom multi-head cross-attention mechanism (MHCA) to comprehensively integrate facial dictionary information with facial features, showcasing significantly superior performance compared to multi-head self-attention (MHSA) alone. Similarly, ASFFNet~\cite{li2020enhanced} enhances the fusion of prior information with facial semantic features through a specially crafted adaptive spatial feature fusion block. VQFR~\cite{gu2022vqfr} employs a parallel decoder structure to blend the generated prior information with low-level features, ensuring enhanced fidelity without compromising the quality of the prior guidance.

\noindent$\bullet$~\textbf{Prior Guide.} The way the prior is bootstrapped plays a crucial role in determining its effectiveness, as different bootstrapping methods yield varying restoration outcomes. For example, PFSRGAN~\cite{chen2021progressive} aims to enable the model to more effectively leverage the raw input information by directly estimating the prior knowledge from the LQ facial images to guide the restoration. In contrast, FSRNet~\cite{chen2018fsrnet} partially restores the LQ faces before estimating the prior to address inaccuracies in prior knowledge estimation. JASRNet~\cite{yin2020joint} adopts a bootstrapping structure with parallel communication to fully leverage the interaction between prior estimation and restoration. Furthermore, as illustrated in Fig.~\ref{fig:CHNet}, CHNet~\cite{lu2022rethinking} modifies the process of estimating priors by opting to estimate them from HQ faces instead of directly or indirectly from LQ faces. For more comprehensive generalizations regarding the prior guide approach, please refer to the provided \emph{supplementary material}. 


\begin{table*}[t!]
\setlength\tabcolsep{2pt}
\begin{minipage}{0.5\linewidth}
\centering
\vspace{-1pt}
\resizebox{1\textwidth}{!}{
\begin{tabular}{l||cccc|cccc}
\hline
\toprule
\rowcolor{lightgray}
& \multicolumn{4}{c|}{CelebA} 
& \multicolumn{4}{c}{Helen} 
\\ 
\cmidrule{2-9}
    \rowcolor{lightgray}
    \multicolumn{1}{l||}{\multirow{-2}{*}{Methods ($\times 8$)}} 
    & PSNR$\uparrow$  & SSIM$\uparrow$  & LPIPS$\downarrow$  & FID$\downarrow$   
    & PSNR$\uparrow$  & SSIM$\uparrow$  & LPIPS$\downarrow$  & FID$\downarrow$  
    \\ 
    \hline\hline
    RCAN~\cite{zhang2018image}      & 27.45   & 0.7824   & 0.2205  & 174.0  & 25.50  &  0.7383  & 0.3437  & 219.3 \\ 
    
    FSRNet~\cite{chen2018fsrnet}    & 27.05   & 0.7714   & 0.2127  & 170.4  & 25.45  & 0.7364  & 0.3090  & 228.8 \\
    
    FACN~\cite{xin2020facial}       & 27.22   & 0.7802   & 0.1828  & 167.7  & 25.06  & 0.7189  & 0.3113  & 218.0 \\
    
    SPARNet~\cite{chen2020learning} & 27.73   & 0.7949   & 0.1995  & 161.2  & 26.43  & 0.7839  & 0.2674  & 211.5 \\

    DIC~\cite{ma2020deep}           & -       & -       & -      & -      & 26.15  & 0.7717  & 0.2158  & 214.1 \\
    
    SISN~\cite{lu2021face}          & 27.91   & 0.7971   & 0.2005  & 162.3  & 26.64  & 0.7908  & 0.2571  & 210.7 \\

    SwinIR~\cite{liang2021swinir} & 27.88   & 0.7967   & 0.2001  & 163.2  & 26.53  & 0.7856  & 0.2644  & 213.2 \\
    
    CTCNet~\cite{gao2023ctcnet}     & \multicolumn{1}{>{\columncolor{lightgray}}c}{ \bf{28.37}}   & \multicolumn{1}{>{\columncolor{lightgray}}c}{ \bf{0.8115}}   & \multicolumn{1}{>{\columncolor{lightgray}}c}{ \bf{0.1702}}   & 156.9  & \multicolumn{1}{>{\columncolor{lightgray}}c}{ \bf{27.08}}  & 0.8007  & 0.2094  & 205.8 \\

    SCTANet~\cite{bao2023sctanet}   & \cellcolor{tinygray}\underline{28.26}   & \cellcolor{tinygray}\underline{0.8100}    & \cellcolor{tinygray}\underline{0.1710}      & \cellcolor{tinygray}\underline{156.8}  & \cellcolor{tinygray}\underline{27.01}   & \multicolumn{1}{>{\columncolor{lightgray}}c}{ \bf{0.8068}}  & \cellcolor{tinygray}\underline{0.1901}  & \cellcolor{tinygray}\underline{203.3} \\

    SFMNet~\cite{wang2023spatial}   & 27.85   & 0.7967   & 0.1837  & \cellcolor{lightgray}\bf{156.5}  & 26.98  & \cellcolor{tinygray}\underline{0.8049}  & \multicolumn{1}{>{\columncolor{lightgray}}c}{ \bf{0.1865}}  & \multicolumn{1}{>{\columncolor{lightgray}}c}{ \bf{199.5}} \\    
    \hline\hline
    Input                           & 23.61   & 0.6779   & 0.4899  & 362.2
    & 22.95   & 0.6762  & 0.4912  & 289.1
    \\ 
    \bottomrule
\end{tabular}}
\vspace{-1mm}
\captionof{table}{\small Performance comparison of key non-blind methods on CelebA and Helen Test Sets. In this paper, the best and the second best values are highlighted and underlined respectively.}
\label{tab:non_blind_performance}
\end{minipage}
~
\begin{minipage}{0.48\linewidth}
\begin{subfigure}[h]{\linewidth}
\begin{overpic}[width=\linewidth]{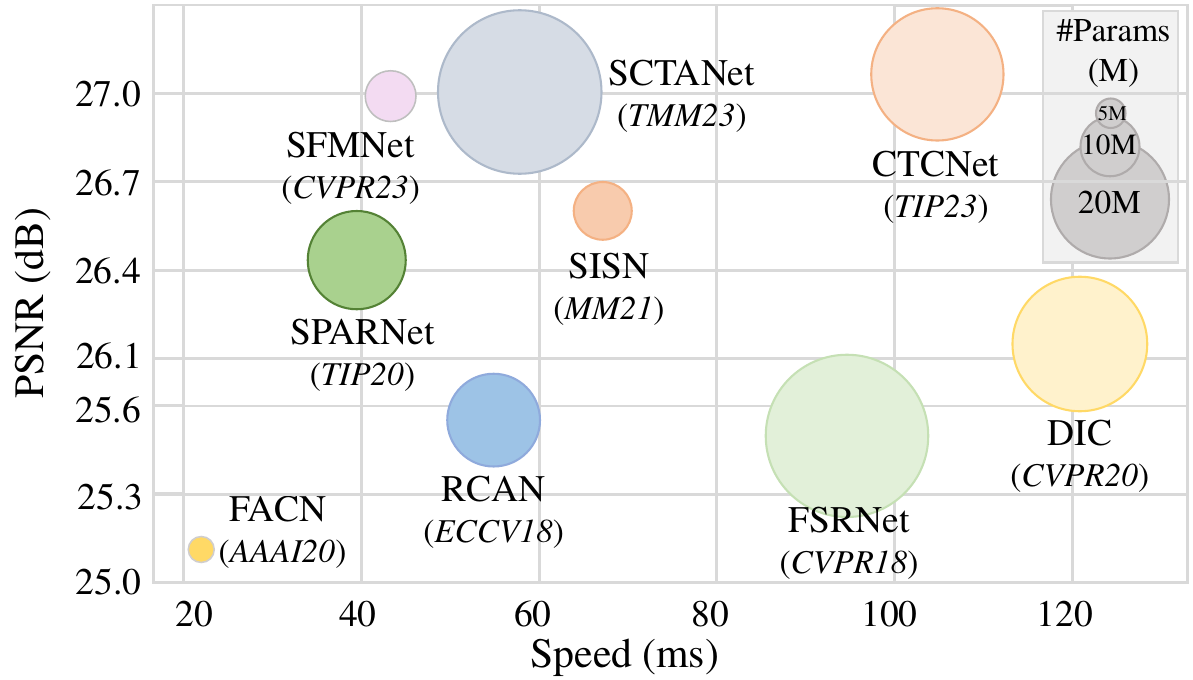}
\put(45, 14.5)
{\scriptsize\linethickness{0.3mm}~\cite{zhang2018image}}
\put(75.5, 12)
{\scriptsize\linethickness{0.3mm}~\cite{chen2018fsrnet}}
\put(35.5, 28){\scriptsize\linethickness{0.3mm}~\cite{chen2020learning}}
\put(26.2, 13.2)
{\scriptsize\linethickness{0.3mm}~\cite{xin2020facial}}
\put(87.5, 13.5)
{\scriptsize\linethickness{0.3mm}~\cite{ma2020deep}}
\put(53.5, 34)
{\scriptsize\linethickness{0.3mm}~\cite{lu2021face}}
\put(75, 36)
{\scriptsize\linethickness{0.3mm}~\cite{gao2023ctcnet}}
\put(62.5, 49)
{\scriptsize\linethickness{0.3mm}~\cite{bao2023sctanet}}
\put(34.1, 42){\scriptsize\linethickness{0.3mm}~\cite{wang2023spatial}}
\end{overpic}
\centering
\end{subfigure}
\vspace{-2mm}
\captionof{figure}{\small Complexity analysis of non-blind methods on Helen.}
\label{fig:non_blind_time_psnr}
\end{minipage}
\vspace{-2.5mm}
\end{table*}

\begin{figure*}
\begin{center}
\begin{overpic}[width=1\linewidth]{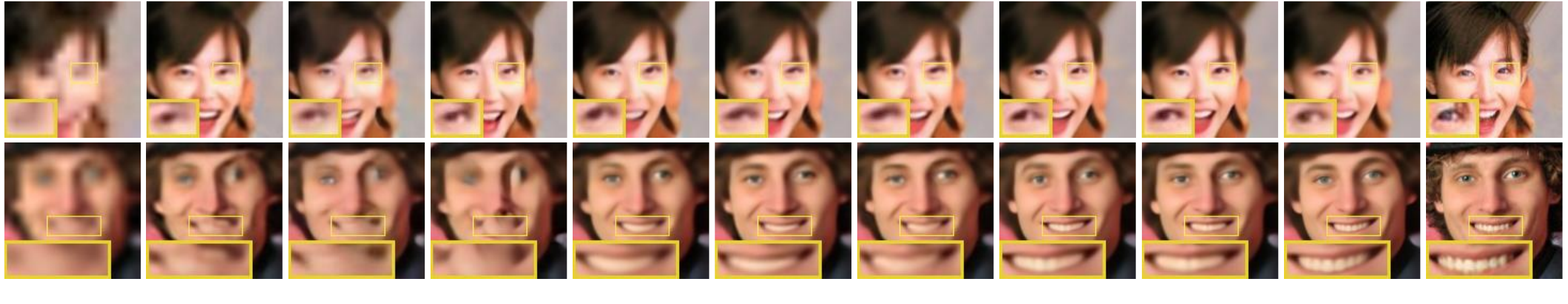}
\put(0.15,0.35){\color{black}{\fontsize{6.8pt}{1pt}\selectfont Synthetic Input}}
\put(10.2,0.35){\color{black}{\scriptsize RCAN~\cite{zhang2018image}}}
\put(19.4,0.35){\color{black}{\scriptsize FSRNet~\cite{chen2018fsrnet}}}
\put(28.2,0.35){\color{black}{\scriptsize FACN~\cite{xin2020facial}}}
\put(36.9,0.35){\color{black}{\scriptsize SPARNet~\cite{chen2020learning}}}
\put(46.9,0.35){\color{black}{\scriptsize SISN~\cite{lu2021face}}}
\put(55.5,0.35){\color{black}{\scriptsize SwinIR~\cite{liang2021swinir}}}
\put(64.25,0.35){\color{black}{\scriptsize CTCNet~\cite{gao2023ctcnet}}}
\put(73.1,0.35){\color{black}{\scriptsize SCTANet~\cite{bao2023sctanet}}}
\put(82.35,0.35){\color{black}{\scriptsize SFMNet~\cite{wang2023spatial}}}
\put(91.4,0.35){\color{black}{\scriptsize Grouth-truth}}
\end{overpic}
\end{center}
\vspace{-4mm}
\caption{\small Visual comparison of different non-blind methods on the CelebA (first row) test set and Helen (second row) test set.}
\vspace{-1mm}
\label{non_blind_visual}
\end{figure*}
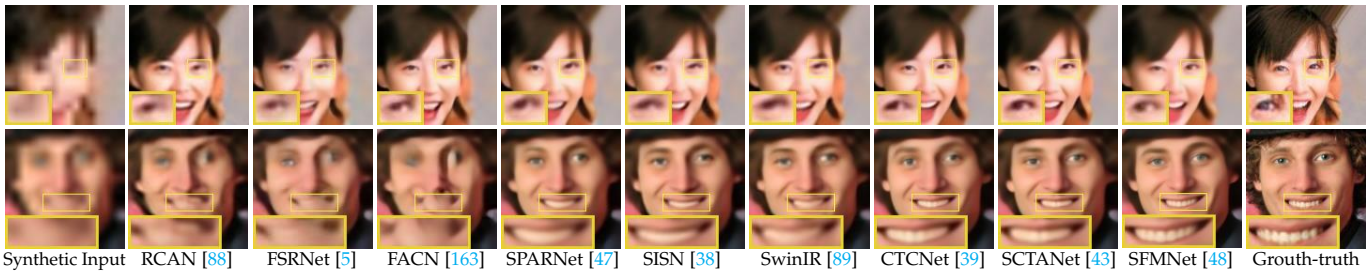

\begin{table*}[t]
\setlength\tabcolsep{2pt}
\begin{minipage}{0.45\linewidth}
  \centering 
    \scalebox{0.99}{
    \begin{tabular}{c||c|c|c|c}
    \toprule
    \rowcolor{lightgray} Method  & RCAN~\cite{zhang2018image}  & FSRNet~\cite{chen2018fsrnet}  & FACN~\cite{xin2020facial}  & SPARNet~\cite{chen2020learning} \\
    \hline 
    \hline
    Params  & 15.7M  & 27.5M  & 4.4M   & 16.6M  \\
    MACs    & 4.7G   & 40.7G  & 12.5M  & 7.1G   \\
    Speed   & 56ms   & 89ms   & 22ms   & 40ms   \\
    \hline
    \hline
    DIC~\cite{ma2020deep}  & SISN~\cite{lu2021face}  & CTCNet~\cite{gao2023ctcnet}  & SCTANet~\cite{bao2023sctanet} & SFMNet~\cite{wang2023spatial}  \\
    \hline 
    22.8M   & 9.8M   & 22.4M   & 27.7M  & 8.6M  \\
    35.5G   & 2.3G   & 47.2G   & 10.4G  & 30.6G \\
    122ms   & 68ms   & 106ms   & 58ms   & 48ms  \\
    \bottomrule
    \end{tabular}}
    \vspace{-1mm}
    \caption{\small Speed and overhead comparison of typical non-blind methods that measured on $128 \times 128$ images. We test all models using an NVIDIA RTX 3090 GPU.}
    \label{tab:non_blind_speed}
\end{minipage}
~
\begin{minipage}{0.52\linewidth}
\begin{subfigure}[h]{\linewidth}
\begin{overpic}[width=\linewidth]{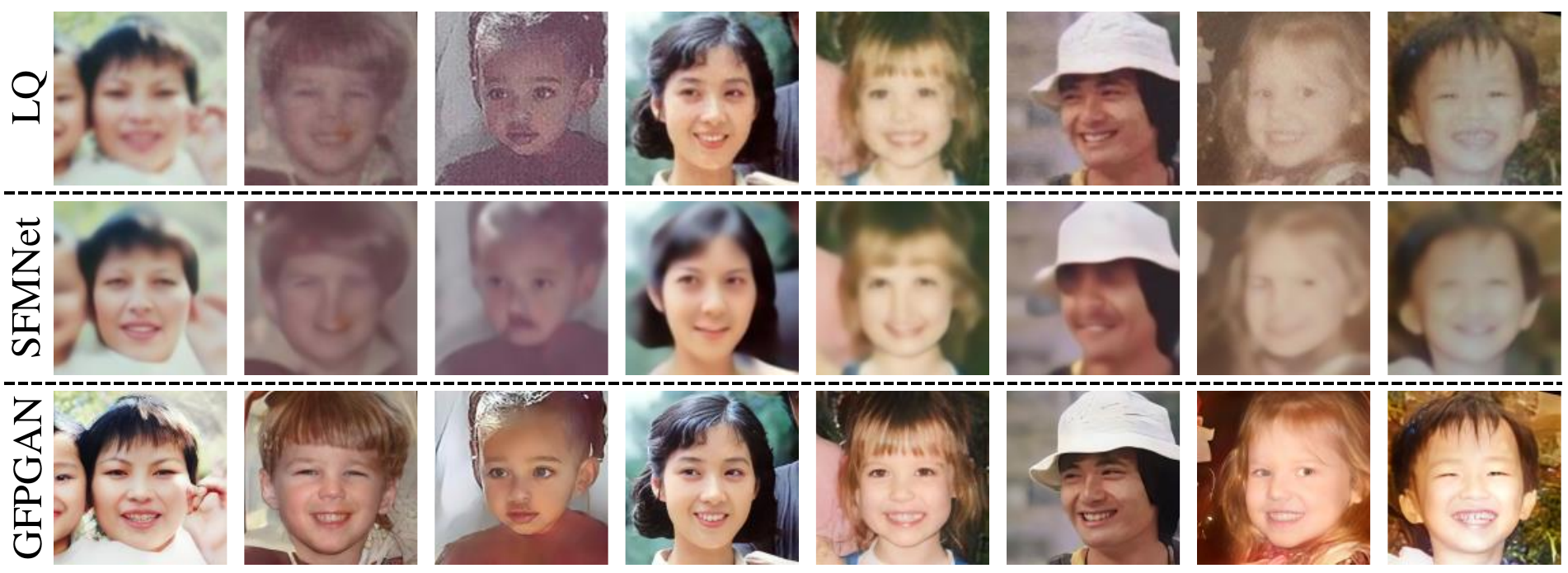}
\end{overpic}
\centering
\end{subfigure}
\vspace{-0.5mm}
\captionof{figure}{\small Comparison of non-blind/blind methods in reality.}
\label{fig:non_and_blind_comparsion}
\end{minipage}
\vspace{-4mm}
\end{table*}

\section{Methods Analysis}~\label{DS}
In this section, we conducted a comprehensive evaluation of the key non-blind and blind face restoration methods. And due to the extensive range of joint tasks, their comparisons are in the \emph{supplementary material}.

\subsection{Experimental Setting} 
\noindent$\bullet$~\textbf{Non-blind Tasks.} We utilized the initial 18,000 images from CelebA dataset ~\cite{liu2015deep} for training purpuse. For testing, we randomly selected 1,000 images from CelebA dataset and 50 random images from Helen dataset ~\cite{le2012interactive}. All images were cropped and resized to a size of 128$ \times $128. The LQ images were derived by downsampling the HQ images using bicubic interpolation, as described in Eq. \ref{bicubic}.

\noindent$\bullet$~\textbf{Blind Tasks.} 
We followed the degradation model used in GFPGAN~\cite{wang2021towards} and conducted training and testing on the FFHQ and CelebA-HQ datasets, respectively. The degradation process is defined by Eq. \ref{blind} and Eq. \ref{blind_SR}, which represent blind restoration and blind super-resolution respectively. In these equations, the parameters $\sigma $, $\delta $, $r$, and $q$ of the degradation model are randomly drawn from the ranges ${\rm{\{ 0}}{\rm{.2 : 10\} }}$, ${\rm{\{ 1 : 8\} }}$, ${\rm{\{ 0 : 20\} }}$, and ${\rm{\{ 60 : 100\} }}$, respectively. Furthermore, to ensure a more comprehensive evaluation, we incorporated real-world datasets such as LFW-Test, WebPhoto-Test, CelebChild, and CelebAdult. All images were aligned and resized to a size of 512$ \times $512.

\begin{table*}[t!]
\setlength\tabcolsep{2pt}
\centering
\vspace{0mm}
\caption{\small Comparison of primary blind method performance on synthetic test set CelebA-HQ and real datasets LFW-Test, WebPhoto, Celeb-Child, and Celeb-Adult. Speed and overhead comparison of typical non-blind methods that measured on $512 \times 512$ images.}
\vspace{-2.5mm}
\label{tab:blind_performance}
\resizebox{1\textwidth}{!}{
\begin{tabular}{l|c|c|c||cccccc|cc|cc|cc|cc}
\hline
\toprule
\rowcolor{lightgray}
& & & & \multicolumn{6}{c|}{CelebA-HQ} 
& \multicolumn{2}{c|}{LFW-Test} 
& \multicolumn{2}{c|}{WebPhoto} 
& \multicolumn{2}{c|}{CelebChild}
& \multicolumn{2}{c}{CelebAdult}
\\ 
\cmidrule{5-18}
    \rowcolor{lightgray}
    \multicolumn{1}{l|}{\multirow{-2}{*}{Methods}} 
    & \multicolumn{1}{c|}{\multirow{-2}{*}{Param}}
    & \multicolumn{1}{c|}{\multirow{-2}{*}{MACs}}
    & \multicolumn{1}{c||}{\multirow{-2}{*}{Speed}}
    & PSNR$\uparrow$  & SSIM$\uparrow$    & LPIPS$\downarrow$  
    & IDD$\downarrow$ & FID$\downarrow$   & NIQE$\downarrow$  
    
    & FID$\downarrow$  & NIQE$\downarrow$ 
    & FID$\downarrow$  & NIQE$\downarrow$ 
    & FID$\downarrow$  & NIQE$\downarrow$
    & FID$\downarrow$  & NIQE$\downarrow$
    \\ 
    \hline\hline
    PSFRGAN~\cite{chen2021progressive}     & 60.2M   & 464.9G  & \cellcolor{tinygray}\underline{53ms}     & 24.65  & .6443  & .4199  & .6664  & 43.33  & 4.099  & 49.53  & 4.095 & 84.98  & \cellcolor{tinygray}\underline{4.151}  & 106.6  & 4.670  & 104.1  & 4.246  \\ 
    
    HiFaceGAN~\cite{yang2020hifacegan}     & 79.9M   & 40.7G   & 90ms     & 24.92  & .6195  & .4770  & .7310  & 66.09  & 5.002  & 64.50  & 4.520 & 116.1  & 4.943  & 113.0  & 4.871  & \cellcolor{tinygray}\underline{104.0}  & 4.340\\
    
    DFDNet~\cite{li2020blind}              & 133.3M  & 599.8G  & 2.1s     & 24.26  & .6042  & .4421  & .6884  & 54.34  & 5.921  & 59.69  & 4.776 & 93.28  & 5.812  & 107.1  & 4.452  & 105.6  & \cellcolor{tinygray}\underline{3.782}\\

    GPEN~\cite{yang2021gan}                & 71.1M   & 138.1G  & 235ms    & 25.59  & .6894  & .4009  & .6019  & \multicolumn{1}{>{\columncolor{lightgray}}l}{ \bf{36.46}}  & 5.364  & 57.00  & 5.071 & 101.3  & 6.326  & 112.1  & 4.945 & 110.8  & 4.362\\

    GFPGAN~\cite{wang2021towards}          & 48.7M   & \cellcolor{tinygray}\underline{51.6G}   & \cellcolor{lightgray}\bf{46ms}     & 25.08  & .6777  & .3646  & .5709  & 42.59  & 4.158  & 50.04  & 3.965 & 87.13  & 4.228  & 111.4  & 4.447  & 105.0  & 4.033\\

    VQFR~\cite{gu2022vqfr}                 & 71.8M   & 1.07T   & 495ms    & 24.14  & .6360  & .3515  & .5959  & 41.29  & \cellcolor{lightgray}\bf{3.693}  & 50.65  & \cellcolor{lightgray}\bf{3.590} & \multicolumn{1}{>{\columncolor{lightgray}}c}{ \bf{75.41}}  & \cellcolor{lightgray}\bf{3.608}  & \cellcolor{tinygray}\underline{105.2}  & \cellcolor{lightgray}\bf{3.938}  & 105.0  & \multicolumn{1}{>{\columncolor{lightgray}}c}{ \bf{3.756}}\\

    GCFSR~\cite{he2022gcfsr}               & 88.7M   & 119.8G  & 145ms    & \multicolumn{1}{>{\columncolor{lightgray}}c}{ \bf{26.31}}  & \multicolumn{1}{>{\columncolor{lightgray}}c}{ \bf{.7085}}  & \multicolumn{1}{>{\columncolor{lightgray}}c}{ \bf{.3400}}  & \multicolumn{1}{>{\columncolor{lightgray}}c}{ \bf{.5122}}  & 50.10  & 4.943  & 52.23  & 4.998 & 93.27  & 5.640  & 115.1  & 5.326  & 107.1  & 4.824\\

    SGPN~\cite{zhu2022blind}               & \cellcolor{lightgray}\bf{15.2M}   & \cellcolor{lightgray}\bf{18.3G}   & 134ms    & 24.93  & .6583  & .3702  & .6028  & \cellcolor{tinygray}\underline{39.44}  & \cellcolor{tinygray}\underline{4.095}  & \cellcolor{tinygray}\underline{44.95}  & \cellcolor{tinygray}\underline{3.863} & \cellcolor{tinygray}\underline{75.61}  & \cellcolor{tinygray}\underline{4.269}  & 109.4  & 4.234  & 104.9  & 4.402\\

    RestoreFormer~\cite{wang2022restoreformer}  & 72.4M   & 340.8G & 172ms & 24.64  & .6060  & .3655  & \cellcolor{tinygray}\underline{.5339}  & 41.82  & 4.405  & 48.38  & 4.169 & 77.33  & 4.459  & \multicolumn{1}{>{\columncolor{lightgray}}l}{ \bf{101.2}}  & 4.580  & \multicolumn{1}{>{\columncolor{lightgray}}l}{ \bf{103.5}}  & 4.321\\

    CodeFormer~\cite{zhou2022towards}      & 73.6M   & 292.4G  & 98ms     & 25.15  & .6699  & \cellcolor{tinygray}\underline{.3432}  & .6171  & 52.43  & 4.650  & 52.36  & 4.484 & 83.19  & 4.705  & 116.2  & 4.983  & 111.1  & 4.541\\

    DMDNet~\cite{li2023learning}           & \cellcolor{tinygray}\underline{40.4M}   & 187.2G  & 219ms    & \cellcolor{tinygray}\underline{25.62}  & \cellcolor{tinygray}\underline{.6933}  & .3670  & .6179  & 39.94  & 4.786  & \multicolumn{1}{>{\columncolor{lightgray}}l}{ \bf{43.38}}  & 4.617 & 88.55  & 5.154  & 114.2  & 4.884  & 114.2  & 4.884\\
    \hline\hline
    Input
    & -  & -  & -  & 23.35  & .6848  & .4866  & .8577  & 144.0  & 13.2  & 137.6  & 11.0  & 170.1  & 12.7  & 144.4  & 9.03  & 118.3  & 7.56\\ 
    \bottomrule
\end{tabular}}
\end{table*}
\hspace{-2.5mm}

\begin{figure*}[t!]
\hspace{-2.5mm}
\setlength\tabcolsep{2pt}
\begin{minipage}{0.325\linewidth}
\begin{subfigure}[h]{\linewidth}
\begin{overpic}[width=\linewidth]{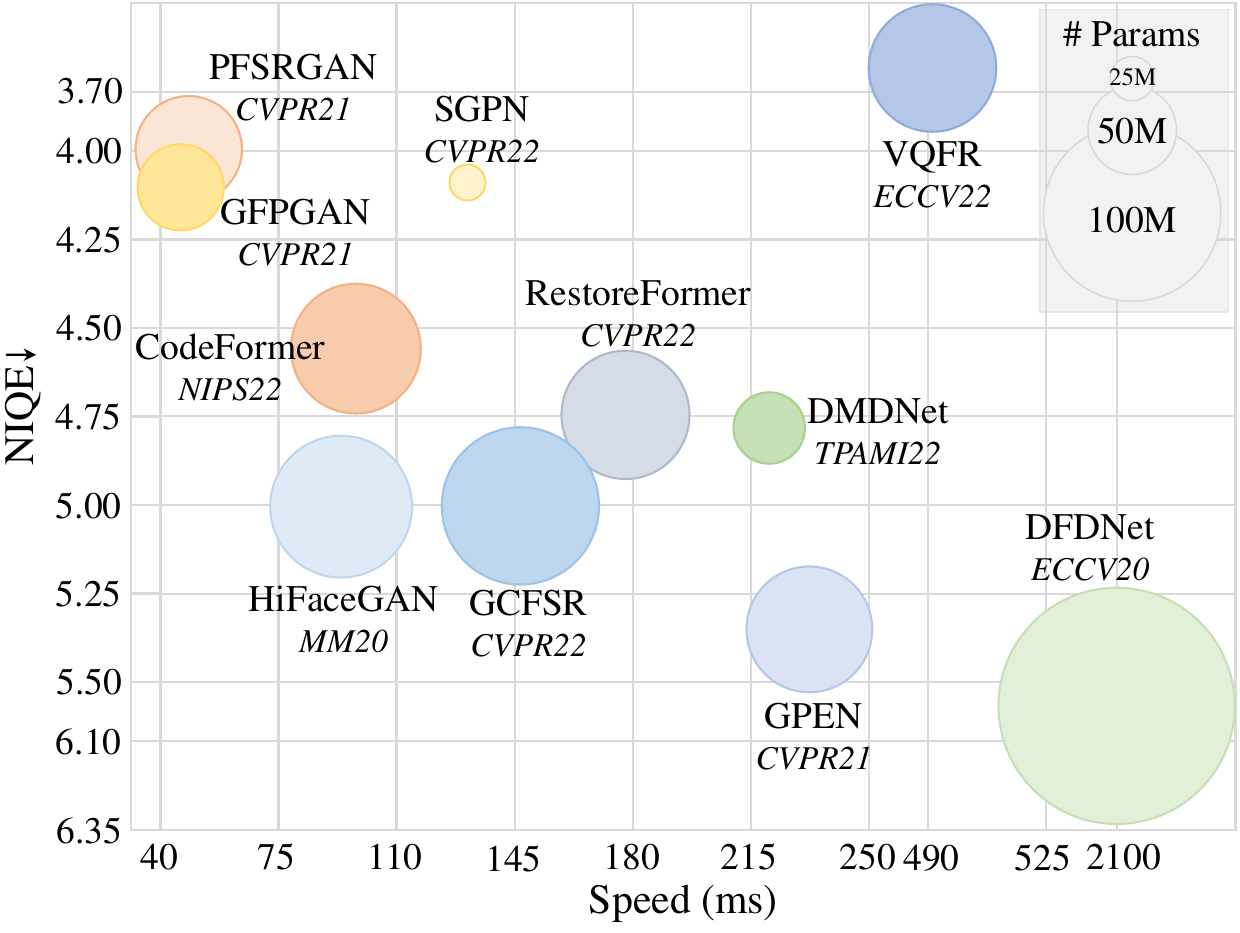}
\put(69, 15)
{\tiny\linethickness{0.3mm}~\cite{yang2021gan}}
\put(71, 55.5)
{\tiny\linethickness{0.3mm}~\cite{gu2022vqfr}}
\put(76, 40)
{\tiny\linethickness{0.3mm}~\cite{li2023learning}}
\put(92, 32)
{\tiny\linethickness{0.3mm}~\cite{li2020blind}}
\put(59.5, 50.5)
{\tiny\linethickness{0.3mm}~\cite{wang2022restoreformer}}
\put(39, 19.5)
{\tiny\linethickness{0.3mm}~\cite{he2022gcfsr}}
\put(24, 20)
{\tiny\linethickness{0.3mm}~\cite{yang2020hifacegan}}
\put(43, 64)
{\tiny\linethickness{0.3mm}~\cite{zhu2022blind}}
\put(29, 57)
{\tiny\linethickness{0.3mm}~\cite{wang2021towards}}
\put(13, 40)
{\tiny\linethickness{0.3mm}~\cite{zhou2022towards}}
\put(30, 69)
{\tiny\linethickness{0.3mm}~\cite{chen2021progressive}}
\end{overpic}
\centering
\end{subfigure}
\vspace{-3mm}
\captionof{figure}{\small Complexity analysis of blind methods on synthetic test set CelebA-HQ.}
\label{fig:blind_time_performance}
\end{minipage}
~
\begin{minipage}{0.325\linewidth}
\begin{subfigure}[h]{\linewidth}
\begin{overpic}[width=\linewidth]{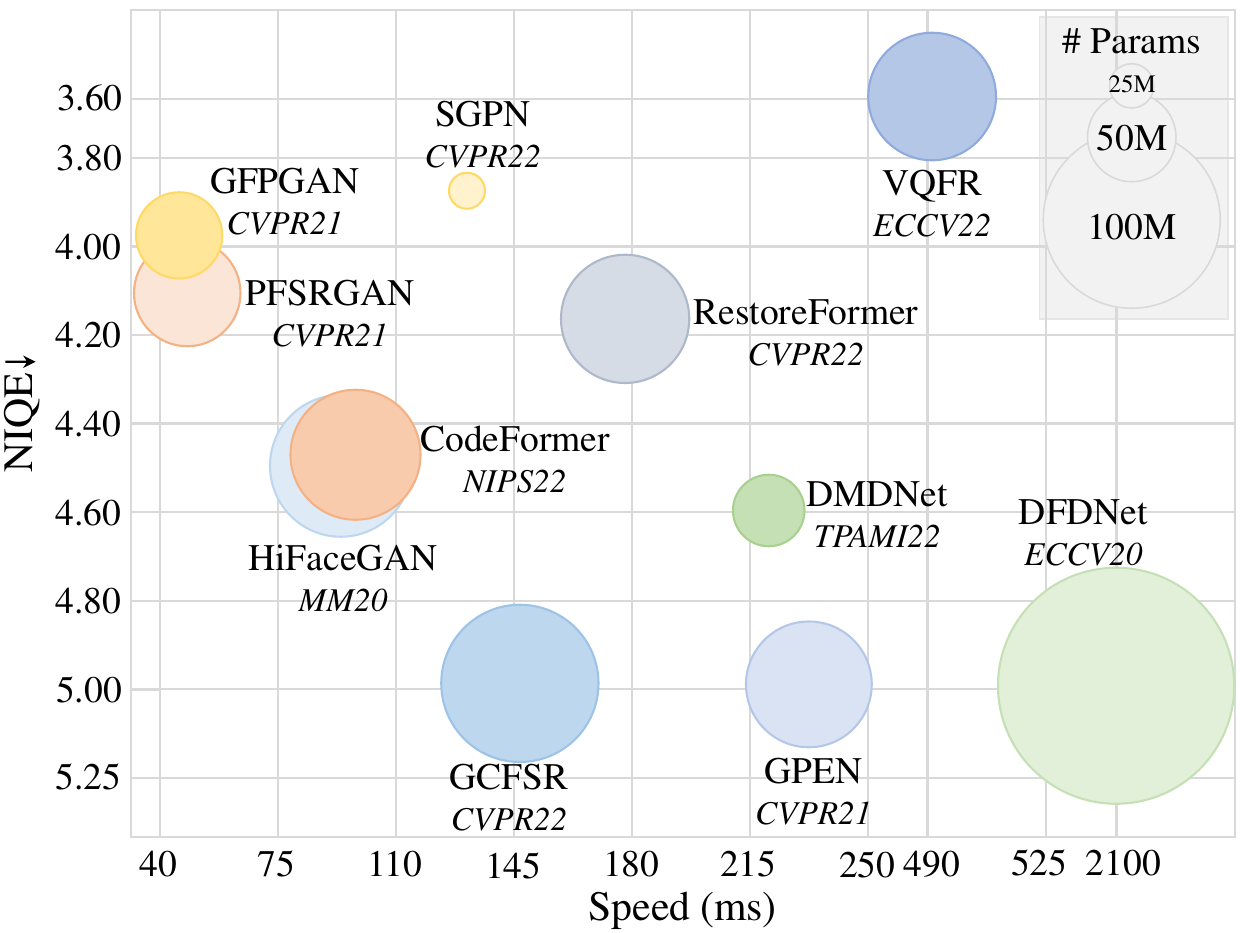}
\put(91.5, 33)
{\tiny\linethickness{0.3mm}~\cite{li2020blind}}
\put(75.5, 34)
{\tiny\linethickness{0.3mm}~\cite{li2023learning}}
\put(69, 12)
{\tiny\linethickness{0.3mm}~\cite{yang2021gan}}
\put(71, 53.5)
{\tiny\linethickness{0.3mm}~\cite{gu2022vqfr}}
\put(73, 48.5)
{\tiny\linethickness{0.3mm}~\cite{wang2022restoreformer}}
\put(45, 11)
{\tiny\linethickness{0.3mm}~\cite{he2022gcfsr}}
\put(48, 38)
{\tiny\linethickness{0.3mm}~\cite{zhou2022towards}}
\put(34, 29)
{\tiny\linethickness{0.3mm}~\cite{yang2020hifacegan}}
\put(32.5, 50.5)
{\tiny\linethickness{0.3mm}~\cite{chen2021progressive}}
\put(42, 65)
{\tiny\linethickness{0.3mm}~\cite{zhu2022blind}}
\put(28, 59.5)
{\tiny\linethickness{0.3mm}~\cite{wang2021towards}}
\end{overpic}
\centering
\end{subfigure}
\vspace{-2mm}
\captionof{figure}{\small Complexity analysis of blind methods on the real test set LFW-Test.}
\label{fig:real_time_performance}
\end{minipage}
~
\begin{minipage}{0.325\linewidth}
\begin{subfigure}[h]{\linewidth}
\begin{overpic}[width=\linewidth]{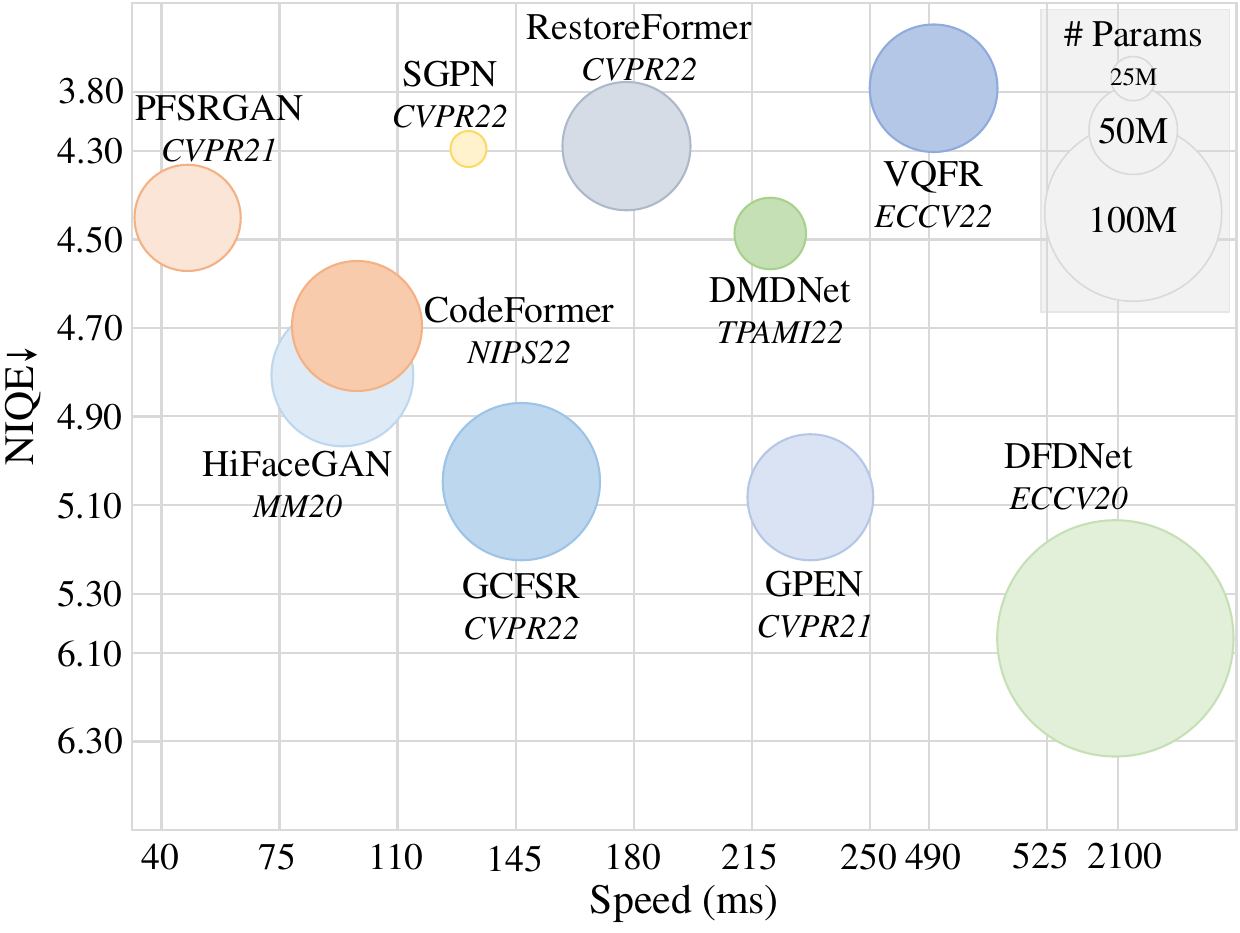}
\put(46, 26)
{\tiny\linethickness{0.3mm}~\cite{he2022gcfsr}}
\put(48.5, 48.5)
{\tiny\linethickness{0.3mm}~\cite{zhou2022towards}}
\put(68.5, 26)
{\tiny\linethickness{0.3mm}~\cite{yang2021gan}}
\put(68, 50)
{\tiny\linethickness{0.3mm}~\cite{li2023learning}}
\put(71, 54)
{\tiny\linethickness{0.3mm}~\cite{gu2022vqfr}}
\put(90.5, 37)
{\tiny\linethickness{0.3mm}~\cite{li2020blind}}
\put(60, 72)
{\tiny\linethickness{0.3mm}~\cite{wang2022restoreformer}}
\put(28, 34.5)
{\tiny\linethickness{0.3mm}~\cite{yang2020hifacegan}}
\put(23.5, 65.5)
{\tiny\linethickness{0.3mm}~\cite{chen2021progressive}}
\put(39, 68.2)
{\tiny\linethickness{0.3mm}~\cite{zhu2022blind}}
\end{overpic}
\centering
\end{subfigure}
\vspace{-2mm}
\captionof{figure}{\small Complexity analysis of blind FSR methods on CelebA-HQ ($\times 8$).}
\label{fig:SRX8_time_performance}
\end{minipage}
\vspace{-2mm}
\end{figure*}

\begin{figure*}
\begin{center}
\begin{overpic}[width=1\linewidth]{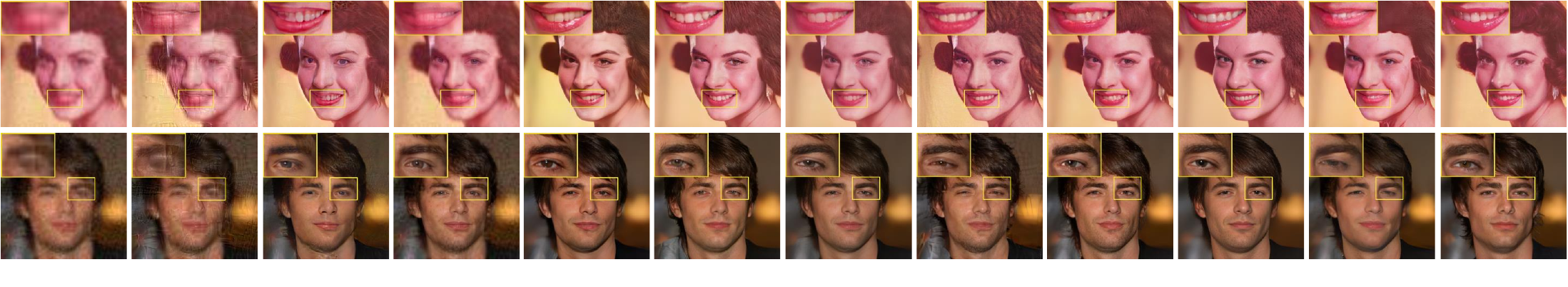}
\put(0.1,0.45){\color{black}{\fontsize{6pt}{1pt}\selectfont Synthetic Input}}
\put(9.1,0.45){\color{black}{\fontsize{6pt}{1pt}\selectfont HiFace~\cite{yang2020hifacegan}}}
\put(17.0,0.45){\color{black}{\fontsize{6pt}{1pt}\selectfont PFSRGAN~\cite{chen2021progressive}}}
\put(26.3,0.45){\color{black}{\fontsize{6pt}{1pt}\selectfont GPEN~\cite{yang2021gan}}}
\put(34.0,0.45){\color{black}{\fontsize{6pt}{1pt}\selectfont GFPGAN~\cite{wang2021towards}}}
\put(42.9,0.45){\color{black}{\fontsize{6pt}{1pt}\selectfont VQFR~\cite{gu2022vqfr}}}
\put(51.0,0.45){\color{black}{\fontsize{6pt}{1pt}\selectfont GCFSR~\cite{he2022gcfsr}}}
\put(59.5,0.45){\color{black}{\fontsize{6pt}{1pt}\selectfont SGPN~\cite{zhu2022blind}}}
\put(67.7,0.45){\color{black}{\fontsize{6pt}{1pt}\selectfont Restore~\cite{wang2022restoreformer}}}
\put(74.8,0.45){\color{black}{\fontsize{5.68pt}{1pt}\selectfont CodeFormer~\cite{zhou2022towards}}}
\put(84.1,0.45){\color{black}
{\fontsize{6pt}{1pt}\selectfont DMDNet~\cite{li2023learning}}}
\put(92.4,0.45){\color{black}
{\fontsize{6pt}{1pt}\selectfont Grouth-truth}}
\end{overpic}
\end{center}
\vspace{-5mm}
\caption{\small Visual comparison of different blind methods on the CelebA-HQ test set for blind face restoration.}
\label{blind_visual}
\vspace{-5mm}
\end{figure*}

\noindent$\bullet$~\textbf{Evaluation Metric.} 
We employed fully reference metrics, such as PSNR, SSIM, LPIPS, and IDD. These metrics  assess various aspects including pixel structure similarity, visual fidelity, and identity preservation. In addition, we also utilized non-reference or semi-reference metrics like NIQE and FID. These metrics allow us to evaluate image fidelity and visual quality without the need for actual landmarks or reference images.

\subsection{Quantitative Evaluation}
Regarding the non-blind task, we chose to focus on evaluating non-blind super-resolution methods due to their predominant emphasis in the field. TABLE~\ref{tab:non_blind_performance} presents a compilation of ten state-of-the-art non-blind methods, including fine-tuned image restoration methods~\cite{zhang2018image,liang2021swinir}, methods based on attention mechanisms~\cite{lu2021face,gao2023ctcnet,bao2023sctanet,wang2023spatial}, and methods relying on various priors~\cite{chen2018fsrnet,xin2020facial,ma2020deep}. Among these, methods employing hybrid attention mechanisms, namely CTCNet~\cite{gao2023ctcnet}, SCTANet~\cite{bao2023sctanet}, and SFMNet~\cite{wang2023spatial}, achieve either the best or second-best performance across all metrics on both test sets. TABLE~\ref{tab:non_blind_speed} provides detailed information about the model characteristics of these methods, including parameters, computation, and inference duration. Furthermore, Fig.~\ref{fig:non_blind_time_psnr} visually illustrates the efficiency of these techniques through three perspectives: performance, inference speed, and model size. Notably, attention-based methods, particularly SFMNet~\cite{wang2023spatial}, stand out as they achieve superior performance while maintaining smaller computational loads. Finally, Fig.~\ref{non_blind_visual} provides a visual comparison of these methods. 

A range of state-of-the-art methods were selected for the blind task, including approaches that do not rely on prior such as network architecture design~\cite{yang2020hifacegan,he2022gcfsr} and diffusion modeling techniques~\cite{saharia2022image}). Additionally, techniques utilizing internally-specific priors such as parsing maps~\cite{chen2021progressive} and 3D face shapes~\cite{zhu2022blind} were considered. Furthermore, methods employing external compensatory prior like pre-trained StyleGAN prior~\cite{yang2021gan,wang2021towards,zhu2022blind}, pre-trained VQGAN prior~\cite{gu2022vqfr,zhou2022towards}, face dictionary~\cite{li2020blind,wang2022restoreformer}, and reference prior~\cite{li2023learning}) were also included. To evaluate the application scope of non-blind and blind methods, we randomly selected several real face photos and restored them using the fine-tuned non-blind method SFMNet~\cite{wang2023spatial} and the blind method GFPGAN~\cite{wang2021towards} respectively. As shown in Fig.~\ref{fig:non_and_blind_comparsion}, it is evident that SFMNet struggles to effectively handle real-world photos, while GFPGAN, despite showing some racial bias in certain images, generally offers superior visual quality. Consequently, the blind method holds greater promise for real-world applications.

\begin{figure*}
\begin{center}
\begin{overpic}[width=0.99\linewidth]{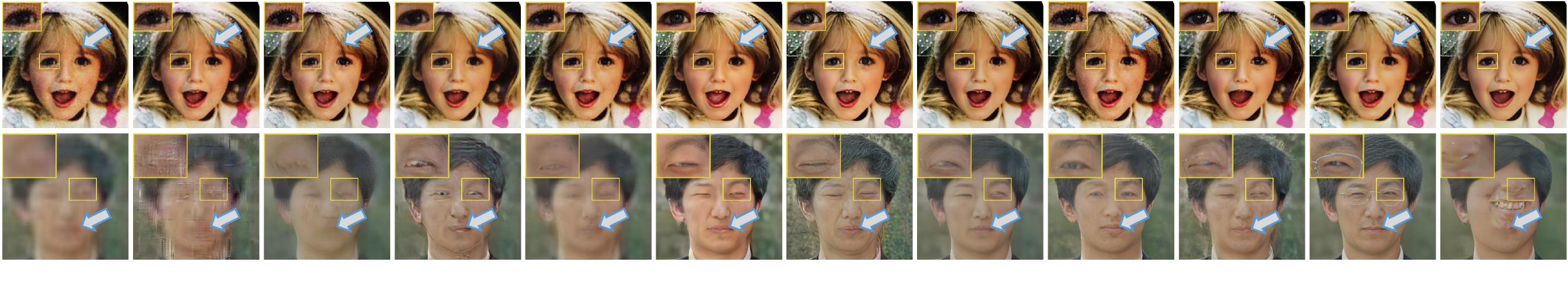}
\put(1.35,0.35){\color{black}{\fontsize{6pt}{1pt}\selectfont Real Input}}
\put(9.3,0.35){\color{black}{\fontsize{6pt}{1pt}\selectfont HiFace~\cite{yang2020hifacegan}}}
\put(17.2,0.35){\color{black}{\fontsize{6pt}{1pt}\selectfont DFDNet~\cite{li2020blind}}}
\put(25.15,0.35){\color{black}{\fontsize{6pt}{1pt}\selectfont PFSRGAN~\cite{chen2021progressive}}}
\put(34.6,0.35){\color{black}{\fontsize{6pt}{1pt}\selectfont GPEN~\cite{yang2021gan}}}
\put(42.4,0.35){\color{black}{\fontsize{6pt}{1pt}\selectfont GFPGAN~\cite{wang2021towards}}}
\put(51.3,0.35){\color{black}{\fontsize{6pt}{1pt}\selectfont VQFR~\cite{gu2022vqfr}}}
\put(59.5,0.35){\color{black}{\fontsize{6pt}{1pt}\selectfont GCFSR~\cite{he2022gcfsr}}}
\put(67.9,0.35){\color{black}{\fontsize{6pt}{1pt}\selectfont SGPN~\cite{zhu2022blind}}}
\put(76.15,0.35){\color{black}{\fontsize{6pt}{1pt}\selectfont Restore~\cite{wang2022restoreformer}}}
\put(83.1,0.35){\color{black}
{\fontsize{5.7pt}{1pt}\selectfont CodeFormer~\cite{zhou2022towards}}}
\put(92.3,0.35){\color{black}
{\fontsize{6pt}{1pt}\selectfont DMDNet~\cite{li2023learning}}}
\end{overpic}
\end{center}
\vspace{-4mm}
\caption{\small Qualitative comparison of restoration for the real test sets, including Celeb-Child (first), and Celeb-Adult (second row).}
\vspace{-1mm}
\label{real_visual}
\end{figure*}

\begin{figure*}
\begin{center}
\includegraphics[width=0.99\linewidth]{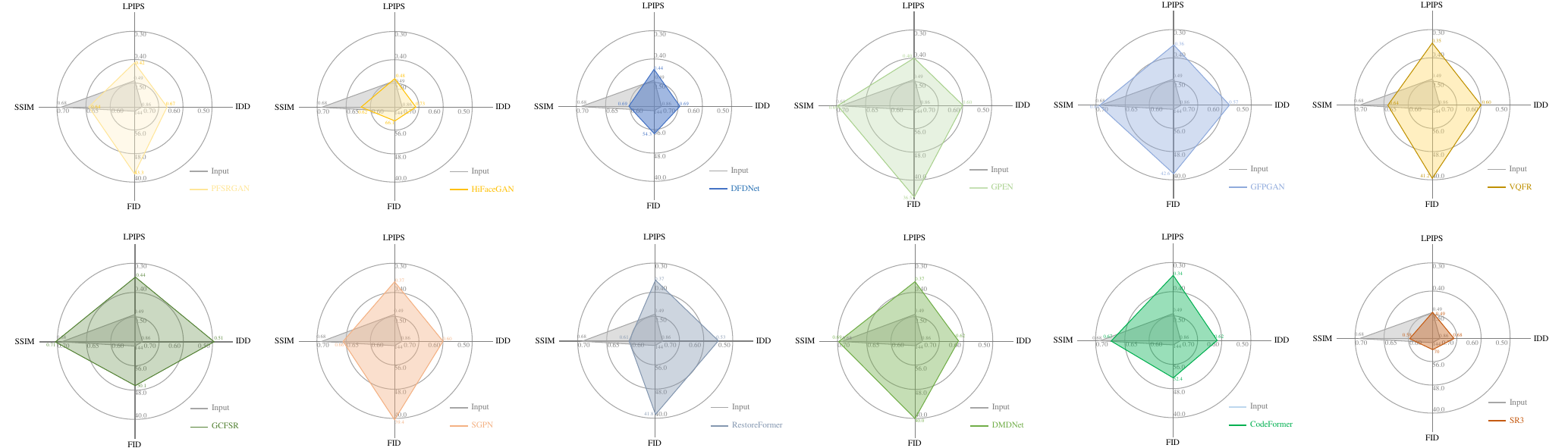}
\vspace{-2mm}
\end{center}
   \caption{\small Balanced analysis of various blind methods across four major metrics: SSIM, LPIPS, FID, and IDD.}
\label{balance}
\vspace{-4mm}
\end{figure*}

In the context of the blind task, our evaluation primarily focuses on blind face restoration, as blind methods primarily emphasize this specific direction. We also complement the evaluation with blind super-resolution. TABLE~\ref{tab:blind_performance} presents a comprehensive quantitative assessment of these techniques across three dimensions: model size, inference speed, and performance on synthetic and real datasets. It can be observed that GCFSR achieves the best performance in several metrics that measure the structural similarity of restored face images. In terms of image fidelity and perceptual quality, pre-trained GAN-based methods, with VQFR~\cite{gu2022vqfr} being a notable representative, exhibit superior performance. Methods such as DMDNet~\cite{wang2022restoreformer}, SGPN~\cite{zhu2022blind}, and GPEN~\cite{yang2021gan} strike a better balance between structural similarity and perceptual quality. Furthermore, to handle more complex degradation, blind methods tend to employ larger models compared to non-blind approaches, resulting in slower inference times. Fig.~\ref{fig:blind_time_performance} and Fig.~\ref{fig:real_time_performance} illustrate the efficiency trade-offs of these methods on the synthetic and real test sets, respectively. In these figures, methods closer to the upper-left corner with smaller circles are considered more efficient. The figure demonstrate that both PFSRGAN~\cite{chen2021progressive} and pre-trained GAN-based methods~\cite{wang2021towards,zhu2022blind,zhou2022towards} are more efficient. Comparative visualization of their visual effects can be observed in Fig.~\ref{blind_visual} and Fig.~\ref{real_visual}. It's noticeable that methods~\cite{wang2021towards,gu2022vqfr,zhou2022towards} relying on pre-trained GAN priors tend to achieve superior performance when dealing with severely degraded facial images. Finally, as depicted in Fig.~\ref{balance}, we have selected four metrics - SSIM for face similarity, IDD for identity consistency, LPIPS for sensory quality assessment, and FID for image fidelity - to highlight the strengths and weaknesses of each method in terms of face image quality. It is evident that some methods~\cite{zhu2022blind,wang2022restoreformer}, while exhibiting better sensory quality, show subpar performance in two metrics, such as SSIM and IDD. On the other hand, methods~\cite{he2022gcfsr,zhou2022towards} with higher structural and identity similarity often display inferior sensory quality. Therefore, there is a clear need for the development of more balanced approaches to address these disparities.

\begin{figure*}
\begin{center}
\begin{overpic}[width=0.99\linewidth]{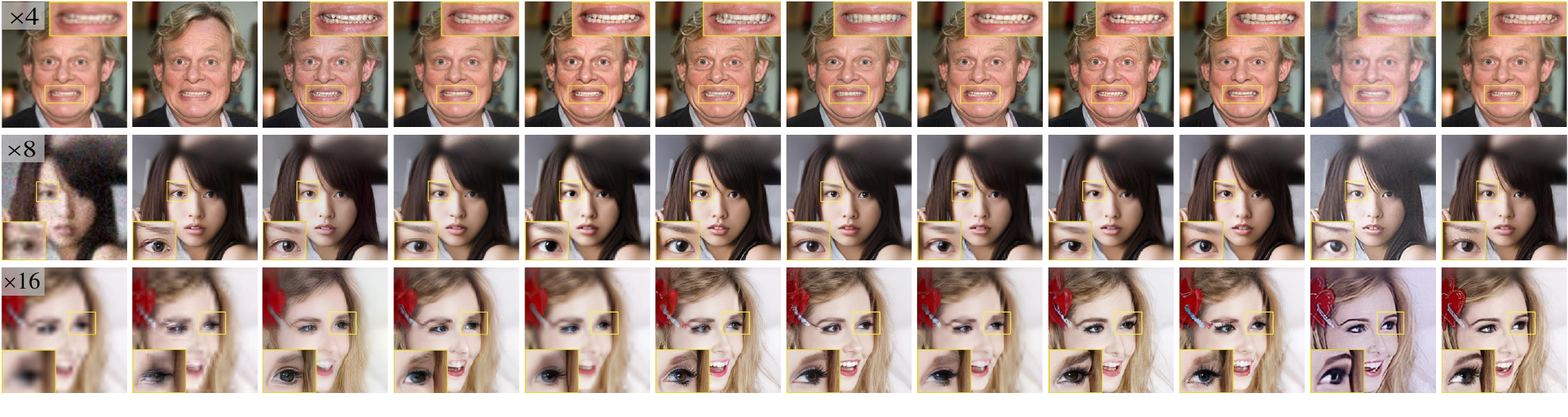}
\put(0.05,0.35){\color{black}{\fontsize{6pt}{1pt}\selectfont Synthetic Input}}
\put(9.3,0.35){\color{black}{\fontsize{6pt}{1pt}\selectfont HiFace~\cite{yang2020hifacegan}}}
\put(17.2,0.35){\color{black}{\fontsize{6pt}{1pt}\selectfont DFDNet~\cite{li2020blind}}}
\put(25.2,0.35){\color{black}{\fontsize{6pt}{1pt}\selectfont PFSRGAN~\cite{chen2021progressive}}}
\put(34.6,0.35){\color{black}{\fontsize{6pt}{1pt}\selectfont GPEN~\cite{yang2021gan}}}
\put(42.8,0.35){\color{black}{\fontsize{6pt}{1pt}\selectfont VQFR~\cite{gu2022vqfr}}}
\put(49.8,0.35){\color{black}{\fontsize{5.7pt}{1pt}\selectfont CodeFormer~\cite{zhou2022towards}}}
\put(59.5,0.35){\color{black}{\fontsize{6pt}{1pt}\selectfont GCFSR~\cite{he2022gcfsr}}}
\put(67.9,0.35){\color{black}{\fontsize{6pt}{1pt}\selectfont SGPN~\cite{zhu2022blind}}}
\put(76.15,0.35){\color{black}{\fontsize{6pt}{1pt}\selectfont Restore~\cite{wang2022restoreformer}}}
\put(85.2,0.35){\color{black}
{\fontsize{6pt}{1pt}\selectfont SR3~\cite{saharia2022image}}}
\put(92.3,0.35){\color{black}
{\fontsize{6pt}{1pt}\selectfont Grouth-truth}}
\end{overpic}
\end{center}
\vspace{-4mm}
\caption{\small Qualitative comparisons on CelebA-HQ for $ \times {\rm{4}}$ (first row), $ \times {\rm{8}}$ (second row), $ \times {\rm{16}}$ (third row) super-resolution.}
\label{SR_visual}
\vspace{-4mm}
\end{figure*}

The second part focuses on blind super-resolution, and TABLE~\ref{tab:blind_SR_performance} provides a comprehensive quantitative performance comparison of these methods at three scales: $ \times $4, $ \times $8, and $ \times $16. It is apparent that priori-free methods like GCFSR~\cite{he2022gcfsr} and HiFaceGAN~\cite{yang2020hifacegan} excel in face structure restoration. However, they exhibit shortcomings in FID and NIQE metrics, suggesting that their restored images might lack realism and may contain artifacts. On the other hand, pre-trained GAN-based approaches such as GPEN~\cite{yang2021gan}, VQFR~\cite{gu2022vqfr}, and SGPN~\cite{zhu2022blind} perform better in these two metrics, indicating more realistic and artifact-free results. Moving forward, Fig.~\ref{fig:SRX8_time_performance} illustrates the efficiency of methods at the $ \times $8 scale, with PFSRGAN~\cite{chen2021progressive} and SGPN~\cite{zhu2022blind} emerging as the more efficient choices. Lastly, in Fig.~\ref{SR_visual}, we present a visual comparison of methods at three scales. Notably, SGPN~\cite{zhu2022blind} and CodeFormer~\cite{zhou2022towards}, leveraging pre-trained GAN priors, perform favorably without introducing artifacts when dealing with substantial downsampling factors.

\begin{table*}[t!]
\setlength\tabcolsep{2pt}
\centering
\caption{\small Quantitative comparisons with primary blind methods on CelebA-HQ for $\times 4$, $\times 8$, $\times 16$ super-resolution.}
\vspace{-2mm}
\label{tab:blind_SR_performance}
\resizebox{1\textwidth}{!}{
\begin{tabular}{l||cccccc|cccccc|cccccc}
\hline
\toprule
\rowcolor{lightgray}
& \multicolumn{6}{c|}{CelebA-HQ ($\times 4$)} 
& \multicolumn{6}{c|}{CelebA-HQ ($\times 8$)} 
& \multicolumn{6}{c}{CelebA-HQ ($\times 16$)} 
\\ 
\cmidrule{2-19}
    \rowcolor{lightgray}
    \multicolumn{1}{l||}{\multirow{-2}{*}{Methods}} 
    & PSNR$\uparrow$  & SSIM$\uparrow$    & LPIPS$\downarrow$  
    & IDD$\downarrow$ & FID$\downarrow$   & NIQE$\downarrow$  
    
    & PSNR$\uparrow$  & SSIM$\uparrow$    & LPIPS$\downarrow$  
    & IDD$\downarrow$ & FID$\downarrow$   & NIQE$\downarrow$
    
    & PSNR$\uparrow$  & SSIM$\uparrow$    & LPIPS$\downarrow$  
    & IDD$\downarrow$ & FID$\downarrow$   & NIQE$\downarrow$
    \\ 
    \hline\hline
    PSFRGAN~\cite{chen2021progressive}     & 27.99  & .7777  & .3055  & .2924  & 42.35  & 4.623  & 25.50  & .6921  & .3639  & .4445  & 47.56 & 4.446 & 23.20  & .6250  & .4216  & .8603 
    & 49.31  & 4.197\\ 
    
    HiFaceGAN~\cite{yang2020hifacegan}     & \cellcolor{tinygray}\underline{29.49}  & .8030  & .2736  & \cellcolor{lightgray}\bf{.2065}  & \cellcolor{tinygray}\underline{39.72}  & 4.535  & \cellcolor{lightgray}\bf{26.76}  & .7156  & .3496  & \cellcolor{lightgray}\bf{.3704}  & 51.32 & 4.830 & \cellcolor{tinygray}\underline{23.68}  & .6179  & .4746  & 1.012 
    & 92.31  & 5.836\\ 
    
    DFDNet~\cite{li2020blind}              & 27.47  & .7542  & .3108  & .2888  & 41.26  & 4.710  & 25.26  & .6336  & .3982  & .4097  & 45.58 & 6.054 & 23.24  & .5768  & .4713  & .9003 
    & 60.06  & 7.070\\ 

    GPEN~\cite{yang2021gan}                & 28.35  & .7974  & .2600  & .2972  & 47.83  & 4.603  & 26.60  & \cellcolor{lightgray}\bf{.7359}  & .3193  & .4052  & 54.17 & 5.086 & \cellcolor{lightgray}\bf{24.12}  & \cellcolor{lightgray}\bf{.6772} & .3950  & .8329 
    & 68.35  & 5.896\\ 

    VQFR~\cite{gu2022vqfr}                 & 26.29  & .7201  & .2989  & .3654  & 43.98  & \cellcolor{lightgray}\bf{3.884}  & 24.84  & .6657  & .3287  & .4600  & 45.72 & \cellcolor{lightgray}\bf{3.826} & 22.17  & .5853  & \cellcolor{tinygray}\underline{.3761}  & .8128 
    & \cellcolor{lightgray}\bf{38.42}  & \cellcolor{lightgray}\bf{3.431}\\ 

    GCFSR~\cite{he2022gcfsr}               & \cellcolor{lightgray}\bf{30.73}  & \cellcolor{lightgray}\bf{.8383}  & \cellcolor{lightgray}\bf{.2369}  & \cellcolor{tinygray}\underline{.2132}  & 52.02  & 4.915  & \cellcolor{tinygray}\underline{26.66}  & \cellcolor{tinygray}\underline{.7299}  & \cellcolor{tinygray}\underline{.3073}  & .4146  & 54.74 & 5.059 & 22.90  & .6371  & .3799  & .8564 
    & 46.99  & 4.622\\ 

    SGPN~\cite{zhu2022blind}               & 28.64  & .8040  & \cellcolor{tinygray}\underline{.2456}  & .2581  & 41.06  & 4.425  & 26.18  & .7127  & \cellcolor{lightgray}\bf{.3033}  & \cellcolor{tinygray}\underline{.3846}  & 44.69 & \cellcolor{tinygray}\underline{4.330} & 23.65  & \cellcolor{tinygray}\underline{.6487}  & \cellcolor{lightgray}\bf{.3602}  & \cellcolor{lightgray}\bf{.7506} 
    & 46.66  & 4.444\\ 

    RestoreFormer~\cite{wang2022restoreformer}    & 24.99  & .6669  & .3353  & .4146  & 41.38  & \cellcolor{tinygray}\underline{4.392}  & 24.65  & .6560 & .3495  & .4525  & \cellcolor{tinygray}\underline{41.66} & 4.340 & 22.60  & .5944  & .3974  & \cellcolor{tinygray}\underline{.8068}  & \cellcolor{tinygray}\underline{38.45}  & \cellcolor{tinygray}\underline{4.216}\\ 

    CodeFormer~\cite{zhou2022towards}      & 27.10  & .7465  & .3020  & .4462  & 51.30  & 4.739  & 25.75  & .6947  & .3229  & .5115  & 51.42 & 4.698 & 23.26  & .6260  & .3666  & .7776 
    & 48.69  & 4.496\\ 

    DMDNet~\cite{li2023learning}           & 28.43  & \cellcolor{tinygray}\underline{.8081}  & .2724  & .3080  & \cellcolor{lightgray}\bf{39.06}  & 4.652  & 26.31  & .7208  & .3292  & .3967  & \cellcolor{lightgray}\bf{41.49} & 4.576 & 22.91  & .6327  & .3890  & .8318 
    & 39.61  & 4.358\\ 

    SR3~\cite{saharia2022image}            & 25.02  & .4797  & .4998  & .3968  & 52.78  & 9.366  & 23.14  & .4781  & .4826  & .4735  & 61.68 & 7.614 & 22.11  & .4057  & .6377  & 1.139 
    & 111.2  & 13.78\\ 
    \hline\hline
    Input
    & 31.05  & .8488  & .2425  & .2216  & 107.0  & 8.155  & 27.51  & .7585  & .3748  & .5898  & 195.7  & 11.24  & 24.21  & .6815  & .5007  & 1.076  & 163.7  & 13.47\\ 
    \bottomrule
\end{tabular}}
\vspace{-2.5mm}
\end{table*}

\section{Challenge And Future Directions}~\label{CS}
After reviewing various tasks and techniques and evaluating some prominent methods, it is clear that significant progress has been made. However, several challenges still persist in this domain. Additionally, there are numerous promising research opportunities to tackle these challenges and further advance the field of facial restoration.

\noindent$\bullet$~\textbf{Unified Large Model.} Prominent advancements in macro-modeling, exemplified by techniques such as Generative Pre-Training (GPT) and the Segment Anything Model (SAM), have had a significant impact on the field of computer vision. However, existing face restoration techniques often have a limited scope. Most models are designed to address specific challenges such as super-resolution or deblurring, or they focus on a single joint task. Consequently, there is a pressing demand in the industry for comprehensive large-scale models that are capable of restoring a wide spectrum of degraded facial images.

\noindent$\bullet$~\textbf{Multimodal Technology.} The successful utilization of GPT-4 in integrating images and text opens up new possibilities. For example, linguistic commands can be input to achieve selective restoration of features such as hair, eyes, and skin. Language-based instructions can also be employed to achieve specific restoration effects, such as emphasizing high resolution or maintaining identity resemblance. However, current models face challenges in precisely controlling these factors due to a lack of interpretability or handling intersectionality across different domains. As a result, the interpretability of FR models and their application in multimodal tasks could emerge as significant research areas.

\noindent$\bullet$~\textbf{Face Fairness.} The majority of FR datasets, such as CelebA and FFHQ, primarily collect facial images from specific geographical regions. It leads to the current trained models focusing on recovering facial features that are typical of those specific regions, while potentially disregarding distinctions in facial characteristics across various areas, such as variations in skin color. As a result, restoration results for individuals with black or yellow skin tones may inadvertently exhibit features characteristic of white individuals. Addressing this challenge requires the development of algorithms that mitigate racial bias in FR or the creation of datasets that prioritize racial balance.

\noindent$\bullet$~\textbf{Face Privacy Protection.} With the widespread use of facial recognition technology, improving recognition accuracy in specific scenarios (such as low light or blur) is closely linked to face restoration techniques. However, during the process of repairing and recognizing faces, there is a risk of facial information leakage. This highly sensitive data is closely associated with financial transactions and access permissions. Unfortunately, current FR methods often ignore this aspect. Therefore, ensuring the protection of facial privacy during restoration remains an ongoing challenge and opportunity. 

\noindent$\bullet$~\textbf{Real-world Applications.} The challenges faced by facial restoration applications are two-fold: the disparity between synthetic and real data domains, and the significant computational costs. The domain difference is evident in the fact that real-world images undergo more complex forms of degradation compared to synthetic counterparts, resulting in persistent artifacts after applying existing restoration techniques. Additionally, the computational overhead of current methods is excessive for deployment on mobile devices, limiting their scalability. To address these challenges, research efforts should focus on developing realistic image degradation models to capture the complexities of real-world degradation, exploring unsupervised restoration approaches to alleviate the reliance on large annotated datasets, and investigating model compression and acceleration techniques to reduce computational costs. These endeavors will contribute to the advancement of applications related to video face restoration and the restoration of aged photographs, ultimately enhancing their practicality and usability.

\noindent$\bullet$~\textbf{Effective Benchmarks.} Several commonly used benchmarks in current face restoration, including datasets, loss functions, baseline network architectures, and evaluation metrics, may not provide optimal solutions. For example, some datasets may lack comprehensive coverage, leading to limited generalization of the models. Flawed loss functions may result in undesired artifacts in the restored faces. Existing network architectures may not be suitable for all restoration tasks, limiting their applicability. Additionally, evaluating restoration results solely based on quantitative metrics may overlook important aspects of human perceptual quality. Ongoing research efforts are actively addressing these issues, leading to improvements in various areas of face restoration. However, these concerns remain focal points for future investigations.

\section{Conclusion}~\label{ES}
In this review, we provide a systematic exploration of deep learning-based approaches for face restoration. We begin by discussing the factors that contribute to the degradation of facial images and artificial degradation processes. Subsequently, we categorize the field into three distinct task categories: non-blind, blind, and joint tasks, and discuss their evolution and technical characteristics. Furthermore, we shed light on prevailing methodologies that utilize facial priors, including both internal proprietary and external compensatory priors. And we summarize the prevalent strategies for enhancing the effectiveness of priors in face restoration. Then, We conduct a thorough comparison of cutting-edge methods, highlighting their respective strengths and weaknesses. Finally, we dissect the prevailing challenges within the existing paradigms and provide insights into potential directions for advancing the field. Overall, This comprehensive review aims to serve as a valuable reference for researchers who are starting their journey in developing techniques aligned with their research aspirations.

\section{Acknowledgements}
This work was supported by China Postdoctoral Science Foundation under Grant 2022M720517 and National Natural Science Foundation of China under Grant 62306043.

\bibliographystyle{IEEEtran}
\bibliography{sample-base}

\begin{thebibliography}{100}
\providecommand{\url}[1]{#1}
\csname url@samestyle\endcsname
\providecommand{\newblock}{\relax}
\providecommand{\bibinfo}[2]{#2}
\providecommand{\BIBentrySTDinterwordspacing}{\spaceskip=0pt\relax}
\providecommand{\BIBentryALTinterwordstretchfactor}{4}
\providecommand{\BIBentryALTinterwordspacing}{\spaceskip=\fontdimen2\font plus
\BIBentryALTinterwordstretchfactor\fontdimen3\font minus \fontdimen4\font\relax}
\providecommand{\BIBforeignlanguage}[2]{{%
\expandafter\ifx\csname l@#1\endcsname\relax
\typeout{** WARNING: IEEEtran.bst: No hyphenation pattern has been}%
\typeout{** loaded for the language `#1'. Using the pattern for}%
\typeout{** the default language instead.}%
\else
\language=\csname l@#1\endcsname
\fi
#2}}
\providecommand{\BIBdecl}{\relax}
\BIBdecl

\bibitem{yang2016wider}
S.~Yang, P.~Luo, C.-C. Loy, and X.~Tang, ``Wider face: A face detection benchmark,'' in \emph{Proceedings of the IEEE Conference on Computer Vision and Pattern Recognition}, 2016, pp. 5525--5533.

\bibitem{huang2008labeled}
G.~B. Huang, M.~Mattar, T.~Berg, and E.~Learned-Miller, ``Labeled faces in the wild: A database forstudying face recognition in unconstrained environments,'' in \emph{Proceedings of Workshop on Faces in 'Real-Life' Images: Detection, Alignment, and Recognition}, 2008.

\bibitem{zhang2022learning}
Z.~Zhang, Y.~Ge, Y.~Tai, X.~Huang, C.~Wang, H.~Tang, D.~Huang, and Z.~Xie, ``Learning to restore 3d face from in-the-wild degraded images,'' in \emph{Proceedings of the IEEE Conference on Computer Vision and Pattern Recognition}, 2022, pp. 4237--4247.

\bibitem{baker2000hallucinating}
S.~Baker and T.~Kanade, ``Hallucinating faces,'' in \emph{Proceedings Fourth IEEE International Conference on Automatic Face and Gesture Recognition}.\hskip 1em plus 0.5em minus 0.4em\relax IEEE, 2000, pp. 83--88.

\bibitem{chen2018fsrnet}
Y.~Chen, Y.~Tai, X.~Liu, C.~Shen, and J.~Yang, ``Fsrnet: End-to-end learning face super-resolution with facial priors,'' in \emph{Proceedings of the IEEE Conference on Computer Vision and Pattern Recognition}, 2018, pp. 2492--2501.

\bibitem{wang2021towards}
X.~Wang, Y.~Li, H.~Zhang, and Y.~Shan, ``Towards real-world blind face restoration with generative facial prior,'' in \emph{Proceedings of the IEEE Conference on Computer Vision and Pattern Recognition}, 2021, pp. 9168--9178.

\bibitem{yu2016ultra}
X.~Yu and F.~Porikli, ``Ultra-resolving face images by discriminative generative networks,'' in \emph{Proceedings of the European Conference on Computer vVision}.\hskip 1em plus 0.5em minus 0.4em\relax Springer, 2016, pp. 318--333.

\bibitem{jiang2018deep}
J.~Jiang, Y.~Yu, J.~Hu, S.~Tang, and J.~Ma, ``Deep cnn denoiser and multi-layer neighbor component embedding for face hallucination,'' \emph{Proceedings of the International Joint Conference on Artificial Intelligence}, 2018.

\bibitem{li2018learning}
X.~Li, M.~Liu, Y.~Ye, W.~Zuo, L.~Lin, and R.~Yang, ``Learning warped guidance for blind face restoration,'' in \emph{Proceedings of the European Conference on Computer vVision}, 2018, pp. 272--289.

\bibitem{yang2021gan}
T.~Yang, P.~Ren, X.~Xie, and L.~Zhang, ``Gan prior embedded network for blind face restoration in the wild,'' in \emph{Proceedings of the IEEE Conference on Computer Vision and Pattern Recognition}, 2021, pp. 672--681.

\bibitem{yu2017face}
X.~Yu and F.~Porikli, ``Face hallucination with tiny unaligned images by transformative discriminative neural networks,'' in \emph{Proceedings of the AAAI Conference on Artificial Intelligence}, vol.~31, no.~1, 2017.

\bibitem{cheng2019identity}
X.~Cheng, J.~Lu, B.~Yuan, and J.~Zhou, ``Identity-preserving face hallucination via deep reinforcement learning,'' \emph{IEEE Transactions on Circuits and Systems for Video Technology}, vol.~30, no.~12, pp. 4796--4809, 2019.

\bibitem{zhang2021recursive}
Y.~Zhang, I.~W. Tsang, Y.~Luo, C.~Hu, X.~Lu, and X.~Yu, ``Recursive copy and paste gan: Face hallucination from shaded thumbnails,'' \emph{IEEE Transactions on Pattern Analysis and Machine Intelligence}, vol.~44, no.~8, pp. 4321--4338, 2021.

\bibitem{zhong2020face}
Y.~Zhong, Y.~Pei, P.~Li, Y.~Guo, G.~Ma, M.~Liu, W.~Bai, W.~Wu, and H.~Zha, ``Face denoising and 3d reconstruction from a single depth image,'' in \emph{Proceedings of the IEEE International Conference on Automatic Face and Gesture Recognition}.\hskip 1em plus 0.5em minus 0.4em\relax IEEE, 2020, pp. 117--124.

\bibitem{jalal2021fairness}
A.~Jalal, S.~Karmalkar, J.~Hoffmann, A.~Dimakis, and E.~Price, ``Fairness for image generation with uncertain sensitive attributes,'' in \emph{Proceedings of the International Conference on Machine Learning}, 2021, pp. 4721--4732.

\bibitem{liu2019survey}
H.~Liu, X.~Zheng, J.~Han, Y.~Chu, and T.~Tao, ``Survey on gan-based face hallucination with its model development,'' \emph{IET Image Processing}, vol.~13, no.~14, pp. 2662--2672, 2019.

\bibitem{jiang2021deep}
J.~Jiang, C.~Wang, X.~Liu, and J.~Ma, ``Deep learning-based face super-resolution: A survey,'' \emph{ACM Computing Surveys}, vol.~55, no.~1, pp. 1--36, 2021.

\bibitem{wang2022survey}
T.~Wang, K.~Zhang, X.~Chen, W.~Luo, J.~Deng, T.~Lu, X.~Cao, W.~Liu, H.~Li, and S.~Zafeiriou, ``A survey of deep face restoration: Denoise, super-resolution, deblur, artifact removal,'' \emph{arXiv preprint arXiv:2211.02831}, 2022.

\bibitem{hore2010image}
A.~Hore and D.~Ziou, ``Image quality metrics: Psnr vs. ssim,'' in \emph{Proceedings of the International Conference on Pattern Recognition}.\hskip 1em plus 0.5em minus 0.4em\relax IEEE, 2010, pp. 2366--2369.

\bibitem{wang2004image}
Z.~Wang, A.~C. Bovik, H.~R. Sheikh, and E.~P. Simoncelli, ``Image quality assessment: from error visibility to structural similarity,'' \emph{IEEE Transactions on Image Processing}, vol.~13, no.~4, pp. 600--612, 2004.

\bibitem{wang2003multiscale}
Z.~Wang, E.~P. Simoncelli, and A.~C. Bovik, ``Multiscale structural similarity for image quality assessment,'' in \emph{Proceedings of the Asilomar Conference on Signals, Systems \& Computers}, vol.~2.\hskip 1em plus 0.5em minus 0.4em\relax Ieee, 2003, pp. 1398--1402.

\bibitem{zhang2018unreasonable}
R.~Zhang, P.~Isola, A.~A. Efros, E.~Shechtman, and O.~Wang, ``The unreasonable effectiveness of deep features as a perceptual metric,'' in \emph{Proceedings of the IEEE Conference on Computer Vision and Pattern Recognition}, 2018, pp. 586--595.

\bibitem{wang2022restoreformer}
Z.~Wang, J.~Zhang, R.~Chen, W.~Wang, and P.~Luo, ``Restoreformer: High-quality blind face restoration from undegraded key-value pairs,'' in \emph{Proceedings of the IEEE Conference on Computer Vision and Pattern Recognition}, 2022, pp. 17\,512--17\,521.

\bibitem{heusel2017gans}
M.~Heusel, H.~Ramsauer, T.~Unterthiner, B.~Nessler, and S.~Hochreiter, ``Gans trained by a two time-scale update rule converge to a local nash equilibrium,'' \emph{Advances in Neural Information Processing Systems}, vol.~30, 2017.

\bibitem{mittal2012making}
A.~Mittal, R.~Soundararajan, and A.~C. Bovik, ``Making a “completely blind” image quality analyzer,'' \emph{IEEE Signal Processing Letters}, vol.~20, no.~3, pp. 209--212, 2012.

\bibitem{hossfeld2016qoe}
T.~Ho{\ss}feld, P.~E. Heegaard, M.~Varela, and S.~M{\"o}ller, ``Qoe beyond the mos: an in-depth look at qoe via better metrics and their relation to mos,'' \emph{Quality and User Experience}, vol.~1, pp. 1--23, 2016.

\bibitem{grm2019face}
K.~Grm, W.~J. Scheirer, and V.~{\v{S}}truc, ``Face hallucination using cascaded super-resolution and identity priors,'' \emph{IEEE Transactions on Image Processing}, vol.~29, pp. 2150--2165, 2019.

\bibitem{lai2019low}
S.-C. Lai, C.-H. He, and K.-M. Lam, ``Low-resolution face recognition based on identity-preserved face hallucination,'' in \emph{Proceedings of the IEEE International Conference on Image Processing}.\hskip 1em plus 0.5em minus 0.4em\relax IEEE, 2019, pp. 1173--1177.

\bibitem{jiang2020dual}
K.~Jiang, Z.~Wang, P.~Yi, T.~Lu, J.~Jiang, and Z.~Xiong, ``Dual-path deep fusion network for face image hallucination,'' \emph{IEEE Transactions on Neural Networks and Learning Systems}, vol.~33, no.~1, pp. 378--391, 2020.

\bibitem{gross2010multi}
R.~Gross, I.~Matthews, J.~Cohn, T.~Kanade, and S.~Baker, ``Multi-pie,'' \emph{Image and Vision Computing}, vol.~28, no.~5, pp. 807--813, 2010.

\bibitem{song2019joint}
Y.~Song, J.~Zhang, L.~Gong, S.~He, L.~Bao, J.~Pan, Q.~Yang, and M.-H. Yang, ``Joint face hallucination and deblurring via structure generation and detail enhancement,'' \emph{International journal of computer vision}, vol. 127, pp. 785--800, 2019.

\bibitem{zhang2020copy}
Y.~Zhang, I.~W. Tsang, Y.~Luo, C.-H. Hu, X.~Lu, and X.~Yu, ``Copy and paste gan: Face hallucination from shaded thumbnails,'' in \emph{Proceedings of the IEEE Conference on Computer Vision and Pattern Recognition}, 2020, pp. 7355--7364.

\bibitem{shen2020exploiting}
Z.~Shen, W.-S. Lai, T.~Xu, J.~Kautz, and M.-H. Yang, ``Exploiting semantics for face image deblurring,'' \emph{International Journal of Computer Vision}, vol. 128, pp. 1829--1846, 2020.

\bibitem{koestinger2011annotated}
M.~Koestinger, P.~Wohlhart, P.~M. Roth, and H.~Bischof, ``Annotated facial landmarks in the wild: A large-scale, real-world database for facial landmark localization,'' in \emph{Proceedings of the IEEE International Conference on Computer Vision Workshops}, 2011, pp. 2144--2151.

\bibitem{kim2019progressive}
D.~Kim, M.~Kim, G.~Kwon, and D.-S. Kim, ``Progressive face super-resolution via attention to facial landmark,'' \emph{arXiv preprint arXiv:1908.08239}, 2019.

\bibitem{yin2020joint}
Y.~Yin, J.~Robinson, Y.~Zhang, and Y.~Fu, ``Joint super-resolution and alignment of tiny faces,'' in \emph{Proceedings of the AAAI Conference on Artificial Intelligence}, vol.~34, no.~07, 2020, pp. 12\,693--12\,700.

\bibitem{grgic2011scface}
M.~Grgic, K.~Delac, and S.~Grgic, ``Scface--surveillance cameras face database,'' \emph{Multimedia Tools and Applications}, vol.~51, pp. 863--879, 2011.

\bibitem{lu2021face}
T.~Lu, Y.~Wang, Y.~Zhang, Y.~Wang, L.~Wei, Z.~Wang, and J.~Jiang, ``Face hallucination via split-attention in split-attention network,'' in \emph{Proceedings of the ACM International Conference on Multimedia}, 2021, pp. 5501--5509.

\bibitem{gao2023ctcnet}
G.~Gao, Z.~Xu, J.~Li, J.~Yang, T.~Zeng, and G.-J. Qi, ``Ctcnet: A cnn-transformer cooperation network for face image super-resolution,'' \emph{IEEE Transactions on Image Processing}, vol.~32, pp. 1978--1991, 2023.

\bibitem{le2012interactive}
V.~Le, J.~Brandt, Z.~Lin, L.~Bourdev, and T.~S. Huang, ``Interactive facial feature localization,'' in \emph{Proceedings of the European Conference on Computer Vision}, 2012, pp. 679--692.

\bibitem{ma2020deep}
C.~Ma, Z.~Jiang, Y.~Rao, J.~Lu, and J.~Zhou, ``Deep face super-resolution with iterative collaboration between attentive recovery and landmark estimation,'' in \emph{Proceedings of the IEEE conference on computer vision and pattern recognition}, 2020, pp. 5569--5578.

\bibitem{zhao2020saan}
T.~Zhao and C.~Zhang, ``Saan: Semantic attention adaptation network for face super-resolution,'' in \emph{Proceedings of the IEEE International Conference on Multimedia and Expo}.\hskip 1em plus 0.5em minus 0.4em\relax IEEE, 2020, pp. 1--6.

\bibitem{bao2023sctanet}
Q.~Bao, Y.~Liu, B.~Gang, W.~Yang, and Q.~Liao, ``Sctanet: A spatial attention-guided cnn-transformer aggregation network for deep face image super-resolution,'' \emph{IEEE Transactions on Multimedia}, 2023.

\bibitem{yi2014learning}
D.~Yi, Z.~Lei, S.~Liao, and S.~Z. Li, ``Learning face representation from scratch,'' \emph{arXiv preprint arXiv:1411.7923}, 2014.

\bibitem{tu2021joint}
X.~Tu, J.~Zhao, Q.~Liu, W.~Ai, G.~Guo, Z.~Li, W.~Liu, and J.~Feng, ``Joint face image restoration and frontalization for recognition,'' \emph{IEEE Transactions on Circuits and Systems for Video Technology}, vol.~32, no.~3, pp. 1285--1298, 2021.

\bibitem{liu2015deep}
Z.~Liu, P.~Luo, X.~Wang, and X.~Tang, ``Deep learning face attributes in the wild,'' in \emph{Proceedings of the IEEE International Conference on Computer Vision}, 2015, pp. 3730--3738.

\bibitem{chen2020learning}
C.~Chen, D.~Gong, H.~Wang, Z.~Li, and K.-Y.~K. Wong, ``Learning spatial attention for face super-resolution,'' \emph{IEEE Transactions on Image Processing}, vol.~30, pp. 1219--1231, 2020.

\bibitem{wang2023spatial}
C.~Wang, J.~Jiang, Z.~Zhong, and X.~Liu, ``Spatial-frequency mutual learning for face super-resolution,'' in \emph{Proceedings of the IEEE Conference on Computer Vision and Pattern Recognition}, 2023, pp. 22\,356--22\,366.

\bibitem{yu2021semantic}
X.~Yu, L.~Zhang, and W.~Xie, ``Semantic-driven face hallucination based on residual network,'' \emph{IEEE Transactions on Biometrics, Behavior, and Identity Science}, vol.~3, no.~2, pp. 214--228, 2021.

\bibitem{hou2023semi}
H.~Hou, J.~Xu, Y.~Hou, X.~Hu, B.~Wei, and D.~Shen, ``Semi-cycled generative adversarial networks for real-world face super-resolution,'' \emph{IEEE Transactions on Image Processing}, vol.~32, pp. 1184--1199, 2023.

\bibitem{bulat2017far}
A.~Bulat and G.~Tzimiropoulos, ``How far are we from solving the 2d \& 3d face alignment problem?(and a dataset of 230,000 3d facial landmarks),'' in \emph{Proceedings of the IEEE International Conference on Computer Vision}, 2017, pp. 1021--1030.

\bibitem{bulat2018super}
------, ``Super-fan: Integrated facial landmark localization and super-resolution of real-world low resolution faces in arbitrary poses with gans,'' in \emph{Proceedings of the IEEE Conference on Computer Vision and Pattern Recognition}, 2018, pp. 109--117.

\bibitem{zafeiriou2017menpo}
S.~Zafeiriou, G.~Trigeorgis, G.~Chrysos, J.~Deng, and J.~Shen, ``The menpo facial landmark localisation challenge: A step towards the solution,'' in \emph{Proceedings of the IEEE Conference on Computer Vision and Pattern Recognition Workshops}, 2017, pp. 170--179.

\bibitem{hu2020face}
X.~Hu, W.~Ren, J.~LaMaster, X.~Cao, X.~Li, Z.~Li, B.~Menze, and W.~Liu, ``Face super-resolution guided by 3d facial priors,'' in \emph{Proceedings of the European Conference on Computer Vision}.\hskip 1em plus 0.5em minus 0.4em\relax Springer, 2020, pp. 763--780.

\bibitem{hu2021face}
X.~Hu, W.~Ren, J.~Yang, X.~Cao, D.~Wipf, B.~Menze, X.~Tong, and H.~Zha, ``Face restoration via plug-and-play 3d facial priors,'' \emph{IEEE Transactions on Pattern Analysis and Machine Intelligence}, vol.~44, no.~12, pp. 8910--8926, 2021.

\bibitem{cao2018vggface2}
Q.~Cao, L.~Shen, W.~Xie, O.~M. Parkhi, and A.~Zisserman, ``Vggface2: A dataset for recognising faces across pose and age,'' in \emph{Proceedings of the IEEE International Conference on Automatic Face \& Gesture Recognition}.\hskip 1em plus 0.5em minus 0.4em\relax IEEE, 2018, pp. 67--74.

\bibitem{li2020enhanced}
X.~Li, W.~Li, D.~Ren, H.~Zhang, M.~Wang, and W.~Zuo, ``Enhanced blind face restoration with multi-exemplar images and adaptive spatial feature fusion,'' in \emph{Proceedings of the IEEE Conference on Computer Vision and Pattern Recognition}, 2020, pp. 2706--2715.

\bibitem{dogan2019exemplar}
B.~Dogan, S.~Gu, and R.~Timofte, ``Exemplar guided face image super-resolution without facial landmarks,'' in \emph{Proceedings of the IEEE Conference on Computer Vision and Pattern Recognition Workshops}, 2019, pp. 0--0.

\bibitem{karras2019style}
T.~Karras, S.~Laine, and T.~Aila, ``A style-based generator architecture for generative adversarial networks,'' in \emph{Proceedings of the IEEE Conference on Computer Vision and Pattern Recognition}, 2019, pp. 4401--4410.

\bibitem{gu2020image}
J.~Gu, Y.~Shen, and B.~Zhou, ``Image processing using multi-code gan prior,'' in \emph{Proceedings of the IEEE conference on computer vision and pattern recognition}, 2020, pp. 3012--3021.

\bibitem{gu2022vqfr}
Y.~Gu, X.~Wang, L.~Xie, C.~Dong, G.~Li, Y.~Shan, and M.-M. Cheng, ``Vqfr: Blind face restoration with vector-quantized dictionary and parallel decoder,'' in \emph{Proceedings of the European Conference on Computer Vision}.\hskip 1em plus 0.5em minus 0.4em\relax Springer, 2022, pp. 126--143.

\bibitem{lee2020maskgan}
C.-H. Lee, Z.~Liu, L.~Wu, and P.~Luo, ``Maskgan: Towards diverse and interactive facial image manipulation,'' in \emph{Proceedings of the IEEE Conference on Computer Vision and Pattern Recognition}, 2020, pp. 5549--5558.

\bibitem{karnewar2020msg}
A.~Karnewar and O.~Wang, ``Msg-gan: Multi-scale gradients for generative adversarial networks,'' in \emph{Proceedings of the IEEE Conference on Computer Vision and Pattern Recognition}, 2020, pp. 7799--7808.

\bibitem{zhang2022pro}
Y.~Zhang, X.~Yu, X.~Lu, and P.~Liu, ``Pro-uigan: Progressive face hallucination from occluded thumbnails,'' \emph{IEEE Transactions on Image Processing}, vol.~31, pp. 3236--3250, 2022.

\bibitem{zhang2022edface}
K.~Zhang, D.~Li, W.~Luo, J.~Liu, J.~Deng, W.~Liu, and S.~Zafeiriou, ``Edface-celeb-1 m: Benchmarking face hallucination with a million-scale dataset,'' \emph{IEEE Transactions on Pattern Analysis and Machine Intelligence}, 2022.

\bibitem{zhang2022blind}
P.~Zhang, K.~Zhang, W.~Luo, C.~Li, and G.~Wang, ``Blind face restoration: Benchmark datasets and a baseline model,'' \emph{arXiv preprint arXiv:2206.03697}, 2022.

\bibitem{li2023learning}
X.~Li, S.~Zhang, S.~Zhou, L.~Zhang, and W.~Zuo, ``Learning dual memory dictionaries for blind face restoration,'' \emph{IEEE Transactions on Pattern Analysis and Machine Intelligence}, 2023.

\bibitem{cheng2021nbnet}
S.~Cheng, Y.~Wang, H.~Huang, D.~Liu, H.~Fan, and S.~Liu, ``Nbnet: Noise basis learning for image denoising with subspace projection,'' in \emph{Proceedings of the IEEE conference on computer vision and pattern recognition}, 2021, pp. 4896--4906.

\bibitem{shen2018deep}
Z.~Shen, W.-S. Lai, T.~Xu, J.~Kautz, and M.-H. Yang, ``Deep semantic face deblurring,'' in \emph{Proceedings of the IEEE Conference on Computer Vision and Pattern Recognition}, 2018, pp. 8260--8269.

\bibitem{lai2022face}
W.-S. Lai, Y.~Shih, L.-C. Chu, X.~Wu, S.-F. Tsai, M.~Krainin, D.~Sun, and C.-K. Liang, ``Face deblurring using dual camera fusion on mobile phones,'' \emph{ACM Transactions on Graphics}, vol.~41, no.~4, pp. 1--16, 2022.

\bibitem{chen2021progressive}
C.~Chen, X.~Li, L.~Yang, X.~Lin, L.~Zhang, and K.-Y.~K. Wong, ``Progressive semantic-aware style transformation for blind face restoration,'' in \emph{Proceedings of the IEEE Conference on Computer Vision and Pattern Recognition}, 2021, pp. 11\,896--11\,905.

\bibitem{he2022gcfsr}
J.~He, W.~Shi, K.~Chen, L.~Fu, and C.~Dong, ``Gcfsr: a generative and controllable face super resolution method without facial and gan priors,'' in \emph{Proceedings of the IEEE Conference on Computer Vision and Pattern Recognition}, 2022, pp. 1889--1898.

\bibitem{cai2019fcsr}
J.~Cai, H.~Han, S.~Shan, and X.~Chen, ``Fcsr-gan: Joint face completion and super-resolution via multi-task learning,'' \emph{IEEE Transactions on Biometrics, Behavior, and Identity Science}, vol.~2, no.~2, pp. 109--121, 2019.

\bibitem{yu2019can}
X.~Yu, F.~Shiri, B.~Ghanem, and F.~Porikli, ``Can we see more? joint frontalization and hallucination of unaligned tiny faces,'' \emph{IEEE Transactions on Pattern Analysis and Machine Intelligence}, vol.~42, no.~9, pp. 2148--2164, 2019.

\bibitem{kalarot2020component}
R.~Kalarot, T.~Li, and F.~Porikli, ``Component attention guided face super-resolution network: Cagface,'' in \emph{Proceedings of the IEEE Winter Conference on Applications of Computer Vision}, 2020, pp. 370--380.

\bibitem{chan2021glean}
K.~C. Chan, X.~Wang, X.~Xu, J.~Gu, and C.~C. Loy, ``Glean: Generative latent bank for large-factor image super-resolution,'' in \emph{Proceedings of the IEEE Conference on Computer Vision and Pattern Recognition}, 2021, pp. 14\,245--14\,254.

\bibitem{zhou2015learning}
E.~Zhou, H.~Fan, Z.~Cao, Y.~Jiang, and Q.~Yin, ``Learning face hallucination in the wild,'' in \emph{Proceedings of the AAAI Conference on Artificial Intelligence}, vol.~29, no.~1, 2015.

\bibitem{feng2016face}
Z.~Feng, J.~Lai, X.~Xie, D.~Yang, and L.~Mei, ``Face hallucination by deep traversal network,'' in \emph{Proceedings of the International Conference on Pattern Recognition}.\hskip 1em plus 0.5em minus 0.4em\relax IEEE, 2016, pp. 3276--3281.

\bibitem{lu2017face}
T.~Lu, H.~Wang, Z.~Xiong, J.~Jiang, Y.~Zhang, H.~Zhou, and Z.~Wang, ``Face hallucination using region-based deep convolutional networks,'' in \emph{Proceedings of the IEEE International Conference on Image Processing}.\hskip 1em plus 0.5em minus 0.4em\relax IEEE, 2017, pp. 1657--1661.

\bibitem{chen2019sequential}
Z.~Chen, J.~Lin, T.~Zhou, and F.~Wu, ``Sequential gating ensemble network for noise robust multiscale face restoration,'' \emph{IEEE Transactions on Cybernetics}, vol.~51, no.~1, pp. 451--461, 2019.

\bibitem{zhu2016deep}
S.~Zhu, S.~Liu, C.~C. Loy, and X.~Tang, ``Deep cascaded bi-network for face hallucination,'' in \emph{Proceedings of the European Conference on Computer Vision}.\hskip 1em plus 0.5em minus 0.4em\relax Springer, 2016, pp. 614--630.

\bibitem{song2017learning}
Y.~Song, J.~Zhang, S.~He, L.~Bao, and Q.~Yang, ``Learning to hallucinate face images via component generation and enhancement,'' \emph{arXiv preprint arXiv:1708.00223}, 2017.

\bibitem{cheng2021face}
F.~Cheng, T.~Lu, Y.~Wang, and Y.~Zhang, ``Face super-resolution through dual-identity constraint,'' in \emph{Proceedings of the IEEE International Conference on Multimedia and Expo}.\hskip 1em plus 0.5em minus 0.4em\relax IEEE, 2021, pp. 1--6.

\bibitem{li2020learning}
M.~Li, Z.~Zhang, J.~Yu, and C.~W. Chen, ``Learning face image super-resolution through facial semantic attribute transformation and self-attentive structure enhancement,'' \emph{IEEE Transactions on Multimedia}, vol.~23, pp. 468--483, 2020.

\bibitem{li2018face}
K.~Li, B.~Bare, B.~Yan, B.~Feng, and C.~Yao, ``Face hallucination based on key parts enhancement,'' in \emph{Proceedings of the IEEE International Conference on Acoustics, Speech and Signal Processing}.\hskip 1em plus 0.5em minus 0.4em\relax IEEE, 2018, pp. 1378--1382.

\bibitem{yu2018face}
X.~Yu, B.~Fernando, B.~Ghanem, F.~Porikli, and R.~Hartley, ``Face super-resolution guided by facial component heatmaps,'' in \emph{Proceedings of the European Conference on Computer Vision}, 2018, pp. 217--233.

\bibitem{yasarla2020deblurring}
R.~Yasarla, F.~Perazzi, and V.~M. Patel, ``Deblurring face images using uncertainty guided multi-stream semantic networks,'' \emph{IEEE Transactions on Image Processing}, vol.~29, pp. 6251--6263, 2020.

\bibitem{zhang2018image}
Y.~Zhang, K.~Li, K.~Li, L.~Wang, B.~Zhong, and Y.~Fu, ``Image super-resolution using very deep residual channel attention networks,'' in \emph{Proceedings of the European Conference on Computer Vision}, 2018, pp. 286--301.

\bibitem{liang2021swinir}
J.~Liang, J.~Cao, G.~Sun, K.~Zhang, L.~Van~Gool, and R.~Timofte, ``Swinir: Image restoration using swin transformer,'' in \emph{Proceedings of the IEEE International Conference on Computer Vision}, 2021, pp. 1833--1844.

\bibitem{cao2017attention}
Q.~Cao, L.~Lin, Y.~Shi, X.~Liang, and G.~Li, ``Attention-aware face hallucination via deep reinforcement learning,'' in \emph{Proceedings of the IEEE Conference on Computer Vision and Pattern Recognition}, 2017, pp. 690--698.

\bibitem{jiang2019atmfn}
K.~Jiang, Z.~Wang, P.~Yi, G.~Wang, K.~Gu, and J.~Jiang, ``Atmfn: Adaptive-threshold-based multi-model fusion network for compressed face hallucination,'' \emph{IEEE Transactions on Multimedia}, vol.~22, no.~10, pp. 2734--2747, 2019.

\bibitem{xin2019residual}
J.~Xin, N.~Wang, X.~Gao, and J.~Li, ``Residual attribute attention network for face image super-resolution,'' in \emph{Proceedings of the AAAI conference on artificial intelligence}, vol.~33, no.~01, 2019, pp. 9054--9061.

\bibitem{chudasama2021comsupresnet}
V.~Chudasama, K.~Nighania, K.~Upla, K.~Raja, R.~Ramachandra, and C.~Busch, ``E-comsupresnet: Enhanced face super-resolution through compact network,'' \emph{IEEE Transactions on Biometrics, Behavior, and Identity Science}, vol.~3, no.~2, pp. 166--179, 2021.

\bibitem{lu2022rethinking}
T.~Lu, Y.~Wang, Y.~Zhang, J.~Jiang, Z.~Wang, and Z.~Xiong, ``Rethinking prior-guided face super-resolution: a new paradigm with facial component prior,'' \emph{IEEE Transactions on Neural Networks and Learning Systems}, 2022.

\bibitem{wang2022faceformer}
Y.~Wang, T.~Lu, Y.~Zhang, Z.~Wang, J.~Jiang, and Z.~Xiong, ``Faceformer: aggregating global and local representation for face hallucination,'' \emph{IEEE Transactions on Circuits and Systems for Video Technology}, 2022.

\bibitem{qi2023efficient}
H.~Qi, Y.~Qiu, X.~Luo, and Z.~Jin, ``An efficient latent style guided transformer-cnn framework for face super-resolution,'' \emph{IEEE Transactions on Multimedia}, 2023.

\bibitem{LAATransformer}
G.~Li, J.~Shi, Y.~Zong, F.~Wang, T.~Wang, and Y.~Gong, ``Learning attention from attention: Efficient self-refinement transformer for face super-resolution,'' \emph{Proceedings of the International Joint Conference on Artificial Intelligence}, 2023.

\bibitem{bao2022attention}
Q.~Bao, B.~Gang, W.~Yang, J.~Zhou, and Q.~Liao, ``Attention-driven graph neural network for deep face super-resolution,'' \emph{IEEE Transactions on Image Processing}, vol.~31, pp. 6455--6470, 2022.

\bibitem{wang2021heatmap}
C.~Wang, J.~Jiang, and X.~Liu, ``Heatmap-aware pyramid face hallucination,'' in \emph{Proceedings of the IEEE International Conference on Multimedia and Expo}.\hskip 1em plus 0.5em minus 0.4em\relax IEEE, 2021, pp. 1--6.

\bibitem{huang2017wavelet}
H.~Huang, R.~He, Z.~Sun, and T.~Tan, ``Wavelet-srnet: A wavelet-based cnn for multi-scale face super resolution,'' in \emph{Proceedings of the IEEE International Conference on Computer Vision}, 2017, pp. 1689--1697.

\bibitem{hu2019face}
X.~Hu, P.~Ma, Z.~Mai, S.~Peng, Z.~Yang, and L.~Wang, ``Face hallucination from low quality images using definition-scalable inference,'' \emph{Pattern Recognition}, vol.~94, pp. 110--121, 2019.

\bibitem{dou2020pca}
H.~Dou, C.~Chen, X.~Hu, Z.~Xuan, Z.~Hu, and S.~Peng, ``Pca-srgan: Incremental orthogonal projection discrimination for face super-resolution,'' in \emph{Proceedings of the ACM International Conference on Multimedia}, 2020, pp. 1891--1899.

\bibitem{lin2020learning}
S.~Lin, J.~Zhang, J.~Pan, Y.~Liu, Y.~Wang, J.~Chen, and J.~Ren, ``Learning to deblur face images via sketch synthesis,'' in \emph{Proceedings of the AAAI Conference on Artificial Intelligence}, vol.~34, no.~07, 2020, pp. 11\,523--11\,530.

\bibitem{li2021organ}
J.~Li, B.~Bare, S.~Zhou, B.~Yan, and K.~Li, ``Organ-branched cnn for robust face super-resolution,'' in \emph{Proceedings of the IEEE International Conference on Multimedia and Expo}.\hskip 1em plus 0.5em minus 0.4em\relax IEEE, 2021, pp. 1--6.

\bibitem{wang2022propagating}
C.~Wang, J.~Jiang, Z.~Zhong, and X.~Liu, ``Propagating facial prior knowledge for multitask learning in face super-resolution,'' \emph{IEEE Transactions on Circuits and Systems for Video Technology}, vol.~32, no.~11, pp. 7317--7331, 2022.

\bibitem{lu2018attribute}
Y.~Lu, Y.-W. Tai, and C.-K. Tang, ``Attribute-guided face generation using conditional cyclegan,'' in \emph{Proceedings of the European Conference on Computer Vision}, 2018, pp. 282--297.

\bibitem{zhang2020supervised}
M.~Zhang and Q.~Ling, ``Supervised pixel-wise gan for face super-resolution,'' \emph{IEEE Transactions on Multimedia}, vol.~23, pp. 1938--1950, 2020.

\bibitem{li2022srdiff}
H.~Li, Y.~Yang, M.~Chang, S.~Chen, H.~Feng, Z.~Xu, Q.~Li, and Y.~Chen, ``Srdiff: Single image super-resolution with diffusion probabilistic models,'' \emph{Neurocomputing}, vol. 479, pp. 47--59, 2022.

\bibitem{saharia2022image}
C.~Saharia, J.~Ho, W.~Chan, T.~Salimans, D.~J. Fleet, and M.~Norouzi, ``Image super-resolution via iterative refinement,'' \emph{IEEE Transactions on Pattern Analysis and Machine Intelligence}, vol.~45, no.~4, pp. 4713--4726, 2022.

\bibitem{gao2023implicit}
S.~Gao, X.~Liu, B.~Zeng, S.~Xu, Y.~Li, X.~Luo, J.~Liu, X.~Zhen, and B.~Zhang, ``Implicit diffusion models for continuous super-resolution,'' in \emph{Proceedings of the IEEE Conference on Computer Vision and Pattern Recognition}, 2023, pp. 10\,021--10\,030.

\bibitem{chrysos2017deep}
G.~G. Chrysos and S.~Zafeiriou, ``Deep face deblurring,'' in \emph{Proceedings of the IEEE Conference on Computer Vision and Pattern Recognition Workshops}, 2017, pp. 69--78.

\bibitem{xu2017learning}
X.~Xu, D.~Sun, J.~Pan, Y.~Zhang, H.~Pfister, and M.-H. Yang, ``Learning to super-resolve blurry face and text images,'' in \emph{Proceedings of the IEEE International Conference on Computer Vision}, 2017, pp. 251--260.

\bibitem{kupyn2019deblurgan}
O.~Kupyn, T.~Martyniuk, J.~Wu, and Z.~Wang, ``Deblurgan-v2: Deblurring (orders-of-magnitude) faster and better,'' in \emph{Proceedings of the IEEE International Conference on Computer Vision}, 2019, pp. 8878--8887.

\bibitem{yang2020hifacegan}
L.~Yang, S.~Wang, S.~Ma, W.~Gao, C.~Liu, P.~Wang, and P.~Ren, ``Hifacegan: Face renovation via collaborative suppression and replenishment,'' in \emph{Proceedings of the ACM International Conference on Multimedia}, 2020, pp. 1551--1560.

\bibitem{li2022faceformer}
A.~Li, G.~Li, L.~Sun, and X.~Wang, ``Faceformer: Scale-aware blind face restoration with transformers,'' \emph{arXiv preprint arXiv:2207.09790}, 2022.

\bibitem{li2020blind}
X.~Li, C.~Chen, S.~Zhou, X.~Lin, W.~Zuo, and L.~Zhang, ``Blind face restoration via deep multi-scale component dictionaries,'' in \emph{Proceedings of the European Conference on Computer Vision}.\hskip 1em plus 0.5em minus 0.4em\relax Springer, 2020, pp. 399--415.

\bibitem{wang2022ft}
J.~Wang, S.~Chen, Z.~Wu, and Y.-G. Jiang, ``Ft-tdr: Frequency-guided transformer and top-down refinement network for blind face inpainting,'' \emph{IEEE Transactions on Multimedia}, 2022.

\bibitem{yu2020hallucinating}
X.~Yu, F.~Porikli, B.~Fernando, and R.~Hartley, ``Hallucinating unaligned face images by multiscale transformative discriminative networks,'' \emph{International Journal of Computer Vision}, vol. 128, no.~2, pp. 500--526, 2020.

\bibitem{karras2020analyzing}
T.~Karras, S.~Laine, M.~Aittala, J.~Hellsten, J.~Lehtinen, and T.~Aila, ``Analyzing and improving the image quality of stylegan,'' in \emph{Proceedings of the IEEE Conference on Computer Vision and Pattern Recognition}, 2020, pp. 8110--8119.

\bibitem{esser2021taming}
P.~Esser, R.~Rombach, and B.~Ommer, ``Taming transformers for high-resolution image synthesis,'' in \emph{Proceedings of the IEEE Conference on Computer Vision and Pattern Recognition}, 2021, pp. 12\,873--12\,883.

\bibitem{wang2022panini}
Y.~Wang, Y.~Hu, and J.~Zhang, ``Panini-net: Gan prior based degradation-aware feature interpolation for face restoration,'' in \emph{Proceedings of the AAAI Conference on Artificial Intelligence}, vol.~36, no.~3, 2022, pp. 2576--2584.

\bibitem{zhu2022blind}
F.~Zhu, J.~Zhu, W.~Chu, X.~Zhang, X.~Ji, C.~Wang, and Y.~Tai, ``Blind face restoration via integrating face shape and generative priors,'' in \emph{Proceedings of the IEEE Conference on Computer Vision and Pattern Recognition}, 2022, pp. 7662--7671.

\bibitem{hu2023dear}
Y.~Hu, Y.~Wang, and J.~Zhang, ``Dear-gan: Degradation-aware face restoration with gan prior,'' \emph{IEEE Transactions on Circuits and Systems for Video Technology}, 2023.

\bibitem{lianalyzing}
Z.~Li, D.~Zeng, X.~Yan, Q.~Shen, and B.~Tang, ``Analyzing and combating attribute bias for face restoration,'' \emph{Proceedings of the International Joint Conference on Artificial Intelligence}, 2023.

\bibitem{wang2023gan}
Y.~Wang, Y.~Hu, J.~Yu, and J.~Zhang, ``Gan prior based null-space learning for consistent super-resolution,'' in \emph{Proceedings of the AAAI Conference on Artificial Intelligence}, vol.~37, no.~3, 2023, pp. 2724--2732.

\bibitem{zhou2022towards}
S.~Zhou, K.~Chan, C.~Li, and C.~C. Loy, ``Towards robust blind face restoration with codebook lookup transformer,'' \emph{Advances in Neural Information Processing Systems}, vol.~35, pp. 30\,599--30\,611, 2022.

\bibitem{zhu2017unpaired}
J.-Y. Zhu, T.~Park, P.~Isola, and A.~A. Efros, ``Unpaired image-to-image translation using cycle-consistent adversarial networks,'' in \emph{Proceedings of the IEEE International Conference on Computer Vision}, 2017, pp. 2223--2232.

\bibitem{bulat2018learn}
A.~Bulat, J.~Yang, and G.~Tzimiropoulos, ``To learn image super-resolution, use a gan to learn how to do image degradation first,'' in \emph{Proceedings of the European Conference on Computer Vision}, 2018, pp. 185--200.

\bibitem{wang2023dr2}
Z.~Wang, Z.~Zhang, X.~Zhang, H.~Zheng, M.~Zhou, Y.~Zhang, and Y.~Wang, ``Dr2: Diffusion-based robust degradation remover for blind face restoration,'' in \emph{Proceedings of the IEEE Conference on Computer Vision and Pattern Recognition}, 2023, pp. 1704--1713.

\bibitem{wang2022zero}
Y.~Wang, J.~Yu, and J.~Zhang, ``Zero-shot image restoration using denoising diffusion null-space model,'' \emph{The Eleventh International Conference on Learning Representations}, 2023.

\bibitem{qiu2023diffbfr}
X.~Qiu, C.~Han, Z.~Zhang, B.~Li, T.~Guo, and X.~Nie, ``Diffbfr: Bootstrapping diffusion model towards blind face restoration,'' in \emph{Proceedings of the ACM International Conference on Multimedia}, 2023.

\bibitem{liu2019facial}
J.~Liu and C.~Jung, ``Facial image inpainting using multi-level generative network,'' in \emph{Proceedings of the IEEE International Conference on Multimedia and Expo}.\hskip 1em plus 0.5em minus 0.4em\relax IEEE, 2019, pp. 1168--1173.

\bibitem{zeng2022swin}
C.~Zeng, Y.~Liu, and C.~Song, ``Swin-casunet: Cascaded u-net with swin transformer for masked face restoration,'' in \emph{Proceedings of the International Conference on Pattern Recognition}.\hskip 1em plus 0.5em minus 0.4em\relax IEEE, 2022, pp. 386--392.

\bibitem{ge2020occluded}
S.~Ge, C.~Li, S.~Zhao, and D.~Zeng, ``Occluded face recognition in the wild by identity-diversity inpainting,'' \emph{IEEE Transactions on Circuits and Systems for Video Technology}, vol.~30, no.~10, pp. 3387--3397, 2020.

\bibitem{li2021swapinpaint}
H.~Li, W.~Wang, C.~Yu, and S.~Zhang, ``Swapinpaint: Identity-specific face inpainting with identity swapping,'' \emph{IEEE Transactions on Circuits and Systems for Video Technology}, vol.~32, no.~7, pp. 4271--4281, 2021.

\bibitem{zhang2022pluralistic}
Y.~Zhang, X.~Zhang, C.~Shi, X.~Wu, X.~Li, J.~Peng, K.~Cao, J.~Lv, and J.~Zhou, ``Pluralistic face inpainting with transformation of attribute information,'' \emph{IEEE Transactions on Multimedia}, 2022.

\bibitem{bai2023fine}
Y.~Bai, R.~He, W.~Tan, B.~Yan, and Y.~Lin, ``Fine-grained blind face inpainting with 3d face component disentanglement,'' in \emph{Proceedings of the IEEE International Conference on Acoustics, Speech and Signal Processing}.\hskip 1em plus 0.5em minus 0.4em\relax IEEE, 2023, pp. 1--5.

\bibitem{yang2018hallucinating}
L.~Yang, B.~Shao, T.~Sun, S.~Ding, and X.~Zhang, ``Hallucinating very low-resolution and obscured face images,'' \emph{arXiv preprint arXiv:1811.04645}, 2018.

\bibitem{gao2023jdsr}
G.~Gao, L.~Tang, F.~Wu, H.~Lu, and J.~Yang, ``Jdsr-gan: Constructing an efficient joint learning network for masked face super-resolution,'' \emph{IEEE Transactions on Multimedia}, vol.~25, pp. 1505--1512, 2023.

\bibitem{liu2023joint}
Z.~Liu, C.~Zhang, Y.~Wu, and C.~Zhang, ``Joint face completion and super-resolution using multi-scale feature relation learning,'' \emph{Journal of Visual Communication and Image Representation}, vol.~93, p. 103806, 2023.

\bibitem{zhang2021face}
Y.~Zhang, I.~W. Tsang, J.~Li, P.~Liu, X.~Lu, and X.~Yu, ``Face hallucination with finishing touches,'' \emph{IEEE Transactions on Image Processing}, vol.~30, pp. 1728--1743, 2021.

\bibitem{li2020if}
K.~Li and Q.~Zhao, ``If-gan: Generative adversarial network for identity preserving facial image inpainting and frontalization,'' in \emph{Proceedings of the IEEE International Conference on Automatic Face and Gesture Recognition}.\hskip 1em plus 0.5em minus 0.4em\relax IEEE, 2020, pp. 45--52.

\bibitem{duan2021simultaneous}
Q.~Duan, L.~Zhang, and X.~Gao, ``Simultaneous face completion and frontalization via mask guided two-stage gan,'' \emph{IEEE Transactions on Circuits and Systems for Video Technology}, vol.~32, no.~6, pp. 3761--3773, 2021.

\bibitem{yu2017hallucinating}
X.~Yu and F.~Porikli, ``Hallucinating very low-resolution unaligned and noisy face images by transformative discriminative autoencoders,'' in \emph{Proceedings of the IEEE Conference on Computer Vision and Pattern Recognition}, 2017, pp. 3760--3768.

\bibitem{abbasi2021identity}
A.~Abbasi and M.~Rahmati, ``Identity-preserving pose-robust face hallucination through face subspace prior,'' \emph{arXiv preprint arXiv:2111.10634}, 2021.

\bibitem{menon2020pulse}
S.~Menon, A.~Damian, S.~Hu, N.~Ravi, and C.~Rudin, ``Pulse: Self-supervised photo upsampling via latent space exploration of generative models,'' in \emph{Proceedings of the IEEE Conference on Computer Vision and Pattern Recognition}, 2020, pp. 2437--2445.

\bibitem{zhang2018super}
K.~Zhang, Z.~Zhang, C.-W. Cheng, W.~H. Hsu, Y.~Qiao, W.~Liu, and T.~Zhang, ``Super-identity convolutional neural network for face hallucination,'' in \emph{Proceedings of the European Conference on Computer Vision}, 2018, pp. 183--198.

\bibitem{ataer2019verification}
E.~Ataer-Cansizoglu, M.~Jones, Z.~Zhang, and A.~Sullivan, ``Verification of very low-resolution faces using an identity-preserving deep face super-resolution network,'' \emph{arXiv preprint arXiv:1903.10974}, 2019.

\bibitem{mathai2019does}
J.~Mathai, I.~Masi, and W.~AbdAlmageed, ``Does generative face completion help face recognition?'' in \emph{Proceedings of the International Conference on Biometrics}.\hskip 1em plus 0.5em minus 0.4em\relax IEEE, 2019, pp. 1--8.

\bibitem{kim2021edge}
J.~Kim, G.~Li, I.~Yun, C.~Jung, and J.~Kim, ``Edge and identity preserving network for face super-resolution,'' \emph{Neurocomputing}, vol. 446, pp. 11--22, 2021.

\bibitem{hsu2019sigan}
C.-C. Hsu, C.-W. Lin, W.-T. Su, and G.~Cheung, ``Sigan: Siamese generative adversarial network for identity-preserving face hallucination,'' \emph{IEEE Transactions on Image Processing}, vol.~28, no.~12, pp. 6225--6236, 2019.

\bibitem{bayramli2019fh}
B.~Bayramli, U.~Ali, T.~Qi, and H.~Lu, ``Fh-gan: Face hallucination and recognition using generative adversarial network,'' in \emph{Proceedings of the Neural Information Processing International Conference}.\hskip 1em plus 0.5em minus 0.4em\relax Springer, 2019, pp. 3--15.

\bibitem{huang2019wavelet}
H.~Huang, R.~He, Z.~Sun, and T.~Tan, ``Wavelet domain generative adversarial network for multi-scale face hallucination,'' \emph{International Journal of Computer Vision}, vol. 127, no. 6-7, pp. 763--784, 2019.

\bibitem{le2019selenet}
H.~A. Le and I.~A. Kakadiaris, ``Selenet: A semi-supervised low light face enhancement method for mobile face unlock,'' in \emph{Proceedings of the International Conference on Biometrics}.\hskip 1em plus 0.5em minus 0.4em\relax IEEE, 2019, pp. 1--8.

\bibitem{ding2020learning}
X.~Ding and R.~Hu, ``Learning to see faces in the dark,'' in \emph{Proceedings of the IEEE International Conference on Multimedia and Expo}.\hskip 1em plus 0.5em minus 0.4em\relax IEEE, 2020, pp. 1--6.

\bibitem{yasarla2021network}
R.~Yasarla, H.~R.~V. Joze, and V.~M. Patel, ``Network architecture search for face enhancement,'' \emph{arXiv preprint arXiv:2105.06528}, 2021.

\bibitem{qu2017robust}
C.~Qu, C.~Herrmann, E.~Monari, T.~Schuchert, and J.~Beyerer, ``Robust 3d patch-based face hallucination,'' in \emph{Proceedings of the IEEE Winter Conference on Applications of Computer Vision}.\hskip 1em plus 0.5em minus 0.4em\relax IEEE, 2017, pp. 1105--1114.

\bibitem{li20213d}
J.~Li, F.~Zhu, X.~Yang, and Q.~Zhao, ``3d face point cloud super-resolution network,'' in \emph{Proceedings of the IEEE International Joint Conference on Biometrics}.\hskip 1em plus 0.5em minus 0.4em\relax IEEE, 2021, pp. 1--8.

\bibitem{uddin2022incomplete}
K.~Uddin, T.~H. Jeong, and B.~T. Oh, ``Incomplete region estimation and restoration of 3d point cloud human face datasets,'' \emph{Sensors}, vol.~22, no.~3, p. 723, 2022.

\bibitem{gat2021identity}
N.~Gat, S.~Benaim, and L.~Wolf, ``Identity and attribute preserving thumbnail upscaling,'' in \emph{Proceedings of the IEEE International Conference on Image Processing}.\hskip 1em plus 0.5em minus 0.4em\relax IEEE, 2021, pp. 2708--2712.

\bibitem{yu2022multi}
Y.~Yu, P.~Zhang, K.~Zhang, W.~Luo, C.~Li, Y.~Yuan, and G.~Wang, ``Multi-prior learning via neural architecture search for blind face restoration,'' \emph{arXiv preprint arXiv:2206.13962}, 2022.

\bibitem{yu2018super}
X.~Yu, B.~Fernando, R.~Hartley, and F.~Porikli, ``Super-resolving very low-resolution face images with supplementary attributes,'' in \emph{Proceedings of the IEEE Conference on Computer Vision and Pattern Recognition}, 2018, pp. 908--917.

\bibitem{xin2020facial}
J.~Xin, N.~Wang, X.~Jiang, J.~Li, X.~Gao, and Z.~Li, ``Facial attribute capsules for noise face super resolution,'' in \emph{Proceedings of the AAAI Conference on Artificial Intelligence}, vol.~34, no.~07, 2020, pp. 12\,476--12\,483.

\bibitem{liu2021features}
Z.-S. Liu, W.-C. Siu, and Y.-L. Chan, ``Features guided face super-resolution via hybrid model of deep learning and random forests,'' \emph{IEEE Transactions on Image Processing}, vol.~30, pp. 4157--4170, 2021.

\bibitem{zhao2023towards}
Y.~Zhao, T.~Hou, Y.-C. Su, X.~J. Li, M.~Grundmann \emph{et~al.}, ``Towards authentic face restoration with iterative diffusion models and beyond,'' in \emph{Proceedings of the IEEE International Conference on Computer Vision}, 2023.

\bibitem{yang2023deep}
W.~Yang, Z.~Chen, C.~Chen, G.~Chen, and K.-Y.~K. Wong, ``Deep face video inpainting via uv mapping,'' \emph{IEEE Transactions on Image Processing}, vol.~32, pp. 1145--1157, 2023.

\bibitem{li2019deep}
M.~Li, Y.~Sun, Z.~Zhang, H.~Xie, and J.~Yu, ``Deep learning face hallucination via attributes transfer and enhancement,'' in \emph{Proceedings of the IEEE International Conference on Multimedia and Expo}.\hskip 1em plus 0.5em minus 0.4em\relax IEEE, 2019, pp. 604--609.

\bibitem{lee2018attribute}
C.-H. Lee, K.~Zhang, H.-C. Lee, C.-W. Cheng, and W.~Hsu, ``Attribute augmented convolutional neural network for face hallucination,'' in \emph{Proceedings of the IEEE Conference on Computer Vision and Pattern Recognition Workshops}, 2018, pp. 721--729.

\bibitem{dey20223dfacefill}
R.~Dey and V.~N. Boddeti, ``3dfacefill: An analysis-by-synthesis approach to face completion,'' in \emph{Proceedings of the IEEE Winter Conference on Applications of Computer Vision}, 2022, pp. 1586--1595.

\bibitem{chan2022glean}
K.~C. Chan, X.~Xu, X.~Wang, J.~Gu, and C.~C. Loy, ``Glean: Generative latent bank for image super-resolution and beyond,'' \emph{IEEE Transactions on Pattern Analysis and Machine Intelligence}, vol.~45, no.~3, pp. 3154--3168, 2022.

\bibitem{hou2022feature}
Z.~Hou, L.~Li, and X.~Guo, ``Feature-guided blind face restoration with gan prior,'' in \emph{Proceedings of the IEEE International Conference on Multimedia and Expo}.\hskip 1em plus 0.5em minus 0.4em\relax IEEE, 2022, pp. 1--6.

\bibitem{zhao2022rethinking}
Y.~Zhao, Y.-C. Su, C.-T. Chu, Y.~Li, M.~Renn, Y.~Zhu, C.~Chen, and X.~Jia, ``Rethinking deep face restoration,'' in \emph{Proceedings of the IEEE Conference on Computer Vision and Pattern Recognition}, 2022, pp. 7652--7661.

\end{thebibliography}

\end{document}